%% file: main.tex
\documentclass[11pt]{article}

\newif\ifreview
\reviewfalse

\input{preamble}
\hypersetup{pdftitle={The information geometry of large language models is shared, learned, and controllable},pdfauthor={Dario Picozzi}}

\begin{document}

\title{\bfseries
  The information geometry of large language models is shared, learned,
  and controllable
}
\author{Dario Picozzi\\[4pt]
  \normalsize Department of Physics and Astronomy, University College London (UCL),\\
  \normalsize Gower Street, London WC1E 6BT, United Kingdom\\[1pt]
  \normalsize London Centre for Nanotechnology, 19 Gordon St, London WC1H 0AH, United Kingdom\\[2pt]
  \normalsize \texttt{picozzi.dario@gmail.com}}
\date{}
\maketitle
\ifreview\linenumbers\fi

\begin{abstract}
Large language models learn similar behaviours, yet it remains unclear what structure they share
or how to change one behaviour without disturbing others. The Fisher--Rao geometry of next-token
probabilities connects these questions: behaviour determines this geometry up to output-preserving
symmetries, whereas activation geometry depends on coordinates. Across transformer, state-space
and recurrent models, output geometries agree more strongly than activation geometries, and shared
geometry supports semantic-category transfer. Agreement with human word choices increases with
predictive accuracy, scale and training, and improves further after model-only calibration. Token
probabilities and read-out geometry jointly predict the spectrum and its effective dimension.
Controlled language assignments show that geometry follows the language law across architectures.
Pretraining corpus statistics predict held-out fact acquisition without recalibration, while
randomised experiments show that deeper evidence substantially delays acquisition across every
tested architecture and evidence construction. Finally, the geometry prescribes
minimum-disturbance local interventions, predicts their relative cost, and supports reusable
control: updates learned on donor prompts transfer to unseen prompts while better preserving
behaviour on reference prompts than Euclidean control. The same geometric correction improves
steering, editing, attribution, dictionary learning and fine-tuning.
\end{abstract}

\input{sections/introduction}
\input{sections/results}
\input{sections/discussion}

\input{methods}

\section*{Data availability}
The public pretrained checkpoints and benchmark datasets used in the reported analyses include
the Pythia suite~\cite{pythia}, GPT-2~\cite{gpt2}, GPT-Neo~\cite{gptneo}, Qwen2~\cite{qwen2},
Qwen1.5~\cite{qwen15}, Mistral-7B~\cite{mistral7b}, Mamba~\cite{mamba},
RWKV~\cite{rwkv}, StarCoder2~\cite{starcoder2}, BLOOM~\cite{bloom},
OLMo~\cite{olmo2}, Sentence-BERT~\cite{sbert} and DINOv2~\cite{dinov2}; and WikiText~\cite{wikitext},
LAMBADA~\cite{lambada}, SST-2~\cite{sst2}, TruthfulQA~\cite{truthfulqa},
CounterFact~\cite{rome}, CIFAR-100~\cite{cifar100}, and the human prediction datasets of
Peelle et al.~\cite{peelle2020norms}, de Varda et al.~\cite{devarda2024cloze} and
DERCo~\cite{quach2024derco}. Derived data supporting the findings are
available from the corresponding author on reasonable request.

\section*{Code availability}
The matrix-free implementation and analysis scripts supporting the reported computational results
are available from the corresponding author on reasonable request.

\section*{Acknowledgements}
D.P.\ acknowledges support from the Engineering and Physical Sciences Research Council (EPSRC)
under grant numbers EP/T517793/1, EP/S021582/1, and EP/W524335/1.

\section*{Author contributions}
D.P.\ conceived the project, performed all analytical derivations and numerical experiments, and
wrote the manuscript.

\section*{Competing interests}
The author declares no competing interests.

\bibliographystyle{unsrtnat}
\bibliography{refs}

\input{extended_data}

\end{document}

\maketitle
\begingroup
\footnotesize
\tableofcontents
\endgroup
\clearpage

\paragraph{Organisation.}
The Supplementary Information follows the paper's argument. Part~I defines the canonical output
geometry, derives its origin in cross-entropy
training and establishes what predictive behaviour identifies. Part~II proves and measures the
geometry shared across independently trained models. Part~III shows how language statistics and
the location of evidence shape the geometry and its acquisition. Part~IV derives
minimal-disturbance control and gives the computational and operational results. Part~V records
the experimental designs, controls and reproducibility information. Original numbered mathematical
results are accompanied by proofs; the classical Chentsov theorem is given as an
attributed proof sketch with references to complete proofs.

\sipart{I}{Origin and identification}

\section{The canonical output metric and its pullback}\label{si:theory}
This section supports the main text's definition of the metric and the
instrument: it states canonicity and the exact chart-covariance of the
pullback solve.

\subsection{Canonicity of the Fisher--Rao metric (Chentsov's theorem)}
Let $\Delta_{n-1}$ denote the open simplex of probability distributions on $n$ outcomes. A
\emph{congruent Markov embedding} $T:\Delta_{n-1}\to\Delta_{m-1}$, $m\ge n$, is the
linear map induced by a stochastic matrix that splits each outcome $a$ into sub-outcomes with fixed
conditional probabilities and admits a stochastic left inverse that recovers $a$. Such an embedding
therefore preserves statistical information; the corresponding left inverse is a sufficient
statistic (coarse-graining) on the embedded family.

\begin{theorem}[Chentsov]
Up to a positive multiplicative constant, the Fisher--Rao metric
$g_p(u,v)=\sum_a u_a v_a / p_a$ is the unique family of Riemannian metrics on the simplices
$\{\Delta_{n-1}\}_{n\ge 2}$ invariant under every congruent Markov embedding.
\end{theorem}

\begin{proof}[Proof sketch]
Invariance under outcome permutations forces the metric at the barycentre
$p=(1/n,\dots,1/n)$ into the form $\alpha_n\sum_a u_a v_a+\beta_n(\sum_a u_a)(\sum_a v_a)$; on the
tangent space of the simplex, where $\sum_a u_a=0$, only the $\alpha_n$ term survives. Invariance
under the equal-split morphisms $\Delta_{n-1}\to\Delta_{kn-1}$ ties the constants $\alpha_n$
together across $n$ and propagates the barycentric form to every rational point, yielding
$g_p(u,v)=c\sum_a u_a v_a/p_a$ there; continuity extends this to all of $\Delta_{n-1}$, and any
invariant metric must agree with the construction, giving uniqueness up to the scale $c>0$.
A complete proof is given by Chentsov~\cite{cencov1982}; a modern treatment is given by Ay,
Jost, L\^e and Schwachh\"ofer~\cite{ay2017information}.
\end{proof}

Chentsov uniqueness is up to scale and applies within the class of
sufficient-statistic--invariant metrics; the matching unique invariant \emph{3-tensor} is the
Amari--Chentsov cubic used in Supplementary Note~\ref{si:cubic}. The metric statement is
infinitesimal: Kullback--Leibler (KL) divergence satisfies
$\mathrm{KL}(p\Vert p')=\tfrac12 d_{\mathrm{FR}}(p,p')^2+O(d^3)$, where
$d_{\mathrm{FR}}$ is Fisher--Rao distance. Finite steps may instead follow a metric geodesic or
exponential tilt. Coordinate covariance is a separate property: every Riemannian metric transforms
tensorially under a smooth reparametrisation, whereas Chentsov's
theorem is what singles out Fisher--Rao, up to scale, after requiring invariance under the statistical
transformations above.

\subsection{The pullback metric and reparametrisation invariance}\label{si:pullback}
The possibly degenerate pullback form $G=J^\top H J$ need not preserve the rank or spectrum of $H$. Under an invertible
activation reparametrisation, however, $G$ itself transforms by congruence, which preserves its rank
and inertia across charts while changing its ordinary coordinate spectrum. The null space of $G$ is the set
of activation directions the read-out maps to no change in $p$, which are invisible in the output by
construction. Under a change of activation chart $h\mapsto Ah$ the metric transforms as a tensor,
$G\mapsto A^{-\top}GA^{-1}$. The intrinsic natural gradient is the unique class on the quotient of
activation tangents by the read-out kernel; Euclidean Moore--Penrose inversion merely selects a
coordinate-dependent representative of that class. For finite damping the solved problem is
$\delta h=(G+\alpha R)^{-1}q$ with a declared reference metric $R$. If
$R\mapsto A^{-\top}RA^{-1}$, then $\delta h\mapsto A\delta h$ exactly; replacing it by a fresh
identity after the coordinate change does not preserve the problem. The current native-coordinate
implementation uses $R=I$. Thus the quotient result is canonical, while the finite damping/reference
is an explicit modelling choice, tested in the numerical coordinate analysis below.

The end-to-end reparameterisation test evaluates this distinction at the final intervention site. For
each of 12 prompts in Pythia-410M, a shared random orthogonal basis is drawn and transformations applied
with condition number $\kappa(A)=1,3,10,30,100$. Solving with
$R'=A^{-\top}IA^{-1}$ and mapping the result back reproduces the native damped direction with maximum
relative error $7.4\times10^{-13}$; the generalised spectrum of $(G',R')$ reproduces that of
$(G,I)$ to $3.9\times10^{-14}$. In contrast, setting a fresh $I$ in the transformed chart changes
the direction by a median relative error of 1.23 over non-identity transformations (minimum 0.54).
Thus, resetting the reference metric to the identity after a coordinate transformation is not
invariant, whereas transporting the native identity is covariant to numerical precision. The
transported native identity provides a coordinate-consistent reference, but is not a canonical
choice; aggregated output-Fisher and data-Mahalanobis references are evaluated separately.

\section{Cross-entropy induces and determines the output geometry}\label{si:gibbs}
This section gives the training origin of the object defined above. At every affine-softmax
checkpoint, fitting an output distribution by cross-entropy induces a Gibbs joint law in
which the read-out rows are natural parameters and the output Fisher is one of two paired
curvatures. Excess predictive risk then controls the canonical observables used throughout the
paper.

Let $P$ be the population law of a learned hidden state $h\in\R^d$ and next token
$Y\in\{1,\ldots,n\}$. Once a checkpoint is fixed, write
\begin{equation}\label{eq:si-softmax}
 q(a\mid h)=\frac{\exp(w_a^\top h+b_a)}{Z(h)},\qquad
 Z(h)=\sum_{j=1}^n\exp(w_j^\top h+b_j).
\end{equation}
No assumption about the optimiser is needed; the distributional regularity required for derivatives
of the hidden-side log-partition is stated below.

\begin{theorem}[excess-risk Gibbs duality]\label{thm:si-gibbs}
Assume the population cross-entropy of $q$ is finite and define the model-induced joint law
$Q(dh,a)=P_H(dh)q(a\mid h)$. Then:
\resultclause{Risk}
The excess population cross-entropy over the Bayes predictor is exactly
\begin{equation}\label{eq:si-gibbs-risk}
 \mathcal L(q)-H_P(Y\mid H)=\KL(P\Vert Q);
\end{equation}
\resultclause{Gibbs family}
With $d\nu(h)=dP_H(h)/Z(h)$, the joint and conditional laws are
\begin{align}
 Q(dh,a)&=\exp(w_a^\top h+b_a)d\nu(h),\label{eq:si-gibbs-joint}\\
 Q(dh\mid a)&=\exp\{w_a^\top h-A(w_a)\}d\nu(h),
 \qquad A(w)=\log\int e^{w^\top h}d\nu(h);\label{eq:si-gibbs-tilt}
\end{align}
thus all token-conditioned laws form one natural exponential family whose natural parameters are
the read-out rows;
\resultclause{Paired curvatures}
Writing $B(h)=\log\sum_a\exp(w_a^\top h+b_a)$, and assuming that $A$ is twice
differentiable in a neighbourhood of each $w_a$ for which the hidden-side moments are used,
\begin{align}
 \nabla^2B(h)&=\Cov_Q(w_Y\mid h),\label{eq:si-output-curvature}\\
 \nabla^2A(w_a)&=\Cov_Q(h\mid a),\label{eq:si-hidden-curvature}
\end{align}
and the first Hessian is exactly the output Fisher metric;
\resultclause{Classwise control}
If $\pi_a^P=P(Y=a)$, $P_a=P(H\in\cdot\mid Y=a)$ for classes with
$\pi_a^P>0$, and $Q_a$ is defined analogously, then
\begin{equation}\label{eq:si-gibbs-chain}
 \KL(P\Vert Q)=\KL(\pi^P\Vert\pi^Q)+
 \sum_{a:\pi_a^P>0}\pi_a^P\KL(P_a\Vert Q_a).
\end{equation}
Consequently, if the excess risk is $\varepsilon$ and $\norm h\le R_0$ almost surely,
\begin{align}
 \sum_{a:\pi_a^P>0}\pi_a^P\KL(P_a\Vert Q_a)&\le\varepsilon,\label{eq:si-class-kl}\\
 \sum_{a:\pi_a^P>0}\pi_a^P\norm{\E_P[h\mid a]-\nabla A(w_a)}
 &\le R_0\sqrt{2\varepsilon}.\label{eq:si-class-mean}
\end{align}
\end{theorem}

\begin{proof}
Because $P$ and $Q$ have the same hidden-state marginal,
\[
 \KL(P\Vert Q)=\E_{P_H}\KL\{P(Y\mid H)\Vert q(Y\mid H)\}
 =\E_P[-\log q(Y\mid H)]-H_P(Y\mid H),
\]
which proves Eq.~\eqref{eq:si-gibbs-risk}. Substitution of Eq.~\eqref{eq:si-softmax}
in the definition of $Q$ gives Eq.~\eqref{eq:si-gibbs-joint}. Integrating its
$a$th slice yields $\pi_a^Q=\exp\{b_a+A(w_a)\}$, and division by this marginal gives
Eq.~\eqref{eq:si-gibbs-tilt}. Differentiation of the two log-partition functions
gives their means and covariances, proving Eqs.~\eqref{eq:si-output-curvature} and
\eqref{eq:si-hidden-curvature}. Equation~\eqref{eq:si-gibbs-chain} is the chain rule
for relative entropy after factorising both laws by $Y$.

Dropping the non-negative marginal divergence in Eq.~\eqref{eq:si-gibbs-chain}
proves Eq.~\eqref{eq:si-class-kl}. For every class with $\pi_a^P>0$, Pinsker's inequality and the
boundedness of $h$ give
\[
 \norm{\E_{P_a}h-\E_{Q_a}h}
 \le 2R_0\norm{P_a-Q_a}_{\rm TV}
 \le R_0\sqrt{2\KL(P_a\Vert Q_a)}.
\]
Average with weights $\pi_a^P$, apply Cauchy--Schwarz and use
Eq.~\eqref{eq:si-class-kl}. This proves Eq.~\eqref{eq:si-class-mean}.
\end{proof}

The theorem explains why the output Fisher used in this work is tied to the training objective:
it is the conditional curvature of the Gibbs law induced by the affine-softmax read-out. The
class-conditional hidden covariance is the opposite conditional curvature of the same joint law.
The identity is exact at a checkpoint; it does not require the representation itself to have been
trained by a particular optimiser.

For a law $p$, write $p^\downarrow$ for its decreasing rearrangement and define the smooth response
at resolution $\alpha>0$ by
\[
 \dfop_p(\alpha)=\sum_k\frac{p_k^\downarrow}{p_k^\downarrow+\alpha}.
\]
For read-out rows $W=(w_a)$ let
$H_{p,W}=\Cov_p(w)$ and let
$R_W=\inf_g\max_a\norm{w_a-g}$ be their gauge-optimal radius.

\begin{theorem}[determination by predictive risk]\label{thm:si-determination}
Let $p$ and $q$ be laws on the same outcome space and put
$\rho=\norm{\sqrt p-\sqrt q}_2$. Then
\begin{align}
 \norm{p^\downarrow-q^\downarrow}_1&\le2\rho,\label{eq:si-profile-det}\\
 |\dfop_p(\alpha)-\dfop_q(\alpha)|&\le\frac{2\rho}{\alpha},\label{eq:si-response-det}\\
 \norm{H_{p,W}-H_{q,W}}_{\rm op}&\le6R_W^2\rho.\label{eq:si-metric-det}
\end{align}
Hence every eigenvalue of the fixed-read-out metric changes by at most $6R_W^2\rho$.
Moreover $\rho^2\le\KL(p\Vert q)$. If two models have excess risks
$\varepsilon_m$ and $\varepsilon_{m'}$ to a common conditional reference law, their
battery-averaged profile, response and fixed-read-out metric discrepancies are therefore bounded
at rate $O(\sqrt{\varepsilon_m}+\sqrt{\varepsilon_{m'}})$. The relational counterpart is proved
in Theorem~\ref{si:t:relational-risk}.
\end{theorem}

\begin{proof}
For scalars $x\ge x'$ and $y\ge y'$,
$|x-y|+|x'-y'|\le|x-y'|+|x'-y|$. Repeatedly uncrossing a matching shows that sorted
matching minimises total absolute deviation, so
$\norm{p^\downarrow-q^\downarrow}_1\le\norm{p-q}_1$. Cauchy--Schwarz gives
\[
 \norm{p-q}_1
 =\sum_a|\sqrt{p_a}-\sqrt{q_a}|(\sqrt{p_a}+\sqrt{q_a})
 \le2\rho,
\]
proving Eq.~\eqref{eq:si-profile-det}. Since $t\mapsto t/(t+\alpha)$ is
$1/\alpha$-Lipschitz on $[0,\infty)$, Eq.~\eqref{eq:si-response-det} follows.

Shift all read-out rows by the gauge minimiser, which leaves the covariance unchanged, so that
$\norm{w_a}\le R_W$. Put $c_p=\sum_ap_aw_a$. For any unit vector $u$,
\begin{align*}
 |u^\top(H_{p,W}-H_{q,W})u|
 &\le \sum_a|p_a-q_a|\langle w_a,u\rangle^2
   +|\langle c_p,u\rangle^2-\langle c_q,u\rangle^2|\\
 &\le 2R_W^2\rho+4R_W^2\rho.
\end{align*}
Taking the supremum proves Eq.~\eqref{eq:si-metric-det}; Weyl's inequality gives the
eigenvalue statement. Finally, with
$\mathrm{BC}(p,q)=\sum_a\sqrt{p_aq_a}$,
\[
 \KL(p\Vert q)=-2\sum_ap_a\log\sqrt{q_a/p_a}
 \ge2\{1-\mathrm{BC}(p,q)\}=\rho^2,
\]
using $\log t\le t-1$. Jensen's inequality and the triangle inequality through the
common reference law give the battery-averaged conclusion.
\end{proof}

\section{Identification of the read-out geometry}\label{si:identification}
This section supports the main text's identifiability results: local
identification with quantitative curvature, global identification with
explicit constants, and the non-identifiability of activation geometry.

Fix a strictly positive reference law $\rho$ on a common outcome space of size $V$ and define the
gauge-free weighted natural-coordinate space
\[
 D_\rho=\operatorname{diag}(\rho),\qquad
 \mathcal H_\rho=\{u\in\R^V:\sqrt\rho^{\,\top}u=0\}.
\]
For $\theta\in\mathcal H_\rho$, let
\[
 p_\theta=\operatorname{softmax}(D_\rho^{-1/2}\theta),\qquad
 H_\theta=D_\rho^{-1/2}\{\operatorname{diag}(p_\theta)-p_\theta p_\theta^\top\}
 D_\rho^{-1/2}.
\]
The Fisher operator $H_\theta$ is positive definite on $\mathcal H_\rho$. Let $X$ have finite
support with strictly positive weights, and suppose the target conditional family has the resolved
representation
\begin{equation}\label{eq:si-ident-target}
 \pi_X=p_{\theta_X},\qquad \theta_X=\beta+Uz_X,\qquad
 U^\top U=I_r,\qquad \E z_X=0,\qquad \Sigma_z=\Cov(z_X)\succ0.
\end{equation}
Write $T=\operatorname{col}(U)$. For a $d$-dimensional candidate subspace
$S\subset\mathcal H_\rho$, $d\ge r$, define the fully profiled risk
\begin{equation}\label{eq:si-profile-risk}
 \mathcal K_{D,d}(S)=\inf_{b\in\mathcal H_\rho}\inf_{f:X\to S}
 \E_X D(\pi_X,p_{b+f(X)}),\qquad D\in\{\KL,h^2\},
\end{equation}
where $h(p,q)=\norm{\sqrt p-\sqrt q}_2$. The zero-risk manifold expected for an
overcomplete read-out is
\[
 \mathcal M_d=\{S\in\operatorname{Gr}(d,\mathcal H_\rho):T\subset S\},
 \qquad
 \Delta_d(S,T)=\norm{P_{S^\perp}P_T}_{\rm F}.
\]
When differentiating the profile, fix the unique moving gauge
$b\in S^\perp\cap\mathcal H_\rho$.

\paragraph{Proof map.}
The local theorem first profiles the intercept and context coordinates and identifies the
Fisher Schur complement that supplies curvature normal to the containment fibre. The
inverse-softmax lemma then separates uniformly near candidates from candidates with a fixed
Hellinger penalty. The global theorem combines those two branches, after which the recovery and
capture corollaries translate predictive error into subspace and partition guarantees.

\begin{theorem}[local identification by profiled Fisher curvature]
\label{thm:si-local-ident}
Fix $S_0\in\mathcal M_d$ with orthonormal basis $V=(U,E)$. A Grassmann tangent is
$B=(B_T,B_E)$ with columns in $S_0^\perp\cap\mathcal H_\rho$, and the orthonormal retraction is
$V_B(t)=(V+tB)(I+t^2B^\top B)^{-1/2}$. Define
\begin{align}
 R_{X,S_0}&=H_{\theta_X}-H_{\theta_X}V(V^\top H_{\theta_X}V)^{-1}
 V^\top H_{\theta_X},\label{eq:si-schur}\\
 Q_{S_0}(B_T)&=\inf_{a\in S_0^\perp\cap\mathcal H_\rho}
 \E(B_Tz_X-a)^\top R_{X,S_0}(B_Tz_X-a).\label{eq:si-ident-Q}
\end{align}
Then
\begin{align}
 \mathcal K_{\KL,d}(\operatorname{col}V_B(t))
 &=\tfrac12t^2Q_{S_0}(B_T)+o(t^2),\label{eq:si-ident-local-kl}\\
 \mathcal K_{h^2,d}(\operatorname{col}V_B(t))
 &=\tfrac14t^2Q_{S_0}(B_T)+o(t^2).\label{eq:si-ident-local-h}
\end{align}
If
$m_H=\min_X\lambda_{\min}(H_{\theta_X}|_{\mathcal H_\rho})$ and
$m_z=\lambda_{\min}(\Sigma_z)$, then
\begin{equation}\label{eq:si-ident-local-floor}
 Q_{S_0}(B_T)\ge m_Hm_z\norm{B_T}_{\rm F}^2.
\end{equation}
Thus the Hessian kernel is exactly the tangent space of $\mathcal M_d$ and every
normal direction has strictly positive curvature.
\end{theorem}

\begin{proof}
At $S_0$, choose the identifiable intercept $b_0=P_{S_0^\perp}\beta$. Rotating the
basis produces a common normal term $BV^\top\beta$, which is cancelled exactly by
the freely optimised intercept. The remaining first-order candidate-minus-target
displacement is $B_Tz_X-a+Vc_X$, where $a$ is the normal change of the intercept and
$c_X$ is the freely optimised context coordinate. The local expansions are
\[
 \KL(p_\theta,p_{\theta+u})=\tfrac12u^\top H_\theta u+o(\norm u^2),\qquad
 h^2(p_\theta,p_{\theta+u})=\tfrac14u^\top H_\theta u+o(\norm u^2).
\]
Minimising first over $c_X$ gives the Fisher Schur complement
Eq.~\eqref{eq:si-schur}; minimising over $a$ gives Eq.~\eqref{eq:si-ident-Q}.
After the moving gauge is fixed, the nuisance Hessian is positive definite, so the
implicit-function theorem gives a unique smooth local stationary branch with the displayed
quadratic expansions.

It remains to identify this local branch with the fully profiled infimum. Its risk is $O(t^2)$.
For any sequence $t_n\to0$, a corresponding sequence of near-minimisers has mean
squared-Hellinger error tending to zero, using $h^2\le\KL$ in the KL case. Finite context support
with positive weights makes this convergence contextwise. Every target probability is positive,
so the candidate probabilities eventually lie in a common compact subset of the open simplex.
The gauge-fixed inverse-softmax map is bounded there, as are the moving-gauge intercepts and
context coordinates. Every subsequential limit is therefore an exact representation at $S_0$ and
lies in the implicit-function neighbourhood. The same compactness gives attainment for all
sufficiently small $t$; uniqueness then makes every global minimiser the local branch. This proves
Eqs.~\eqref{eq:si-ident-local-kl} and \eqref{eq:si-ident-local-h}.

For $v\in S_0^\perp\cap\mathcal H_\rho$,
\[
 v^\top R_{X,S_0}v
 =\min_c(v+Vc)^\top H_{\theta_X}(v+Vc)
 \ge m_H\min_c\norm{v+Vc}^2=m_H\norm v^2.
\]
Therefore
\[
 Q_{S_0}(B_T)\ge m_H\inf_a\E\norm{B_Tz_X-a}^2
 =m_H\E\norm{B_Tz_X}^2
 \ge m_Hm_z\norm{B_T}_{\rm F}^2,
\]
where centring makes $a=0$ optimal in the Euclidean lower bound. The form vanishes
exactly when $B_T=0$, which is precisely the tangent condition for retaining
$T\subset S$.
\end{proof}

\begin{lemma}[inverse-softmax dichotomy]\label{lem:si-inverse-softmax}
Let $p=p_\theta$ and $q=p_\eta$ for $\theta,\eta\in\mathcal H_\rho$, and suppose
$\min_i p_i\ge p_*>0$. Then either
\begin{equation}\label{eq:si-ident-near}
 h^2(p,q)\ge\frac{p_*^2}{64}\norm{\theta-\eta}^2,
\end{equation}
or
\begin{equation}\label{eq:si-ident-far}
 h^2(p,q)>\frac{p_*}{4}.
\end{equation}
The first branch holds whenever $h(p,q)\le\sqrt{p_*}/2$.
\end{lemma}

\begin{proof}
The gauge-fixed inverse-softmax chart is
$\theta=P_{\mathcal H_\rho}D_\rho^{1/2}\log p$ and similarly for $\eta$.
If $h(p,q)\le\sqrt{p_*}/2$, then coordinatewise
$q_i\ge p_*/4$. Contractivity of orthogonal projection, the mean-value theorem on
$[p_*/4,1]$ and $|p_i-q_i|\le2|\sqrt{p_i}-\sqrt{q_i}|$ give
\[
 \norm{\theta-\eta}\le\norm{\log p-\log q}_2
 \le\frac4{p_*}\norm{p-q}_2\le\frac8{p_*}h(p,q).
\]
Squaring proves Eq.~\eqref{eq:si-ident-near}. Otherwise
$h^2(p,q)>p_*/4$, which is Eq.~\eqref{eq:si-ident-far}.
\end{proof}

\begin{theorem}[global identification and the overcomplete fibre]
\label{thm:si-global-ident}
Assume $X$ has finite support, put
\[
 p_*=\min_{X,i}\pi_X(i)>0,\qquad
 w_*=\min_{x:P_X(x)>0}P_X(x)>0,
\]
and retain Eq.~\eqref{eq:si-ident-target}. For either $D=\KL$ or $D=h^2$,
\begin{equation}\label{eq:si-global-growth}
 \mathcal K_{D,d}(S)\ge c_{D,d}\Delta_d(S,T)^2,
 \qquad
 c_{D,d}:=
 \min\left\{\frac{p_*^2m_z}{64},\frac{w_*p_*}{4r}\right\}>0.
\end{equation}
Moreover $\mathcal K_{D,d}(S)=0$ if and only if $T\subset S$. Hence a rank-matched
read-out identifies the unique subspace $T$, while an overcomplete read-out is
identified exactly up to the containment fibre $\mathcal M_d$.
\end{theorem}

\begin{proof}
If $T\subset S$, choose $b=\beta$ and $f(X)=Uz_X$ in
Eq.~\eqref{eq:si-profile-risk}. Conversely consider any candidate natural parameters
$\eta_X=b+f(X)$ with $f(X)\in S$. Since $\KL\ge h^2$, it is enough to lower-bound
the Hellinger risk.

If the near branch of Lemma~\ref{lem:si-inverse-softmax} holds at every context,
orthogonal projection, $\E z_X=0$ and $\Sigma_z\succeq m_zI$ give
\begin{align*}
 \E\norm{\theta_X-\eta_X}^2
 &\ge\E\norm{P_{S^\perp}(\beta+Uz_X-b)}^2\\
 &\ge m_z\norm{P_{S^\perp}U}_{\rm F}^2
 =m_z\Delta_d(S,T)^2.
\end{align*}
The candidate risk is therefore at least
$p_*^2m_z\Delta_d(S,T)^2/64$. If the far branch holds at any context, that
context has weight at least $w_*$, so the candidate risk exceeds $w_*p_*/4$.
Since $\Delta_d(S,T)^2\le r$, this is at least
$w_*p_*\Delta_d(S,T)^2/(4r)$. Taking the smaller branch constant and then the
profile infimum proves Eq.~\eqref{eq:si-global-growth} and the stated zero set.
\end{proof}

\begin{corollary}[quantitative approximate recovery]\label{si:c:approx-recovery}
Let $q$ be the language law, $q_r$ a rank-$r$ language core, and $p_m$ a model law.
Write $R_H$ for expected squared root-probability distance and
$\Delta_d=\lVert P_{S_m^\perp}P_{T_r}\rVert_F$. Suppose
\[
 R_H(q,q_r)\le\zeta_r,
 \qquad
 \mathbb E_x\mathrm{KL}(q_x\Vert p_{m,x})\le\varepsilon_m,
 \qquad
 c_{h^2,r}\Delta_d^2\le R_H(q_r,p_m),
 \qquad c_{h^2,r}>0.
\]
Then
\begin{equation}\label{eq:si-approx-recovery}
 \Delta_d\le
 \frac{\sqrt{\zeta_r}+\sqrt{\varepsilon_m}}
      {\sqrt{c_{h^2,r}}}.
\end{equation}
For the finite-support global theorem one may take
$c_{h^2,r}=\min\{p_*^2m_z/64,\,w_*p_*/(4r)\}$ under its stated
inverse-chart, context-floor and covariance hypotheses.
\end{corollary}

\begin{proof}
Minkowski's inequality in the joint context--outcome root space and the
Hellinger--KL inequality give
\[
 \sqrt{R_H(q_r,p_m)}
 \le \sqrt{R_H(q_r,q)}+\sqrt{R_H(q,p_m)}
 \le \sqrt{\zeta_r}+\sqrt{\varepsilon_m}.
\]
The margin premise and nonnegativity give
$\sqrt{c_{h^2,r}}\,\Delta_d\le\sqrt{R_H(q_r,p_m)}$.
Division by the positive square root proves Eq.~\eqref{eq:si-approx-recovery}.
The displayed finite constant is exactly the near/far global margin in
Theorem~\ref{thm:si-global-ident}; it is not introduced by this final scalar step.
\end{proof}

\begin{corollary}[capture inherited at the recovery rate]\label{si:c:capture-recovery}
Let the rank-$r$ language core $T$ and a rank-$r$ model core $S_m$ be at chordal
distance at most $B_m$. For a fixed partition projector $0\preceq\Pi_G\preceq I$,
write $q_G(S)=\operatorname{tr}(\Pi_GP_S)/r$ for mean capture and
$q_{\min,G}(S)$ for the smallest eigenvalue of its compression to $S$. Then
\begin{equation}\label{eq:si-capture-recovery}
 q_G(S_m)\ge q_G(T)-\frac{B_m}{\sqrt r},
 \qquad
 q_{\min,G}(S_m)\ge q_{\min,G}(T)-\min\{1,2B_m\}.
\end{equation}
In particular Corollary~\ref{si:c:approx-recovery} supplies
$B_m=(\sqrt{\zeta_r}+\sqrt{\varepsilon_m})/\sqrt{c_{h^2,r}}$.
\end{corollary}

\begin{proof}
If $\theta_1,\ldots,\theta_r$ are the principal angles, $P_T-P_{S_m}$ has
nonzero eigenvalues $\pm\sin\theta_i$. Positivity and contractivity of $\Pi_G$
give
\[
 |\operatorname{tr}(\Pi_G(P_T-P_{S_m}))|
 \le\sum_i\sin\theta_i
 \le\sqrt r\left(\sum_i\sin^2\theta_i\right)^{1/2}
 =\sqrt r\,d_{\mathrm{ch}}(S_m,T).
\]
Division by $r$ proves the mean bound. For worst capture,
\[
 P\Pi_GP-Q\Pi_GQ=(P-Q)\Pi_GP+Q\Pi_G(P-Q),
\]
so the compression difference has operator norm at most
$2\lVert P-Q\rVert_{\mathrm{op}}$. Weyl's inequality and
$\lVert P-Q\rVert_{\mathrm{op}}=\max_i\sin\theta_i
\le d_{\mathrm{ch}}(S_m,T)$ give the $2B_m$ bound. Both captures lie in
$[0,1]$, which gives the independent cap by one.
\end{proof}

The measured profiled-curvature margin is positive in every evaluated configuration. The separate
fractional-frame measurements concern spectral resolution rather than identification and are
reported with the spectral measurements below.

The contrast that completes the identification picture is that no analogous
statement holds on the hidden face.

\begin{proposition}[activation geometry is not behaviourally identifiable]\label{si:p:gauge}
Let a network factor as $f=g\circ h$ through a $d$-dimensional hidden layer. For every
$A\in\mathrm{GL}(d)$, the factorisation
$f=(g\circ A^{-1})\circ(Ah)$ has identical conditional laws. If $Z$ contains sampled
activations as rows, its Gram matrix becomes $ZA^{\top}AZ^{\top}$; unless $A$ is
conformal on the sampled activation span (or on the span of activation differences for
distance geometry), Euclidean and generally cosine relational geometry changes.
Therefore behaviour does not determine a unique Euclidean activation geometry, whereas
the output geometry is a functional of the conditional laws.
\end{proposition}

\begin{proof}
The two factorisations compute exactly the same function. The Gram transformation is
direct substitution. Non-conformal choices of $A^{\top}A$ change some inner products
or squared distances whenever the sampled span has dimension at least two. In the
gauge test, re-chartings with condition numbers up to $10^4$ leave output geometry
constant to machine precision while the evaluated activation measures degrade.
\end{proof}

\paragraph{Profiled-loss experimental protocol.}
The rank-32 reference is constructed from a Pythia-6.9B teacher: its marginal law, affine mean
and leading weighted log-probability modes are estimated on 500 calibration contexts.
The target probabilities are those of this resolved reference family. Principal-vector paths
move from the reference subspace towards the read-outs of Pythia 70M, 160M, 410M and 1.4B;
one random-normal control per model preserves every principal angle. Path fractions are
$0$, $0.125$, $0.25$, $0.5$ and $1$.

At each fraction, the affine intercept and context coordinates are jointly profiled on each of
two disjoint 100-context calibration folds, followed by alternating convex coordinate/intercept
refinements. Each fitted intercept is then frozen and only coordinates are optimised on
100 held-out contexts, disjoint from both calibration folds and the 500-context reference fit.
Optimisation uses float64 and L-BFGS. Maximum absolute and root-mean-square gradient tolerances
are respectively $10^{-3}$ and $10^{-5}$ for the calibration intercept, $5\times10^{-4}$ and
$5\times10^{-5}$ for calibration coordinates, and $3\times10^{-4}$ and $3\times10^{-5}$ for
held-out coordinates. The two held-out losses are averaged. The local slope is fitted through the origin
against squared chordal distance at fractions $0.125$, $0.25$ and $0.5$, after subtracting
the zero-angle loss. Its interval uses 1,000 held-out-context bootstrap draws shared across
path fractions; fold-specific estimates are retained. This tests the resolved reference family
along the specified paths.

\subsection{Causal language assignment}
The controlled comparison crossed three architectures and two capacities with eight independent
pairs of 64-context, 64-outcome synthetic languages. Two separate pilot pairs selected the training
schedule for each architecture--capacity cell using excess predictive risk alone. The same selected
schedule was applied to every test pair. All 96 labelled test runs met the common requirement
$\epsilon<\Delta^2/32$, where $\Delta^2$ is the mean squared difference between the two true
scaled chord-distance matrices. This defines a common accuracy requirement rather than exact
equality of training loss.

For a learned distance matrix, the assigned-law advantage is its mean squared error relative to
the counterfactual law minus that relative to the assigned law, divided by $\Delta^2$. The mean
over the eight language pairs is $0.997836$ (95\% language-pair bootstrap interval
$0.997739$--$0.997933$). The normalised geometry change across laws exceeds the same-law
cross-architecture change by $0.995678$ ($0.995485$--$0.995871$). Both contrasts are positive in
all eight independent pairs (exact one-sided sign $P=1/256$).
The controlled assignment isolates the effect of the language law in
these synthetic systems.

\sipart{II}{Shared}

\section{Stability and attenuation of cross-model agreement}\label{si:stabatten}
This section supports the main text's convergence results: rank stability,
the excess-risk route to rank convergence, and the exact attenuation account
of the measured agreement.

\subsection{Stability of the shared output geometry}\label{si:stability}
This subsection proves entrywise and rank stability on a shared outcome space, and then
derives population rank convergence from vanishing excess risk under an explicit margin
condition. It also proves that behaviour does not identify a unique Euclidean activation
geometry and gives a common-outcome coarse-graining extension for different tokenizers.

\paragraph{Setting.}
A model $m$ enters only through its conditional laws $p_m(\cdot\,|\,x)$ on the context
battery; write $u_m(x) = \sqrt{p_m(\cdot\,|\,x)}$, a unit vector. The relational
geometry compared in the main text is the matrix of Fisher--Rao distances $D_m(x,y) =
2\arccos C_m(x,y)$, with $C_m(x,y) = \langle u_m(x), u_m(y)\rangle \in [0,1]$ the
Bhattacharyya affinity. Since $D$ is a strictly decreasing function of $C$, reversing
both models' entry rankings alike, every rank-agreement statistic of the distance
matrices coincides with that of the affinity matrices, so all statements are proved for
$C$. Models with a shared tokenizer share the outcome space; vocabulary positions
beyond the common range are merged into a single bucket, a deterministic
coarse-graining applied to each model (for the smaller vocabulary the bucket is empty).

\begin{lemma}[determination and per-context control]\label{si:l:lip}
(a) \emph{Determination.} $C_m$ is a functional of the conditional laws alone.
(b) \emph{Entrywise control.} For two models $m, m'$ on
a shared outcome space, let $w_z = u_{m'}(z) - u_m(z)$ and $\rho(z) = \lVert w_z
\rVert$ (the root-probability distance). Then, exactly,
\[
  e(x,y) \;:=\; C_{m'}(x,y) - C_m(x,y)
  \;=\; \langle w_x,\, u_m(y)\rangle + \langle u_{m'}(x),\, w_y\rangle ,
  \qquad
  \lvert e(x,y)\rvert \;\le\; \rho(x) + \rho(y) .
\]
(c) \emph{Risk control.} $\rho(z)^2 \le \mathrm{KL}\bigl(p_m(\cdot|z)\,\Vert\,p_{m'}(\cdot|z)\bigr)$ and
likewise with the arguments exchanged. (d) \emph{Reference triangle.} For any reference law $q(\cdot\,|\,z)$,
$\rho(z) \le \delta_m(z) + \delta_{m'}(z)$ with $\delta_m(z) = \lVert u_m(z) -
\sqrt{q(\cdot|z)}\rVert$.
\end{lemma}

\begin{proof}
(a) is immediate. (b): expand $\langle u_{m'}(x), u_{m'}(y)\rangle - \langle u_m(x),
u_m(y)\rangle$ and collect; the bound is Cauchy--Schwarz on each term with $\lVert u
\rVert = 1$. (c): with $\mathrm{BC} = \langle u_m, u_{m'}\rangle$, $\rho^2 = 2(1 -
\mathrm{BC})$, and
$\mathrm{KL}(p\,\Vert\,p') = -2\sum_a p_a \log\sqrt{p'_a/p_a} \ge -2\sum_a
p_a(\sqrt{p'_a/p_a} - 1) = 2(1 - \mathrm{BC}) = \rho^2$, using $\log t \le t - 1$.
(d) is the norm triangle inequality.
\end{proof}

\begin{theorem}[excess risk forces relational convergence]
\label{si:t:relational-risk}
Let $A$ and $B$ be two models on a common outcome space and let $q$ be a common
reference conditional law. For context $x$, put
$u_M(x)=\sqrt{p_M(\cdot\mid x)}$ and define the root-probability distance matrix
$d_M(x,y)=\norm{u_M(x)-u_M(y)}_2$. If
\[
 \varepsilon_M=\E_x\KL\{q(\cdot\mid x)\Vert p_M(\cdot\mid x)\},
\]
then, for independent contexts $X,Y$ drawn from the battery law,
\begin{equation}\label{eq:si-relational-mse}
 \E_{X,Y}\{d_A(X,Y)-d_B(X,Y)\}^2
 \le4(\varepsilon_A+\varepsilon_B).
\end{equation}
For any finite vectorisation of the distance matrices, let $a_c,b_c$ denote their
centred vectors and suppose both have non-zero norm. Then
\begin{equation}\label{eq:si-rsa-bound}
 1-\operatorname{corr}(a,b)
 \le\frac{\norm{a_c-b_c}_2^2}{2\norm{a_c}_2\norm{b_c}_2}.
\end{equation}
Thus excess-risk minimisation forces the relational geometry to converge at
root-risk rate, with the realised spread of the battery appearing only as the
conditioning factor in Eq.~\eqref{eq:si-rsa-bound}.
\end{theorem}

\begin{proof}
Put $w(x)=u_A(x)-u_B(x)$. The reverse triangle inequality gives
\[
 |d_A(x,y)-d_B(x,y)|
 \le\norm{w(x)-w(y)}.
\]
Squaring and averaging over independent contexts gives an upper bound
$2\E_x\norm{w(x)}^2-2\norm{\E_xw(x)}^2\le2\E_x\norm{w(x)}^2$.
Passing through $\sqrt q$ and applying
$(s+t)^2\le2s^2+2t^2$ yields
\[
 \E_x\norm{u_A-u_B}^2
 \le2\E_x\norm{u_A-\sqrt q}^2+2\E_x\norm{u_B-\sqrt q}^2
 \le2(\varepsilon_A+\varepsilon_B),
\]
where the final inequality is the Hellinger--KL bound. This proves
Eq.~\eqref{eq:si-relational-mse}.

For the correlation statement, expand
\[
 \norm{a_c-b_c}^2=\norm{a_c}^2+\norm{b_c}^2
 -2\langle a_c,b_c\rangle.
\]
Because $\norm{a_c}^2+\norm{b_c}^2\ge2\norm{a_c}\norm{b_c}$,
\[
 \norm{a_c-b_c}^2
 \ge2\norm{a_c}\norm{b_c}
 \{1-\operatorname{corr}(a,b)\},
\]
which rearranges to Eq.~\eqref{eq:si-rsa-bound}.
\end{proof}

\begin{theorem}[rank stability of the relational geometry]\label{si:t:rank}
Let $(x,y)$ and $(u,v)$ be two entries of the affinity matrices, with margin $\mu =
\lvert C_m(x,y) - C_m(u,v)\rvert$. If the two models order these entries discordantly,
then
\[
  \mu \;\le\; \lvert e(x,y) - e(u,v)\rvert
  \;\le\; \min\bigl\{\, \lvert e(x,y)\rvert + \lvert e(u,v)\rvert,\;\;
  \rho(x) + \rho(y) + \rho(u) + \rho(v) \,\bigr\} .
\]
Consequently the Kendall discordance rate, the probability of discordance over entry
pairs drawn uniformly from the battery, is at most $\Pr[\mu \le \lvert\Delta e\rvert]$.
\end{theorem}

\begin{proof}
Discordance means $C_m(x,y) - C_m(u,v)$ and $C_{m'}(x,y) - C_{m'}(u,v)$ have strictly
opposite signs. Their difference is $e(x,y) - e(u,v)$, and a difference of two
opposite-signed quantities has absolute value at least each of them, so
$\lvert\Delta e\rvert \ge \mu$. The upper bounds are the triangle inequality and
Lemma~\ref{si:l:lip}(b).
\end{proof}

\begin{theorem}[population excess risk forces rank convergence]\label{si:t:kl}
Let $q(\cdot|z)$ be a reference conditional law on the shared outcome space and let
$Z\sim\nu$ be the context law. Draw two compared entries $A=(X,Y)$ and $B=(U,V)$ by
any scheme for which each of $X,Y,U,V$ has marginal $\nu$. Define
\[
 R_m=\mathbb E_{Z\sim\nu}\mathrm{KL}\bigl(q(\cdot|Z)\,\Vert\,p_m(\cdot|Z)\bigr),
 \qquad
 F_q(t)=\Pr\bigl[|C_q(A)-C_q(B)|\le t\bigr].
\]
If the compared rankings have no ties, then for every $t>0$,
\[
  1-\tau(m,m')\;\le\;2F_q(t)+\frac{32(R_m+R_{m'})}{t^2}.
\]
Consequently, if $F_q(0)=0$ and $R_m+R_{m'}\to0$, then $\tau(m,m')\to1$. More
quantitatively, if $F_q(t)\le Lt^\kappa$ near zero for some $\kappa>0$, optimising in
$t$ gives $1-\tau=O((R_m+R_{m'})^{\kappa/(\kappa+2)})$.
\end{theorem}

\begin{proof}
Put $\delta_m(z)=\lVert\sqrt{p_m(\cdot|z)}-\sqrt{q(\cdot|z)}\rVert$. Lemma
\ref{si:l:lip}(c) gives $\mathbb E\delta_m(Z)^2\le R_m$. If model $m$ reverses the
reference ordering of $A$ and $B$ while the reference margin exceeds $t$, then
\[
 t<\delta_m(X)+\delta_m(Y)+\delta_m(U)+\delta_m(V)=:S_m.
\]
By Cauchy--Schwarz, $S_m^2\le4(\delta_m(X)^2+\delta_m(Y)^2+
\delta_m(U)^2+\delta_m(V)^2)$, so $\mathbb E S_m^2\le16R_m$ and Markov's inequality
gives $\Pr(S_m>t)\le16R_m/t^2$. Away from a reference tie, discordance between $m$
and $m'$ implies that at least one model reverses the reference ordering. Hence their
discordance probability is at most
$F_q(t)+16(R_m+R_{m'})/t^2$. Multiplying by two gives the Kendall bound. If
$F_q(0)=0$, choose $t\downarrow0$ slowly enough that $(R_m+R_{m'})/t^2\to0$; the
rate follows by balancing the two terms.
\end{proof}

\begin{corollary}[finite-list Spearman convergence]\label{si:c:spearman}
For two tie-free rankings of a finite list of length $n\ge2$, Spearman's rank correlation $\rho_S$ and
Kendall's $\tau$ satisfy
\[
  1-\rho_S\le3(1-\tau).
\]
Thus the conclusion of Theorem~\ref{si:t:kl} also forces $\rho_S\to1$.
\end{corollary}

\begin{proof}
For a list of length $n$, let $K$ be the inversion count and $d_i$ the rank
displacements. Since $|d_i|\le n-1$, the Spearman and Kendall formulae, together with
the footrule bound $\sum_i|d_i|\le2K$, give
\[
 1-\rho_S=\frac{6\sum_i d_i^2}{n(n^2-1)}
 \le\frac{6(n-1)\sum_i|d_i|}{n(n^2-1)}
 \le\frac{12K}{n(n+1)}
 \le\frac{12K}{n(n-1)}=3(1-\tau).
\]
\end{proof}

\begin{corollary}[a common outcome space by coarse-graining]\label{si:c:coarse}
Models may have different outcome spaces $\Omega_m$. At each context $z$, let $T_{m,z}$
be a Markov kernel from $\Omega_m$ to a common space $\Omega_0$, and suppose reference
laws satisfy $T_{m,z}q_m(\cdot|z)=q_0(\cdot|z)$. Then the preceding results apply to
$\bar p_m=T_{m,z}p_m$ on $\Omega_0$, with
\[
 \mathrm{KL}(q_0\Vert\bar p_m)\le\mathrm{KL}(q_m\Vert p_m).
\]
For tokenizers with deterministic byte decoding, mapping every token to the first newly
decoded byte after the context (with an empty/special-token bucket) is one such common coarse outcome.
This proves cross-tokenizer comparison after coarse-graining to the common outcome space;
native-vocabulary agreement remains empirical.
\end{corollary}

\begin{proof}
The Kullback--Leibler inequality is data processing under $T_{m,z}$ at each context; apply
Theorem~\ref{si:t:kl} to the pushed-forward laws.
\end{proof}

\subsection{The resolution law}\label{si:resolution-law}

The population theorem above controls average disagreement. The following result
states exactly when that control becomes a resolved count or an identical discrete
structure. It also separates the hard threshold count from the smooth response used
for effective dimension.

For a nonempty finite score vector $s=(s_1,\ldots,s_d)$ define
$\mathcal R_s(\alpha)=\{i:s_i\ge\alpha\}$ and
$D_s(\alpha)=|\mathcal R_s(\alpha)|$.

\begin{theorem}[loss, resolution and margin]\label{si:t:resolution-law}
Let
\[
 \delta_m(x)=\lVert\sqrt{p_m(\cdot|x)}-\sqrt{q(\cdot|x)}\rVert_2,
 \qquad
 \rho_m^2=\mathbb E_x\delta_m(x)^2,
 \qquad
 \varepsilon_m=\mathbb E_x\mathrm{KL}(q_x\Vert p_{m,x}).
\]
\resultclause{Risk}
Then
\begin{equation}\label{eq:si-blur-law}
 \rho_m^2\le\varepsilon_m.
\end{equation}
\resultclause{Hard counts and margins}
More generally, if finite score vectors satisfy
$\max_i|s_i^m-s_i^q|\le b_m$, then
\begin{equation}\label{eq:si-count-sandwich-generic}
 D_{s^q}(\alpha+b_m)\le D_{s^m}(\alpha)
 \le D_{s^q}(\alpha-b_m).
\end{equation}
If $\gamma_q(\alpha)=\min_i|s_i^q-\alpha|$, then
\begin{equation}\label{eq:si-margin-law}
 \gamma_q(\alpha)>\max\{b_m,b_{m'}\}
 \quad\Longrightarrow\quad
 \mathcal R_{s^m}(\alpha)=\mathcal R_{s^{m'}}(\alpha).
\end{equation}

\resultclause{Sorted probability profile}
At a fixed context, take the scores to be the sorted probabilities and let
$D_p(\alpha)=\#\{k:p_{(k)}\ge\alpha\}$. Then
\begin{equation}\label{eq:si-profile-resolution}
 D_q(\alpha+2\delta_m(x))
 \le D_{p_m}(\alpha)
 \le D_q(\alpha-2\delta_m(x)).
\end{equation}
For every $\eta>0$, Eq.~\eqref{eq:si-profile-resolution} therefore holds outside
a context set of probability at most $\eta$ after replacing
$\delta_m(x)$ by $\sqrt{\varepsilon_m/\eta}$.

\resultclause{Smooth response}
Finally, for the smooth probability response
\[
 \Phi_p(\alpha)=\sum_a\frac{p_a}{p_a+\alpha},\qquad \alpha>0,
\]
one has
\begin{align}
 |\Phi_p(\alpha)-\Phi_q(\alpha)|
 &\le\frac{2}{\alpha}\lVert\sqrt p-\sqrt q\rVert_2,
 \label{eq:si-response-pointwise}\\
 \mathbb E_x|\Phi_{p_m(x)}(\alpha)-\Phi_{p_{m'}(x)}(\alpha)|
 &\le\frac{2(\sqrt{\varepsilon_m}+\sqrt{\varepsilon_{m'}})}{\alpha}.
 \label{eq:si-response-population}
\end{align}
\end{theorem}

\begin{proof}
The coordinate Hellinger--KL inequality gives
\[
 \delta_m(x)^2\le\mathrm{KL}(q_x\Vert p_{m,x}),
 \qquad
 \E_x\delta_m(x)^2\le\varepsilon_m;
\]
the second inequality proves
Eq.~\eqref{eq:si-blur-law}.

If $s_i^q\ge\alpha+b_m$, then $s_i^m\ge\alpha$, while
$s_i^m\ge\alpha$ implies $s_i^q\ge\alpha-b_m$. These two set inclusions prove
Eq.~\eqref{eq:si-count-sandwich-generic}. If
$|s_i^q-\alpha|>b_m$, the reference score lies either above
$\alpha+b_m$ or below $\alpha-b_m$, so model $m$ has the same threshold
membership at coordinate $i$. Applying this to both models proves
Eq.~\eqref{eq:si-margin-law}.

Sorting is non-expansive in the sup norm, hence
\[
 |\sqrt{p_{m,(k)}}-\sqrt{q_{(k)}}|\le\delta_m(x).
\]
Both roots are at most one, so
\[
 |p_{m,(k)}-q_{(k)}|
 \le2\delta_m(x).
\]
Equation~\eqref{eq:si-profile-resolution} is now the generic count sandwich.
Markov's inequality applied to $\delta_m^2$ gives
$\Pr\{\delta_m>\sqrt{\varepsilon_m/\eta}\}\le\eta$, proving its population
form. This is why an RMS blur does not yield a uniform contextwise identity.

For $x,y\ge0$,
\[
 \left|\frac{x}{x+\alpha}-\frac{y}{y+\alpha}\right|
 \le\frac{|x-y|}{\alpha}.
\]
Also, by Cauchy--Schwarz,
\[
 \sum_a|p_a-q_a|
 =\sum_a|\sqrt{p_a}-\sqrt{q_a}|(\sqrt{p_a}+\sqrt{q_a})
 \le2\lVert\sqrt p-\sqrt q\rVert_2.
\]
This proves Eq.~\eqref{eq:si-response-pointwise}. Average, use
$\mathbb E\delta_m\le\rho_m\le\sqrt{\varepsilon_m}$, and pass through the
common reference law by the triangle inequality to obtain
Eq.~\eqref{eq:si-response-population}.
\end{proof}

\begin{remark}[scope of the simple margin equation]\label{si:r:resolution-scope}
There is no theorem of the form
$\gamma(F)>2(\rho_m+\rho_{m'})\Rightarrow F_m=F_{m'}$ for an arbitrary
structure $F$ when $\rho$ is an RMS population blur. A correct statement must
declare the observable, its Lipschitz constant, its decision margin and whether
the error is pointwise or aggregate. Equation~\eqref{eq:si-margin-law} is the
exact finite threshold statement. Relational ranks instead use four endpoint
errors and obey Theorem~\ref{si:t:kl}. The hard count
$D_p(\alpha)$ and the smooth response $\Phi_p(\alpha)$ are also different
objects: the former has a horizontal threshold band, while the latter has the
vertical $2/\alpha$ error bound.
\end{remark}

\begin{remark}[the coherence mechanism]\label{si:r:coherence}
On the real models the worst-case form of Theorem~\ref{si:t:rank} is vacuous: the
per-context disagreement is large (median $\rho = 0.30$--$0.71$ across the pairs
below) while the median entry-pair margin is only $0.08$--$0.14$. Observed stability
comes from low coherence, which makes the realised perturbations far smaller than the worst case
allows: the measured coherence $\lvert\langle w_x, u(y)\rangle\rvert/\rho(x)$ has
median $0.024$--$0.034$ between the well-trained pairs (90th percentile
$0.10$--$0.14$), so the direction in which two models disagree at one context is
nearly orthogonal to the affinity directions of the other contexts, and the realised
entry perturbations are 2--3\% of worst case, with margins four to seven times above
them. For calibration, a fully isotropic disagreement direction in the $|V| \approx
5\times10^4$-dimensional outcome space would give coherence of order $1/\sqrt{|V|}
\approx 0.004$; the measured values are several times that, so the disagreement is
structured, yet an order of magnitude below the level that would threaten the relational order.
Model disagreement is idiosyncratic and nearly orthogonal to the relational structure;
correspondingly, the model-specific residual of the shared geometry has negligible external semantic
alignment and does not support cross-model probe transfer.
For the weakest pair (70M against 6.9B) the coherence rises to $0.085$ and the
agreement correspondingly falls, so the observed mechanism tracks the data alongside the bound.
\end{remark}

\paragraph{Empirical evaluation.}
All quantities are computed from the cached $\sqrt p$ vectors of the natural battery
(200 contexts, full vocabulary, float64) for ten same-tokenizer pairs: the
cross-architecture trio Pythia-1.4B / Mamba-1.4B / RWKV-1.5B (attention, state-space,
recurrent; one tokenizer, one corpus), mixed-scale pairs, the Pythia ladder against
Pythia-6.9B, and GPT-2-large against GPT-Neo-1.3B (one tokenizer, two corpora). Ranks
use the affinity matrices; discordance is estimated on $4\times10^5$ sampled entry
pairs; the identity of Lemma~\ref{si:l:lip}(b) is checked to $10^{-9}$; the bound
$\rho^2 \le \mathrm{KL}$ holds at every context of every pair (median
$\rho/\sqrt{\mathrm{KL}} \approx 0.70$); the discordance bounds hold on every pair.
Supplementary Table~\ref{si:tab:stability-certificates} summarises the realised
stability and deterministic lower certificates.

\begin{table}[htbp]
\centering
\caption{\textbf{Held-out stability certificates on the natural-context battery.}
The deterministic lower certificate is evaluated with realised perturbations; it is distinct
from the worst-case excess-risk bound.}
\label{si:tab:stability-certificates}
\small
\begin{tabular}{lccccc}
\toprule
pair & $\rho$ med. & coherence med. & Spearman & $\tau$ observed & $\tau \ge$ (cert.) \\
\midrule
Pythia-1.4B / Mamba-1.4B & 0.34 & 0.028 & 0.947 & 0.816 & 0.67 \\
Pythia-1.4B / RWKV-1.5B & 0.34 & 0.028 & 0.924 & 0.797 & 0.65 \\
Mamba-1.4B / RWKV-1.5B & 0.30 & 0.024 & 0.951 & 0.829 & 0.67 \\
Pythia-410M / Mamba-1.4B & 0.42 & 0.033 & 0.926 & 0.773 & 0.62 \\
Pythia-410M / RWKV-1.5B & 0.40 & 0.034 & 0.905 & 0.759 & 0.61 \\
Pythia-70M / Pythia-6.9B & 0.71 & 0.085 & 0.777 & 0.588 & 0.42 \\
Pythia-160M / Pythia-6.9B & 0.59 & 0.085 & 0.816 & 0.636 & 0.47 \\
Pythia-410M / Pythia-6.9B & 0.44 & 0.034 & 0.907 & 0.750 & 0.59 \\
Pythia-1.4B / Pythia-6.9B & 0.35 & 0.029 & 0.934 & 0.799 & 0.65 \\
GPT-2-large / GPT-Neo-1.3B & 0.38 & 0.032 & 0.943 & 0.805 & 0.63 \\
\bottomrule
\end{tabular}
\end{table}

The deterministic lower certificate from Theorem~\ref{si:t:rank}, evaluated
a posteriori with the realised perturbations, sits $0.14$--$0.17$ below the observed
$\tau$ on every pair. It is not the worst-case excess-risk bound of
Theorem~\ref{si:t:kl}; its residual slack reflects the certificate's one-sidedness (a
large perturbation difference is necessary for a flip but not sufficient). A trajectory
consistent with the loss linkage appears along the ladder: as the weaker Pythia
model improves from 70M to 1.4B, the median symmetrised per-context KL to 6.9B falls
from $1.07$ to $0.23$, the median $\rho$ falls from $0.71$ to $0.35$ in step with
$\rho \le \sqrt{\mathrm{KL}}$, and the observed agreement rises from $0.78$ to $0.93$
(Spearman). Routing the deterministic pairwise bound through the strongest common model as reference
(Lemma~\ref{si:l:lip}(d)) costs a measured factor of $1.8$--$2.0$. The quantitative bound
applies on a shared outcome space and hence directly to same-tokenizer pairs. Cross-tokenizer
native-vocabulary agreement sits at the same level (same-tokenizer pairs agree only marginally
more) but remains empirical. Corollary~\ref{si:c:coarse} supplies a common first-byte outcome:
at this level the squared Hellinger distance does not increase on any of the 28 model pairs, different-tokenizer
agreement is $0.910$ (80\%-subsample stability interval $0.896$--$0.922$), and the
token-level anchor value is reproduced byte-exactly. The derived results comprise determination,
Lipschitz and Kullback--Leibler controls, margin rank stability, population excess-risk convergence,
and non-identifiability of a unique Euclidean activation geometry. The battery supplies the
empirical margin distribution and the coherence distribution of trained-model disagreement.
A model pair that disagreed systematically along relational directions
could agree far less at the same per-context excess risk.

\subsection{The attenuation identity and floors}\label{si:attenuation}

The name is the classical one: attenuation is the weakening of an observed correlation by noise
in the measured quantities, and the identity below is the exact form of that accounting for
relational geometries. Take a fixed reference relational map $S$ over a context battery and, for each model $m$, write
its relational map as $C_m = S + e_m$. Let $v_S=\mathrm{Var}(S)$, $v_m=\mathrm{Var}(e_m)$,
$c_m=\mathrm{Cov}(S,e_m)$ and $k_{mn}=\mathrm{Cov}(e_m,e_n)$ over a declared population of
context pairs.

\begin{theorem}[six-scalar agreement identity]\label{thm:si-sixscalar}
For arbitrary models $m\neq n$ with $\Var(C_m)>0$ and $\Var(C_n)>0$,
\[
\mathrm{Corr}(C_m, C_n)
= \frac{v_S + c_m + c_n + k_{mn}}
       {\sqrt{(v_S + v_m + 2c_m)(v_S + v_n + 2c_n)}} ,
\]
with no exchangeability, homogeneity or distributional assumption.
\end{theorem}

\begin{proof}
Expand
$\Cov(S+e_m,S+e_n)=v_S+c_m+c_n+k_{mn}$ and
$\Var(S+e_j)=v_S+v_j+2c_j$, then divide the covariance by the two standard
deviations.
\end{proof}

\begin{corollary}[exchangeable runs]\label{cor:si-exchangeable}
If the joint law of $S$ and the complete vector of run errors is invariant under
permutations of the run labels, then $v_m=v_e$, $c_m=c$ and
$k_{mn}=\gamma v_e$ for every $m\ne n$. If $v_S+2c+v_e>0$, this gives
\begin{equation}\label{eq:si-exchangeable}
 \Corr(C_m,C_n)=\frac{v_S+2c+\gamma v_e}{v_S+2c+v_e}.
\end{equation}
Its difference from the naive reliability ratio is exactly
\begin{equation}\label{eq:si-naive-reliability}
 \Corr(C_m,C_n)-\frac{v_S}{v_S+v_e}
 =\frac{v_e\{\gamma(v_S+v_e)+2c\}}
 {(v_S+v_e)(v_S+v_e+2c)}.
\end{equation}
\end{corollary}

\begin{proof}
Exchangeability equates the corresponding second moments. Substitution in
Theorem~\ref{thm:si-sixscalar} proves Eq.~\eqref{eq:si-exchangeable}; bringing the
two fractions to a common denominator proves Eq.~\eqref{eq:si-naive-reliability}.
\end{proof}

\begin{lemma}[root-affinity risk lemma]\label{lem:si-risk}
If the reference is a genuine conditional law on the same outcome experiment, both marginals of
the declared context-pair law equal the context law, and $\delta_m(x)$ is the root-probability distance at
context $x$ with risk $\varepsilon_m=\mathbb E\,\delta_m^2$, then
$|e_m|\le \delta_m(X)+\delta_m(Y)$ entrywise, $v_m \le 4\varepsilon_m$,
$|c_m| \le 2\sqrt{v_S\varepsilon_m}$ and $|k_{mn}| \le 4\sqrt{\varepsilon_m\varepsilon_n}$.
\end{lemma}

\begin{proof}
Lemma~\ref{si:l:lip} gives
$|e_m|\le\delta_m(X)+\delta_m(Y)$. Therefore
\[
 \E e_m^2\le2\E\delta_m(X)^2+2\E\delta_m(Y)^2=4\varepsilon_m.
\]
Variance is bounded by the second moment. The remaining inequalities are
Cauchy--Schwarz applied to $\Cov(S,e_m)$ and $\Cov(e_m,e_n)$.
\end{proof}

\begin{theorem}[attenuation floors]\label{thm:si-floors}
Write $u=\sqrt{v_S}$ and $w_j = 2\sqrt{\varepsilon_j}$. Whenever
$u^2 - u(w_m+w_n) - w_mw_n > 0$,
$\mathrm{Corr}(C_m,C_n) \ge \bigl(u^2-u(w_m+w_n)-w_mw_n\bigr)/\bigl((u+w_m)(u+w_n)\bigr)$;
if additionally $k_{mn}\ge 0$ and $u > w_m + w_n$, the numerator improves to $u^2-u(w_m+w_n)$.
For conditionally independent, identically distributed training runs measured against an
external reference, the run-averaged cross covariance is automatically nonnegative; an
individual realised pair may still be negative.
\end{theorem}

\begin{proof}
Lemma~\ref{lem:si-risk} bounds the numerator of
Theorem~\ref{thm:si-sixscalar} below by
$u^2-u(w_m+w_n)-w_mw_n$, improving the final term to zero when
$k_{mn}\ge0$. Its two variance factors obey
$\sqrt{v_S+v_j+2c_j}\le u+w_j$. When the numerator lower bound is positive,
division proves the displayed floors. For conditionally independent identically
distributed runs, the law of total covariance gives
\begin{align*}
 \Cov(e_m,e_n)
 &=\E\{\Cov(e_m,e_n\mid D,X,Y)\}\\
 &\quad+\Cov\{\E(e_m\mid D,X,Y),\E(e_n\mid D,X,Y)\}\\
 &=\Var\{\bar e(D,X,Y)\}\ge0.
\end{align*}
\end{proof}

\begin{proposition}[the $\boldsymbol{\pm1/2}$ idiosyncrasy signature]
\label{prop:si-half-law}
Let $r_i$ be model-specific deviation vectors in an inner-product space and put
$w_{ij}=r_i-r_j$. Then
\begin{equation}\label{eq:si-gromov-product}
 \langle w_{ij},w_{ik}\rangle
 =\tfrac12\{\norm{w_{ij}}^2+\norm{w_{ik}}^2-\norm{w_{jk}}^2\}.
\end{equation}
If distinct deviations are mutually orthogonal and have the same positive norm, then
\begin{equation}\label{eq:si-half-law}
 \cos(w_{ij},w_{ik})=\tfrac12,\qquad
 \cos(w_{ij},w_{jk})=-\tfrac12.
\end{equation}
Thus the signs distinguish edges with a common terminal endpoint from successive oriented edges;
a common disagreement axis would instead approach unit magnitude.
\end{proposition}

\begin{proof}
Equation~\eqref{eq:si-gromov-product} is the polarization identity applied to
$w_{jk}=w_{ik}-w_{ij}$. Under mutual orthogonality and common squared norm $d^2$,
each difference has squared norm $2d^2$ and
$\langle r_i-r_j,r_i-r_k\rangle=d^2$, whereas
$\langle r_i-r_j,r_j-r_k\rangle=-d^2$. Division by $2d^2$ proves
Eq.~\eqref{eq:si-half-law}.
\end{proof}

\begin{proposition}[no hidden-face floor]\label{prop:si-nohiddenfloor}
No analogous floor exists for hidden-layer geometries: there are pairs of models with identical
output behaviour, related by an exact output-preserving symmetry, whose hidden relational
geometries can have negative agreement (an explicit integer construction attains Pearson agreement
$-0.2053$ and rank agreement exactly $-4/17$ at identical outputs).
\end{proposition}

\begin{proof}
Take four hidden rows
\[
 H=\begin{pmatrix}0&2&1\\0&-1&1\\-1&-1&-2\\-1&0&0\end{pmatrix}
\]
and charts $A=\operatorname{diag}(1,3,1)$ and
$B=\operatorname{diag}(1,1,3)$. The two batteries are related by the invertible
change $HB=(HA)\operatorname{diag}(1,1/3,3)$. A read-out using only the first
coordinate is identical and non-constant in both charts. Over the six unordered
context pairs, direct inner products give
\[
 c_A=\left(-\frac{17}{\sqrt{370}},-\frac{20}{\sqrt{518}},0,
 \frac7{\sqrt{140}},0,\frac1{\sqrt{14}}\right),
\]
\[
 c_B=\left(\frac7{\sqrt{130}},-\frac{20}{\sqrt{494}},0,
 -\frac{17}{\sqrt{380}},0,\frac1{\sqrt{38}}\right).
\]
Their centred Pearson correlation is $-0.2053387\ldots$. Their average-rank
vectors are $(1,2,3.5,6,3.5,5)$ and $(6,1,3.5,2,3.5,5)$; the centred vectors
have squared norm $17$ and inner product $-4$, so Spearman agreement is exactly
$-4/17$. Since the output laws and their non-zero relational signal are identical,
no positive hidden-geometry floor can be a function only of output risk and signal.
\end{proof}

Empirically, the identity reconstructs every measured agreement on the eight-model battery (112
model-pair configurations, covering output-level and hidden-layer geometries in raw and rank
conventions, tolerance $10^{-10}$), the theorem identities and inequalities hold in all forty
per-model evaluations, and all sixty realised cross covariances are nonnegative. The floors are
valid but vacuous at the measured model-to-reference distances: realised fluctuation variances
are near $0.005$ against bounds near $1.3$. The floor calculations and held-out prediction study are documented in
Supplementary Note~\ref{si:protocols}.

The held-out prediction test measures the operational performance of the exchangeable formula.
All 120 full-sample attempts and 120{,}000 substream attempts satisfy the
numerical consistency checks. The estimator is undefined in 24 of the 120 full-sample cells
(transported denominator non-positive at steps 1--64), and the defined early-checkpoint
predictions are inaccurate (median error $0.25$--$0.31$ at steps 1--16 and $0.056$ at step 64).
From step 256 onward, where the estimator is defined and stable, the median absolute prediction
error is $0.0036$--$0.014$; the pooled defined median is $0.00938$ and the substream 95th
percentile is $0.01437$. The evidence therefore identifies the onset of an exchangeable
predictive regime during training: the estimator is accurate after step 256 but unreliable and
sometimes undefined before it.
The independently trained model deviations also
show the signature of Proposition~\ref{prop:si-half-law}: $+0.57$ and $-0.51$ on the
cross-architecture trio, and $+0.487$ and $-0.487$ across PolyPythias seeds.

\section{Shared geometry carries meaning and transfers}\label{si:meaning}

The shared component is the mean of the models' rank-transformed distance matrices; a model's
residual is its matrix after the shared component has been partialled out. Against the external
semantic geometry defined by the \texttt{all-MiniLM-L6-v2} Sentence-BERT encoder~\cite{sbert},
residual alignment is negligible on natural, templated and factual-relation batteries
($-0.001$, $+0.003$ and $+0.051$), whereas shared-component alignment is $0.044$, $0.102$ and
$0.440$. On the factual-relation battery, $+0.231$ (95\% interval $+0.159$ to $+0.316$) remains
after partialling out surface form. The shared component also contains structure beyond its
alignment with the external meaning reference.

Transfer is tested through relative representations. Each prompt is represented by its
Fisher--Rao distances to a fixed set of 40 anchors and a multinomial probe is trained on one model
and evaluated on another. Eight-way accuracy is $0.724$ within model, $0.659$ across models and
$0.700$ using the consensus alone, against chance $0.125$. The residual supports $0.407$ within
its own model but $0.091$ across models. The shared component is therefore the transferable one.

A separate activation-space analysis extends the evidence beyond textual co-occurrence. Over 100
concrete concepts, the mean cosine-distance geometry of last-token, last-hidden-state representations
from Pythia-1.4B and GPT-2 Large agrees with a vision-only DINOv2 geometry at $0.385$ (permutation
$P<0.001$), and 84\% of this association remains after controlling textual co-occurrence. The
output-geometric connection to control is supplied independently by the top spectral bands: the first five modes carry approximately half of the
steering weight, and the first 40 carry approximately 78\%; sharing and answer-relevant motion
decay over the same bands on both templated and natural batteries. The geometry that transfers
across models is therefore the geometry in which minimal-disturbance control is concentrated.

\sipart{III}{Learned}

\section{Spectral inheritance, cluster anatomy and multiscale response}\label{si:anatomy}
This section supports the main text's anatomy results: the spectrum tracks
the weighted token profile, the resolved core splits into singleton tokens
and interchangeable clusters, and the effective dimension counts spectrally resolved
structure.

\subsection{Spectral constraints and effective dimension}\label{si:inheritance}
This subsection gives exact conditional bounds relating the sorted spectrum of the output Fisher
metric to the sorted weighted token profile, an exact finite-width majorization constraint, and
the effective-dimension transform of a power-law spectrum. The hypotheses of the conditional
bounds are measurable constants of the read-out, evaluated below on the real models.

\subsubsection*{Exact finite spectral bounds}

\paragraph{Setting.}
With read-out rows $w_a$ and next-token distribution $p$, the output Fisher metric is the
$p$-weighted covariance of the read-out,
\begin{equation}\label{si:eq:cov}
  H \;=\; W_U^{\top}\bigl(\mathrm{diag}(p) - pp^{\top}\bigr)W_U
  \;=\; \sum_{a} p_a\, v_a v_a^{\top},
  \qquad v_a = w_a - \bar w, \quad \bar w = \sum_b p_b w_b ,
\end{equation}
so the centring is intrinsic: $H$ never sees the $p$-weighted mean direction, and the uncentred
second moment differs from $H$ only by the rank-one spike $\bar w\bar w^{\top}$. The natural
variables are the weighted profile $q_a=p_a\lVert v_a\rVert^2$ and, when $v_a\ne0$, the centred
unit direction $\hat v_a=v_a/\lVert v_a\rVert$; for $v_a=0$, choose any unit $\hat v_a$. Relabel so
that $q_1 \ge q_2 \ge \dots \ge 0$; since the read-out norms
vary mildly across tokens, the sorted profile carries the Zipf profile of $p$. Eigenvalues are
listed in nonincreasing order. Because the $\hat v_a$ are unit vectors,
\begin{equation}\label{si:eq:trace}
  \operatorname{tr} H \;=\; \sum_a q_a \;=:\; Q .
\end{equation}
Fix $K \le \min\{d,|V|\}$ and let $M_K = (\langle\hat v_i, \hat v_j\rangle)_{i,j\le K}$ be the Gram matrix of
the top-$K$ directions. Three measurable constants control the spectrum: the frame bounds $a_K =
\lambda_{\min}(M_K)$ and $A_K = \lambda_{\max}(M_K)$, and the tail spread $\varepsilon_K =
\lambda_{\max}(\sum_{k>K} q_k \hat v_k \hat v_k^{\top})$. All three are computable from the model
weights and one forward pass, with no randomness or independence hypothesis; for an exactly
orthonormal top-$K$ read-out, $a_K = A_K = 1$.

\begin{theorem}[inheritance band]\label{si:t:main}
For every $k \le K$, the \emph{frame lower bound} and \emph{frame-and-tail upper bound} are
\[
  a_K\, q_k \le \lambda_k(H) \le A_K\, q_k + \varepsilon_K .
\]
\end{theorem}

\begin{proof}
Split $H = H_K + H_{>K}$ into the top block $H_K = \sum_{i\le K} q_i \hat v_i\hat v_i^{\top}$ and
the tail; both are positive semidefinite.

\emph{Lower bound.} If $a_K = 0$ the bound is trivial, so assume $a_K > 0$; then $\hat v_1,\dots,
\hat v_K$ are linearly independent. Fix $k \le K$ and let $S = \mathrm{span}\{\hat v_1,\dots,\hat
v_k\}$, of dimension $k$. Let $G_k$ be the Gram matrix of the first $k$ directions; as a principal
submatrix of $M_K$ its eigenvalues lie between $a_K$ and $A_K$. Any unit $x \in S$ can be written
$x = \sum_{i\le k} c_i \hat v_i$, so that $1 = \lVert x\rVert^2 = c^{\top}G_k c$ and $\langle\hat
v_j, x\rangle = (G_k c)_j$ for $j \le k$. Diagonalising $G_k$,
\[
  \sum_{j\le k}\langle\hat v_j, x\rangle^{2} = c^{\top}G_k^{2}c
  \ge \lambda_{\min}(G_k)\, c^{\top}G_k c \ge a_K .
\]
Dropping nonnegative terms and using $q_j \ge q_k$ for $j \le k$,
\[
  x^{\top}Hx \ge x^{\top}H_K x \ge \sum_{j\le k} q_j\langle\hat v_j, x\rangle^{2}
  \ge q_k \sum_{j\le k}\langle\hat v_j, x\rangle^{2} \ge a_K q_k ,
\]
and the Courant--Fischer max--min characterisation gives $\lambda_k(H) \ge \min_{x\in S,\,\lVert
x\rVert=1} x^{\top}Hx \ge a_K q_k$.

\emph{Upper bound.} Weyl's inequality gives $\lambda_k(H) \le \lambda_k(H_K) + \varepsilon_K$. For
$\lambda_k(H_K)$, choose any $(k-1)$-dimensional subspace $W$ containing the first $k-1$
directions. For a unit $x \perp W$ the terms $j < k$
vanish, so with $U$ the $K\times d$ matrix with rows $\hat v_j^{\top}$,
\[
  x^{\top}H_K x = \sum_{j=k}^{K} q_j\langle\hat v_j, x\rangle^{2}
  \le q_k \sum_{j\le K}\langle\hat v_j, x\rangle^{2}
  = q_k\lVert Ux\rVert^{2} \le q_k\,\lambda_{\max}(UU^{\top}) = A_K q_k ,
\]
using that $U^{\top}U$ and $UU^{\top} = M_K$ have the same nonzero eigenvalues. The
Courant--Fischer min--max characterisation gives $\lambda_k(H_K) \le \max_{x\perp W,\,\lVert
x\rVert=1} x^{\top}H_K x \le A_K q_k$.
\end{proof}

\begin{theorem}[fractional-frame inheritance]\label{si:t:fractional}
For $b>0$, let
\[
 r_K(b)=\#\{j:\lambda_j(M_K)<b\}.
\]
No randomness or independence assumption is required. The following bounds hold.
\resultclause{Lower bound}
If $K\le\min\{d,|V|\}$ and
$k+r_K(b)\le K$, then
\begin{equation}\label{si:eq:fractional-lower}
 \lambda_k(H)\ge b\,q_{k+r_K(b)}.
\end{equation}
\resultclause{Direction-free upper bound}
If $0\le m<k\le d$ and
$Q_{>m}=\sum_{j>m}q_j$, then
\begin{equation}\label{si:eq:fractional-upper}
 \lambda_k(H)\le \frac{Q_{>m}}{k-m}.
\end{equation}
Thus the lower bound requires only that sufficiently many leading Gram modes remain above
resolution $b$, while the upper bound does not depend on the read-out directions.
\end{theorem}

\begin{proof}
Put $D_K=\operatorname{diag}(q_1,\ldots,q_K)$ and
$S_K=D_K^{1/2}M_KD_K^{1/2}$. Its eigenvalues are those of
$H_K=\sum_{j\le K}q_j\hat v_j\hat v_j^\top$, with zeros included, and
$H\succeq H_K$. Let $P=(bI-M_K)_+$. Then $P\succeq0$ has rank
$r=r_K(b)$ and
\[
 S_K+D_K^{1/2}PD_K^{1/2}\succeq bD_K.
\]
A positive rank-$r$ update moves an eigenvalue upward by at most $r$ positions. Hence
rank interlacing and eigenvalue monotonicity give
\[
 \lambda_k(H)\ge\lambda_k(S_K)
 \ge\lambda_{k+r}(S_K+D_K^{1/2}PD_K^{1/2})
 \ge bq_{k+r},
\]
which proves Eq.~\eqref{si:eq:fractional-lower}.

For the upper bound, write $H=A_m+B_m$, where
$A_m=\sum_{j\le m}q_j\hat v_j\hat v_j^\top$ has rank at most $m$ and
$B_m=\sum_{j>m}q_j\hat v_j\hat v_j^\top\succeq0$. Weyl's rank inequality gives
$\lambda_k(H)\le\lambda_{k-m}(B_m)$. The $(k-m)$th eigenvalue of a positive matrix
is at most its trace divided by $k-m$, and $\operatorname{tr}B_m=Q_{>m}$.
\end{proof}

\begin{corollary}[finite constant-factor certificate]\label{si:c:fractional-finite}
Fix $k$. Suppose that, for some $K\le\min\{d,|V|\}$, $b>0$, $\alpha\in(0,1]$
and $c_{\rm d},c_{\rm t}>0$,
at least $\alpha K$ eigenvalues of $M_K$ are at least $b$, $k\le\alpha K$, and
$q_K\ge c_{\rm d}q_k$. If also $Q_{>m}\le c_{\rm t}(k-m)q_k$ for some $m<k$, then
\[
 bc_{\rm d}q_k\le\lambda_k(H)\le c_{\rm t}q_k.
\]
\end{corollary}

\begin{proof}
The Gram-mode premise gives $r_K(b)\le K-\alpha K\le K-k$, so
Theorem~\ref{si:t:fractional} yields
$\lambda_k(H)\ge bq_{k+r_K(b)}\ge bq_K$. Its direction-free upper bound gives the
other inequality.
\end{proof}

\begin{corollary}[upper-outlier rank-shift refinement]\label{si:c:shift}
Fix $b_{\max}>0$ and let $r_{\uparrow}$ be the number of eigenvalues of $M_K$ above
$b_{\max}$. For $r_{\uparrow}<k\le K$,
\[
 \lambda_k(H)\le b_{\max}q_{k-r_{\uparrow}}+\varepsilon_K.
\]
Together with Theorem~\ref{si:t:fractional} at $b=b_{\min}$, this gives
\[
 b_{\min}q_{k+r_K(b_{\min})}\le\lambda_k(H)
 \le b_{\max}q_{k-r_{\uparrow}}+\varepsilon_K
\]
whenever both rank conditions hold. On a power-law profile a fixed index shift preserves the
exponent for $k$ well above the shift; exponent inheritance additionally requires control of the
additive tail term relative to $q_k$.
\end{corollary}

\begin{proof}
Write $M_K=B+P_{\uparrow}$ by clipping its eigenvalues from above at $b_{\max}$.
Then $0\preceq B\preceq b_{\max}I$ and $P_{\uparrow}\succeq0$ has rank at most
$r_{\uparrow}$. With $D_K=\operatorname{diag}(q_1,\ldots,q_K)$, rank interlacing gives
\[
 \lambda_k(D_K^{1/2}M_KD_K^{1/2})
 \le\lambda_{k-r_{\uparrow}}(D_K^{1/2}BD_K^{1/2})
 \le b_{\max}q_{k-r_{\uparrow}}.
\]
The leading block has these eigenvalues, and adding the tail contributes at most
$\varepsilon_K$ by Weyl's inequality. The lower clause is exactly
Theorem~\ref{si:t:fractional}.
\end{proof}

\begin{corollary}[cluster-deflated inheritance]\label{si:c:projected}
Let $P$ be any orthogonal projector of rank $r<d$, define
$H_{\perp}=P^{\perp}HP^{\perp}|_{P^{\perp}}$, and put
$N_{\perp}=\#\{a:P^{\perp}v_a\ne0\}$. After omitting zero projected vectors,
sort $q_a^{\perp}=p_a\lVert P^{\perp}v_a\rVert^2$ and define the corresponding
projected constants $a_K^{\perp},A_K^{\perp},\varepsilon_K^{\perp}$. If
$K\le\min\{d-r,N_{\perp}\}$, $k\le K$, and $\mu_1\ge\cdots\ge\mu_{d-r}$ are the eigenvalues of
$H_{\perp}$, then
\[
 a_K^{\perp}q_k^{\perp}\le\mu_k
 \le A_K^{\perp}q_k^{\perp}+\varepsilon_K^{\perp},
 \qquad
 \lambda_k(H)\ge\mu_k\ge\lambda_{k+r}(H).
\]
Thus removing an $r$-dimensional cluster subspace gives a valid residual-profile
certificate and locates the associated original eigenvalues up to a rank shift $r$.
\end{corollary}

\begin{proof}
Apply Theorem~\ref{si:t:main} in $P^{\perp}$, then use Poincar\'e separation for the
codimension-$r$ compression.
\end{proof}

\subsubsection*{Finite-width and response consequences}

\begin{theorem}[trace floor]\label{si:t:floor}
For every $k \le d$,
\[
  \lambda_k(H) \;\ge\; \frac{Q - \sum_{j<k}\lambda_j(H)}{d-k+1}
  \;\ge\; \frac{Q_{\ge k} - (A_K - 1)\,Q_{<k} - (k-1)\,\varepsilon_K}{d-k+1}
  \qquad (k \le K+1),
\]
where $Q_{<k} = \sum_{j<k} q_j$ and $Q_{\ge k} = Q - Q_{<k}$. The first lower bound is
the average remaining \emph{spectral} mass per remaining dimension. The second is a
profile-based relaxation, informative only when its numerator is positive; it reduces
approximately to $Q_{\ge k}/(d-k+1)$ when $A_K-1$ and $\varepsilon_K$ are sufficiently
small.
\end{theorem}

\begin{proof}
$\lambda_k$ is the largest of the $d-k+1$ trailing eigenvalues, hence at least their average, and
their sum is $Q - \sum_{j<k}\lambda_j$ by \eqref{si:eq:trace}. The second inequality applies the
upper bound of Theorem~\ref{si:t:main} to each $\lambda_j$, $j < k$.
\end{proof}

\begin{theorem}[finite-width majorization]\label{si:t:maj}
Assume the vocabulary has at least $d$ outcomes. Then
\[
  (q_1,\ldots,q_{|V|}) \;\prec\;
  (\lambda_1(H),\ldots,\lambda_d(H),0,\ldots,0).
\]
Equivalently, for every $k\le d$,
\[
  \sum_{j\le k}q_j \;\le\; \sum_{j\le k}\lambda_j(H),
  \qquad
  \sum_{j=k+1}^{d}\lambda_j(H) \;\le\; Q_{>k},
\]
with equality of the total sums. In particular,
$\sum_{j=1}^{d}(\lambda_j-q_j)=Q_{>d}$: finite width forces the profile mass beyond
rank $d$ into the first $d$ spectral modes, but majorization does not identify the
rank at which an individual eigenvalue leaves the profile.
\end{theorem}

\begin{proof}
Let $B$ be the $d\times |V|$ matrix with columns $\sqrt{q_a}\,\hat v_a$. Then
$H=BB^{\top}$, while the Gram matrix $B^{\top}B$ has diagonal $(q_a)$ and eigenvalues
$(\lambda_1(H),\ldots,\lambda_d(H),0,\ldots,0)$. The Schur--Horn theorem gives the
majorization statement. Subtracting its partial-sum inequalities from the common
total $Q$ gives the trailing-mass inequalities.
\end{proof}

\begin{remark}[why there is no profile-only trace floor]\label{si:r:floorcounter}
Take $d=2$, $p_1=p_2=1/2$, and centred directions $v_1=e_1$, $v_2=-e_1$. Then
$q=(1/2,1/2)$ but $H=e_1e_1^{\top}$ has spectrum $(1,0)$. Thus
$\lambda_k\ge Q_{\ge k}/(d-k+1)$ is false: at $k=2$ it would assert $0\ge1/2$.
Theorem~\ref{si:t:floor} instead uses the actual preceding eigenvalues, and its
profile relaxation retains the necessary frame and tail corrections.
\end{remark}

\paragraph{Profile crossover statistic.}
The quantity
\[
  k^{*}_{\mathrm{prof}} = \min\{k<d:q_k\le Q_{>k}/(d-k)\}
\]
compares a profile entry with the mean profile mass available per remaining dimension.
It is a profile-only, mean-field predictor of the measured spectral crossover. Theorems~\ref{si:t:floor}
and~\ref{si:t:maj} supply separate conditional results and do not imply this predictor. For a Zipf profile
$q_k=Z^{-1}k^{-1}$ with $|V|\gg d$, its defining equation gives
$k^{*}_{\mathrm{prof}}(1+\ln(|V|/k^{*}_{\mathrm{prof}}))\approx d$. Numerically,
doubling $d$ at realistic $(|V|,d)$ multiplies this statistic by about $2.1$--$2.3$
(for $|V|=5\times10^4$, $151$ at $d=1024$ and $343$ at $d=2048$), corresponding to a
log--log width slope about $1.1$--$1.2$.

\begin{corollary}[slope stability]\label{si:c:slope}
Suppose $\lambda_k(H) = \theta_k q_k$ with $\theta_k \in [a,b]$, $0 < a \le b$, for all $k$ in a
fit window $\mathcal W$, and let $\beta$ and $s_{\mathcal W}$ be the least-squares slopes of
$-\log\lambda_k$ and $-\log q_k$ against $\log k$ over $\mathcal W$. Then, with $x_k = \log k -
\overline{\log k}$,
\[
  \lvert\beta - s_{\mathcal W}\rvert \;\le\; \frac{\log(b/a)}{2}\,
  \frac{\sum_k \lvert x_k\rvert}{\sum_k x_k^{2}} .
\]
\end{corollary}

\begin{proof}
The least-squares slope is linear in the ordinates and $\sum_k x_k = 0$, so the deviation is
$-\sum_k x_k\eta_k/\sum_k x_k^2$ with $\eta_k = \log\theta_k$; centring $\eta$ in
$[-\tfrac12\log(b/a), +\tfrac12\log(b/a)]$ and applying the triangle inequality gives the bound. A
rank-uniform multiplicative band therefore perturbs the fitted spectral exponent by at most
$O(\log(b/a))$, and far less when the $\theta_k$ are not adversarially ordered.
\end{proof}

\begin{corollary}[bounds for effective dimension]\label{si:c:df}
Let $\mathrm{df}_H(\alpha) = \sum_{k\le d}\lambda_k/(\lambda_k+\alpha)$ (the effective dimension of
the main text) and $\mathrm{df}_q^{(K)}(\alpha) = \sum_{k\le K} q_k/(q_k+\alpha)$. If $a_K > 0$
then, for every $\alpha > 0$,
\[
  \mathrm{df}_q^{(K)}\!\Bigl(\tfrac{\alpha}{a_K}\Bigr) \;\le\; \mathrm{df}_H(\alpha) \;\le\;
  \mathrm{df}_q^{(K)}\!\Bigl(\tfrac{\alpha}{A_K}\Bigr) + \frac{d\,\bar\lambda}{\bar\lambda+\alpha},
  \qquad \bar\lambda = A_K q_K + \varepsilon_K .
\]
\end{corollary}

\begin{proof}
$f_\alpha(x) = x/(x+\alpha)$ is increasing, subadditive on $x \ge 0$ and satisfies
$f_\alpha(tq) = f_{\alpha/t}(q)$. The lower bound drops the terms $k > K$ and applies $\lambda_k
\ge a_K q_k$. For the upper bound, $\lambda_k \le \bar\lambda$ for every $k \ge K$ by
Theorem~\ref{si:t:main} and monotonicity; for $k \le K$, subadditivity gives $f_\alpha(\lambda_k)
\le f_\alpha(A_K q_k) + f_\alpha(\varepsilon_K)$, and the collected remainders total at most
$d f_\alpha(\bar\lambda)$ since $\varepsilon_K \le \bar\lambda$.
\end{proof}

If the additive remainder $d\bar\lambda/(\bar\lambda+\alpha)$ is negligible relative
to the profile term, these bounds identify the response up to the rescalings
$\alpha\mapsto\alpha/a_K$ and $\alpha\mapsto\alpha/A_K$. The condition
$\alpha\gg\bar\lambda$ alone is not sufficient when $d$ is large.

\begin{proposition}[power-law effective-dimension asymptotics]\label{si:p:dfpower}
Let $(\lambda_k)_{k\ge1}$ be a nonincreasing infinite spectrum with
$\lambda_k\sim Ck^{-s}$, $s>1$, and truncate it after $d$ terms. In an
intermediate-resolution limit in which
$n_\alpha=(C/\alpha)^{1/s}\to\infty$ and $d/n_\alpha\to\infty$,
\[
  \mathrm{df}_H(\alpha) \;\sim\; I_s\left(\frac{C}{\alpha}\right)^{1/s},
  \qquad
  I_s=\int_0^\infty\frac{dx}{1+x^s}
      =\frac{\pi}{s\sin(\pi/s)}.
\]
Thus a genuine power-law spectral window gives response exponent $\gamma=1/s$ on the
corresponding resolution window. At $s=1$ the finite-width response instead contains
a logarithmic factor, of order $n_\alpha\log(1+d/n_\alpha)$, and at fixed $d$ the
exact response tends to $\operatorname{rank}(H)$ as $\alpha\downarrow0$.
\end{proposition}

\begin{proof}
For the exact law $\lambda_k=Ck^{-s}$,
\[
 \frac{\mathrm{df}_H(\alpha)}{n_\alpha}
 =\frac1{n_\alpha}\sum_{k=1}^{d}\frac{1}{1+(k/n_\alpha)^s},
\]
which converges to $I_s$ by Riemann sums and the integrable tail for $s>1$.
Sandwiching an asymptotic power law between $(1\pm\eta)Ck^{-s}$ beyond a fixed rank,
then taking $\eta\downarrow0$, gives the stated result. The $s=1$ estimate follows by
integrating $(1+x/n_\alpha)^{-1}$, and the fixed-$d$ limit is termwise.
\end{proof}

\begin{remark}[measured constants reflect clusters]\label{si:r:measured}
On the real models the frame constants are far from $1$: at $K \approx 64$ the measured $A_K$ has
median $14$ (range $5$--$33$ over 56 curves), and on every curve at least nine Gram eigenvalues
fall below $\tfrac12$, so $a_K < \tfrac12$ throughout; the raw (uncentred) directions give the
same order. Centring in \eqref{si:eq:cov} removes the shared mean direction. The large frame
constants arise from clusters of high-probability tokens with strongly correlated read-out
directions. The Gram bulk confirms this: its median eigenvalue is $0.6$--$0.7$, with only
$2$--$4$ eigenvalues above $2$ at $K \le 72$, so a few collective modes absorb the trace while the
bulk stays near-orthonormal. Likewise $\varepsilon_K \ge q_{K+1}$ trivially, but the measured
$\varepsilon_K/q_{K+1}$ has median $26$--$54$ for $K \approx 27$--$64$, because thousands of
low-probability tokens share cluster directions. The worst-case band of
Theorem~\ref{si:t:main} is therefore loose on real read-outs; the realised spectrum nevertheless
tracks the profile tightly (below), because the large collective modes of the tail lie along the
same cluster directions as those of the top block, where Weyl's additive worst case is far from
attained. If near-parallel directions are merged into groups and the grouped frame is
well conditioned, the inheritance theorem applies to the cluster-merged profile.
This is consistent with the metric's eigenvectors being token clusters; by
Corollary~\ref{si:c:slope}, merging neighbouring weights on a power law preserves the
exponent. The exact general statements are the first bound in
Theorem~\ref{si:t:floor} and the majorization in Theorem~\ref{si:t:maj}; the
quantitative exponent statement uses the realised band in Corollary~\ref{si:c:slope}.
Corollary~\ref{si:c:shift} separately controls the multiplicative and rank-shift
effects of the Gram structure, but its additive $\varepsilon_K$ term is large for the
measured $K$ and therefore does not by itself certify tight tracking.
Conditional on a fixed profile $(q_a)$, independent uniform-spherical $\hat v_a$ would obey the standard high-probability bounds
$a_K, A_K = 1 \mp O(\sqrt{K/d})$ and $\varepsilon_K \le q_{K+1} + Q_{>K}/d +
O(\sqrt{q_K Q_{>K}\log d/d})$; real read-outs exceed these predictions by an order of magnitude through
cluster structure, which is why the hypotheses are stated as measured quantities.
\end{remark}

\paragraph{Spectral model selection and estimator validation.}
Five parametric spectral families
(single power law, broken power law, power law with exponential cutoff, stretched exponential,
log-quadratic) are fitted to each spectrum and compared by the Bayesian information criterion
(BIC), finite-sample corrected Akaike information criterion (AICc) and blocked
cross-validation, and a fixed change-point detector locates the core-to-tail crossover with
hierarchical family--model--context bootstrap intervals. A single global power law is preferred
by BIC in under $7$ percent of model--context pairs, and remains a minority choice under blocked
cross-validation; a crossover is detected in every pair; the core
window retains $R^2 = 0.99$ and the median tail exponent beyond the crossover is $1.73$. The
main-text power-law results therefore apply to the rank-$2$--$80$ core window.

Estimator validity is checked against exactly computed spectra on held-out model--context items.
The estimator and evaluation criteria were fixed before these exact spectra were revealed.
Empirical coverage of the nominal 95\% percentile-bootstrap intervals is $0.90$ for continuous
effective dimensions and $0.85$ for discrete threshold counts, so the primary dimension
statements use the continuous forms.

\paragraph{Finite-size prediction protocol.}
A separate finite-size effective-dimension confirmation uses 80 model--prompt curves from ten models in seven
families. For each curve, the bounded effective-dimension curve
$N_{\mathrm{eff}}/d=[1+(c/c^{\star})^{\gamma}]^{-1}$ and an equal-parameter unbroken power curve
are fitted by nonlinear least squares at four fixed anchor dampings
$c\in\{10^{-2},10^{-3},10^{-4},10^{-5}\}$, where $c=\alpha/\lambda_{\max}$, and scored by
root-mean-square error at the three interleaved held-out dampings
$c\in\{3\times10^{-3},3\times10^{-4},3\times10^{-5}\}$. This gives 240 held-out effective-dimension
values. The uncertainty interval for the median curve-level error ratio uses 5,000 crossed draws
that resample families, one model within each selected family and one common prompt-index multiset
across the sampled models.

\paragraph{Synthetic inheritance controls.}
Synthetic
controls draw $512$ read-out directions in $d = 1024$ (independent Gaussian, unit or log-normal
norms, and a coherent control dominated by a rank-$8$ component), assign Zipf$(s)$ profiles with
$s \in [0.6, 1.5]$ under random token permutations, and fit $\hat\beta$ with the identical
protocol (three seeds).

\subsubsection*{Evaluation on real models}

Evaluation began with 64 held-out effective-dimension curves with vocabulary $\le 160$k and width $\le 2048$
(8 prompts $\times$ 8 architectures: GPT-2, GPT-2-XL, GPT-Neo-125M/1.3B, Mamba-130M/1.4B,
Qwen2-0.5B, RWKV-1.5B). The stored next-token distributions were independently reproduced within
relative tolerance $3\times10^{-4}$ for 56 curves; the eight RWKV curves were excluded before
spectrum analysis. Constants and bounds were then evaluated in dense float64 on a 12-point
logarithmic grid in $K$. There are zero violations of Theorem~\ref{si:t:main} across all included curves, grid values and
ranks. Over the exponent-fit window (ranks 8--64) the ratio $\lambda_k/q_k$ has per-curve median
$0.87$ and median extremes $[0.69, 1.13]$; Corollary~\ref{si:c:slope} with this measured band and
window constant $1.50$ gives the worst-case bound $\lvert\beta - s\rvert \le 0.37$, and the
measured inheritance ($0.04$ median) sits well inside it. The profile crossover statistic
$k^{*}_{\mathrm{prof}}$ matches the measured broken-power crossover rank with median ratio $0.93$
(Spearman $0.48$, $n = 56$); per-model medians track the width empirically, from $k^{*}_{\mathrm{prof}} =
172$--$229$ (measured $264$--$319$) at $d = 768$ through $557$ (measured $551$) at $d = 1600$ to
$776$--$852$ (measured $597$--$718$) at $d = 2048$.

The fractional-frame certificate of Theorem~\ref{si:t:fractional}, evaluated at
$k=64$ and $b=1/2$, holds in all 56 evaluated curves in both the standardised and native
read-out representations. At the smallest admissible truncation in each representation, the median surviving
Gram-mode fractions are $0.700$ and $0.572$, at median $K=112$ and $K=145$, respectively.
These different-truncation statistics quantify spectral resolution rather than identification.

\subsubsection*{Prediction on a held-out battery across external model families}

Design choices were fixed using Pythia and GPT-2, before exact spectra were revealed for a
text-disjoint 64-context battery in nine external-family models: two BLOOM, two GPT-Neo, two
Mamba, two Qwen2 and one RWKV checkpoint. The predictor uses the standardised centred
read-out representation. Writing $u_a$ for its rows and $\bar u=\sum_a p_a u_a$, each model--context cell
has centred weighted profile $q_a=p_a\lVert u_a-\bar u\rVert^2$. The construction sets
$\widehat\lambda_k=q_{(k)}$ and applies the effective-dimension transform to its leading $r$ values,
giving $\widehat N_{\mathrm{eff}}(\alpha)=\sum_{k=1}^{r}
q_{(k)}/(q_{(k)}+\alpha)$ at ten fixed relative dampings. The measured effective dimension is instead given by the
identity $N_{\mathrm{eff}}(\alpha)=\sum_k\lambda_k/(\lambda_k+\alpha)$ evaluated after the exact
spectrum is revealed. No eigenvalue or fitted parameter enters the prediction.

Supplementary Table~\ref{si:tab:external-profile} reports family medians and crossed-bootstrap
95\% intervals over the full model-by-context grid. The certificate column is the fraction of cells admitting a measured
frame-and-tail inheritance certificate. The final three columns are multiplicative reductions in
paired error relative to a same-trace flat spectrum, a random permutation of profile ranks, and an
effective-dimension curve induced by the same-trace flat spectrum, respectively, on a separately
fixed 16-context-per-model battery.

\begin{table}[htbp]
\centering
\caption{\textbf{Profile prediction on a held-out battery across external model families.}
Errors and certificate rates are family summaries; the final three columns give paired error
reductions relative to the indicated matched alternatives.}
\label{si:tab:external-profile}
\scriptsize
\setlength{\tabcolsep}{1.0pt}
\begin{tabular}{lcccccccc}
\toprule
family & models/cells & spectral error [95\%] & $|\Delta\beta|$ [95\%] & \shortstack{effective-dimension\\RMSE [95\%]} & certificate & flat $\lambda$ & shuffled $q$ & \shortstack{flat-spectrum\\$N_{\mathrm{eff}}$} \\
\midrule
BLOOM   & 2/128 & 0.0617 [0.0436, 0.1024] & 0.0856 [0.0422, 0.1105] & 0.0270 [0.0201, 0.0371] & 99.2\% & 6.208$\times$ & 19.427$\times$ & 7.831$\times$ \\
GPT-Neo & 2/128 & 0.0771 [0.0627, 0.1055] & 0.0623 [0.0469, 0.1024] & 0.0312 [0.0221, 0.0389] & 100\%  & 4.893$\times$ & 13.547$\times$ & 7.881$\times$ \\
Mamba   & 2/128 & 0.0852 [0.0558, 0.1176] & 0.0734 [0.0532, 0.0927] & 0.0460 [0.0281, 0.0532] & 99.2\% & 5.422$\times$ & 14.986$\times$ & 6.439$\times$ \\
Qwen2   & 2/128 & 0.0884 [0.0680, 0.1084] & 0.0782 [0.0571, 0.0989] & 0.0360 [0.0269, 0.0421] & 98.4\% & 6.324$\times$ & 17.084$\times$ & 7.853$\times$ \\
RWKV    & 1/64  & 0.0833 [0.0512, 0.1118] & 0.0758 [0.0572, 0.0962] & 0.0521 [0.0461, 0.0569] & 100\%  & 9.224$\times$ & 25.630$\times$ & 6.533$\times$ \\
\bottomrule
\end{tabular}
\end{table}

\begin{figure}[htbp]
  \centering
  \makebox[\linewidth]{%
    \vtop{\hbox{\makebox[2.547in][l]{{\fontsize{8}{9}\selectfont\sffamily\bfseries a}}}%
      \vskip 1.5pt\hbox{\makebox[2.547in][c]{\includegraphics{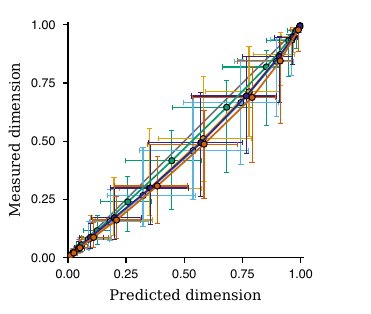}}}}%
    \hspace{0.20in}%
    \vtop{\hbox{\makebox[2.547in][l]{{\fontsize{8}{9}\selectfont\sffamily\bfseries b}}}%
      \vskip 1.5pt\hbox{\makebox[2.547in][c]{\includegraphics{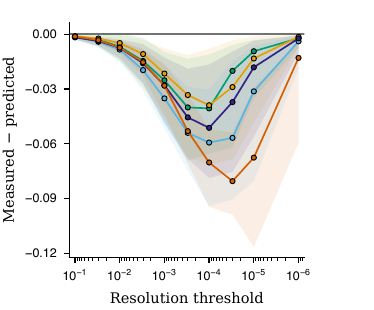}}}}}\par\vspace{2pt}
  \includegraphics{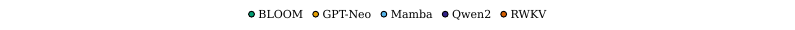}
  \caption{\textbf{Full calibration and residual structure of the token-profile effective-dimension prediction.}
  \textbf{a}, All 5,760 model--context--resolution values from 576 model--context cells in the
  held-out battery are summarized as 50 family-by-resolution points. Opaque coloured circles with
  black borders compare family medians of the normalised effective dimension predicted from the sorted token
  profile with that measured from the exact spectrum; horizontal and vertical bars are the
  corresponding interquartile ranges, point--lines join the ten fixed resolution thresholds and
  the grey line is exact prediction.
  \textbf{b}, Measured minus predicted normalised effective dimension as the relative resolution threshold $\alpha/q_{(1)}$ decreases from
  left to right. Lines are family medians and ribbons are interquartile ranges. The modest
  mid-resolution overprediction is resolved again at the coarse and fine endpoints.}
  \label{si:fig:profile-calibration}
  \label{fig:si-profile-response-diagnostics}
\end{figure}

Across the full 1,664-cell evaluation, the trace identity, profile reproduction, forward
reproduction and majorisation checks have zero violations, as do all applicable inheritance
bands. Exact certificates are available in 98.4--100\% of each external family's cells; where no
admissible frame truncation exists, no band is asserted. The profile prediction generalises on the
held-out battery across all five external families: it reproduces the spectral profile and effective-dimension curve and
outperforms the matched flat-spectrum comparator, the rank-shuffled comparator and the effective-dimension
curve induced by the same-trace flat spectrum in every family. The matched
alternatives quantify the accuracy lost when spectral structure is removed.
Supplementary Figure~\ref{si:fig:profile-calibration} shows the full calibration and the
resolution-dependent residual structure.

\FloatBarrier

\paragraph{Measured-spectrum effective-dimension identity.}
Distribution reproduction was attempted for 80 held-out effective-dimension curves against the stored
top-512 probabilities at relative tolerance $3\times10^{-4}$. Seventy-two curves from nine models
were reproduced to that tolerance and comprise the measured-spectrum effective-dimension analysis; the eight RWKV
curves were excluded. For
each included curve, the dense output Fisher was formed in float64 and eigendecomposed, and the
effective dimension was evaluated at the stored relative dampings with no fitted parameters. The
median held-out error is $0.00245$, against $0.00473$ for the fitted bounded law on the same curves;
the dense core exponent (median $1.155$) confirms the
previously measured $1.14$; and the fitted crossover $\alpha^{*}$ coincides with the resolution at which the
resolved count reaches half its ceiling (Spearman $1.000$, median ratio $1.06$). The
effective dimension is therefore exactly determined by the measured spectrum;
the two-parameter Hill curve accurately interpolates it.

\paragraph{Origin of the tail steepening.}
A factorial attribution varies the profile (real $p$ against a pure power law at the matched core
slope) and the read-out (real $W_U$ against an independent Gaussian read-out with matched column
norms) at three widths ($d$, $d/2$, $d/4$ by column subsampling), for models with vocabulary
$\le 160$k. The steepening of the spectral tail beyond the core window survives the random
read-out with the real profile (median $+0.47$) and disappears for the pure power-law profile
with the random read-out ($-0.07$), ruling out a finite-width random-matrix edge; the real
read-out contributes a secondary component ($+0.27$). The crossover rank scales with width with a
log--log slope of $1.1$--$1.3$ in every cell: the width sets the position of the band edge, the
profile's high-rank depletion sets the shape of the tail.

\subsection{Separating probability concentration from read-out structure}
The matched mechanism panel compared the weighted profile $q_a=p_a\lVert u_a-\bar u\rVert^2$
with the probability-only profile $(r/V)p_a$ in a common 27,450-outcome conditional representation.
Fourteen models covered two sizes each of Pythia, GPT-2, GPT-Neo, Qwen2, BLOOM, Mamba and RWKV.
The 192 fresh contexts per model comprised equal Wikipedia, AG News and LAMBADA subsets.
Absolute effective-dimension error was the root-mean-square count error across ten dampings,
$\alpha/(\operatorname{tr}H/r)\in\{30,10,3,1,0.3,0.1,0.03,0.01,0.003,0.0003\}$.
Exact spectral-exponent error was measured over ranks 8--32 on 48 contexts per model.
Within-family medians were computed for each predictor before taking their log ratio.

In family order BLOOM, GPT-2, GPT-Neo, Mamba, Pythia, Qwen2 and RWKV, the weighted/probability-only
log-error ratios for effective dimension are $0.62149$, $0.16456$, $0.49457$, $0.45938$, $0.43075$,
$0.88203$ and $0.52199$. Their equal-family mean is $0.51068$, favouring the probability-only
profile. For spectral shape, the corresponding ratios are $-0.49943$, $-0.40051$, $-0.25089$,
$-0.21088$, $-0.31919$, $-0.67191$ and $-0.29760$; their mean is $-0.37863$, favouring the weighted
profile. Enumerating all 128 family sign patterns gives one-sided $P=1/128$ for each contrast.
The common-outcome comparison separates probability concentration from read-out weighting;
it complements the native-vocabulary external-family prediction above, whose estimator and
outcome representation are retained in main Figure~\ref*{Main-fig:imprint}a,b.

\subsection{Singleton tokens, interchangeable clusters and multiscale inheritance}\label{si:clusters}

Let $\Pi_L$ denote the conditional-expectation projector of a partition $L$ of outcomes in the
weighted coordinate representation, and let $U^\top U=I_r$ be a resolved rank-$r$ frame with explained energy fraction
$q(L)=\operatorname{tr}\bigl(U^\top \Pi_L U\bigr)/r$.
When $\operatorname{rank}(\Pi_LU)=r$, write $P_T=UU^\top$ and let $P_C$ be the orthogonal
projector onto $\operatorname{col}(\Pi_LU)$.

\begin{proposition}[clusters as conditional expectations]\label{prop:si-condexp}
$\Pi_L$ is the orthogonal projector onto partition-measurable directions; maximising the explained
fraction over partitions with $K$ cells is a weighted $K$-means problem on the frame rows. If
$\operatorname{rank}(\Pi_LU)=r$, then
$\lVert P_C-P_T\rVert_F^2=2r\{1-q(L)\}$ links the explained fraction to projector distance.
\end{proposition}

\begin{proof}
Conditional expectation is self-adjoint, idempotent and has as its image exactly the functions
constant on partition cells, so it is the orthogonal projector onto that subspace. Write the rows
of the weighted frame as $x_a=U_{a,:}/\sqrt{\rho_a}$. The isotropy identity
$\sum_a\rho_ax_ax_a^\top=I_r$ gives
\[
 r\{1-q(L)\}=\sum_{C\in L}\sum_{a\in C}\rho_a\norm{x_a-\bar x_C}^2,
 \qquad
 \bar x_C=\frac{\sum_{a\in C}\rho_ax_a}{\sum_{a\in C}\rho_a}.
\]
Thus maximising capture is exactly weighted $K$-means. If the projected frame retains rank $r$,
the projector formula gives
$\operatorname{tr}(P_CP_T)=\operatorname{tr}(U^\top\Pi_LU)=rq(L)$. The principal-angle
identity for two rank-$r$ projectors then gives
$\norm{P_C-P_T}_{\rm F}^2=2r-2\operatorname{tr}(P_CP_T)=2r\{1-q(L)\}$.
\end{proof}

\begin{theorem}[interchangeability from capture]\label{thm:si-interchangeability}
Let $q_{\min}(L)=\lambda_{\min}(U^\top\Pi_LU)$. For any resolved vector
$y=Uz$ and any within-cluster redistribution $d$ satisfying $\Pi_Ld=0$,
\begin{equation}\label{eq:si-redistribution}
 |\langle d,y\rangle|
 \le\norm d\,\sqrt{1-q_{\min}(L)}\,\norm y.
\end{equation}
If $q_{\min}(L)=1$, every resolved observable is exactly invariant to redistribution within
the cells of $L$. The statement concerns the resolved geometry; the full output distribution can
still distinguish tokens inside a cell.
\end{theorem}

\begin{proof}
Since $\Pi_Ld=0$ and $\Pi_L$ is self-adjoint,
$\langle d,y\rangle=\langle d,(I-\Pi_L)y\rangle$. Cauchy--Schwarz gives the first
factor. Moreover
\[
 \norm{(I-\Pi_L)Uz}^2
 =z^\top\{I-U^\top\Pi_LU\}z
 \le\{1-q_{\min}(L)\}\norm z^2.
\]
Because $U$ is orthonormal, $\norm z=\norm y$. This proves
Eq.~\eqref{eq:si-redistribution}; the exact case follows by setting
$q_{\min}=1$.
\end{proof}

The empirical partition has two regimes. On the fixed language-mode battery, weighted
$K$-means achieves mean capture $0.718$ and worst-direction capture $0.252$, both well above the
capture levels specified before measurement. Twenty-six of 128 cells contain fewer than eight
tokens, including six singleton cells. Those cells contain
probability mass $0.3315$, and every such cell contains a token in the top 158 by reference
probability. The high-probability head therefore requires singleton or near-singleton resolution,
while the remaining mass forms genuine interchangeable clusters. Specificity, reference-swap and
outer-half replications recover the same structure.

Algebraically, let $J_{\rm ex}$ and $J_{\rm bk}$ be the coordinate projectors onto the exceptional
singleton block and clustered bulk, respectively. The partition respects this split. Put
$\Pi_g=J_g\Pi_LJ_g$, $A_g=U^\top\Pi_gU$ and
$E_g=U^\top(J_g-\Pi_g)U$ for $g\in\{\mathrm{ex},\mathrm{bk}\}$. These matrices are
positive semidefinite and give the exact identity
\[
 I_r=A_{\rm ex}+A_{\rm bk}+E_{\rm ex}+E_{\rm bk},
\]
where the two $A$ terms are captured information and the two $E$ terms are unresolved residuals.
The exceptional contribution has rank at most the singleton budget. This separates a
probability-declared head from a support-rich bulk and prevents a high total capture from hiding
poor bulk capture.

Fresh-context tests do not support a universal fixed partition: in a held-out test of the candidate
partitions, the proposed within-cell attenuation failed in all six model-halves, with within-to-cross
median-response ratios of $1.22$--$3.11$. A separate cross-model
partition comparison finds bulk agreement $1.0$ against a matched-null median $0.979$ in both
splits, which is descriptive evidence of shared partition structure. The actual-Fisher inheritance
analysis recovers all 176 theorem certificates and the primary inheritance effect, while the
stronger half-sample directionality estimates do not support a directional effect in any arm.
Together these results establish the within-battery singleton-head/cluster-bulk anatomy and the
exact interchangeability theorem rather than a universal fixed partition across contexts.

\section{Acquisition time on a logarithmic scale}\label{si:acquisition}
This section gives the acquisition evidence reported here. It distinguishes the statistical schedule
measured on pretrained-model checkpoint ladders from the causal depth intervention, then connects
both to the geometric motion and margin results below.

For a fact $f$, let $t(f)$ be the first checkpoint after which its signed answer margin remains
above a fixed threshold specified before model evaluation, and define its logarithmic acquisition time
\begin{equation}\label{eq:si-acquisition-clock}
 \tau(f)=\log_2\{t(f)/t_0\}.
\end{equation}
The surviving depth effect is a translation in this logarithmic time coordinate, and alignment by
held-out loss across model sizes is more accurate than alignment by training step.

\subsection{Corpus $n$-gram statistics predict acquisition before training}

Across Pythia-70M, Pythia-160M and Pythia-410M, the margin trajectory of a fact is predicted from
its ex-ante corpus margins at unigram, bigram and trigram resolution. At 70M and 160M, the selected
log-linear model hands weight from coarse to fine statistics during training; at the final
checkpoints the trigram and bigram terms dominate, with the trigram-to-unigram weight ratio ranging
from $2.7$ to $4.9$. Without refitting on fresh facts with zero prefix overlap, the predictor
explains $0.775$--$0.792$ of margin-trajectory variance, attains acquisition-status agreement
$0.792$--$0.854$ and predicts continuous acquisition time with error $0.77$--$0.96$ in $\tau$.
The corresponding errors are $1.04$, $2.20$ and $0.98$ for the constant acquisition-time baseline
and $2.02$, $2.56$ and $2.34$ for the label-permutation baseline. Corpus $n$-gram margins therefore
predict the acquisition schedule before training.

\subsection{Evidence depth causes a multiplicative delay}

The causal experiment assigns the same synthetic facts to five evidence depths while sharing
initial weights and batch streams across arms. Every statistic below the assigned depth has zero
margin and the complete deciding margin appears at and above it. For the fixed lower-tail
quantile band $p\in[0.10,0.20]$,
\begin{equation}\label{eq:si-logtime-law}
 Q_p\{\tau\mid R4\}-Q_p\{\tau\mid R0\}\simeq\Delta,
 \qquad
 \Delta=2.103\ [1.79,2.40].
\end{equation}
Equivalently, deepest evidence delays acquisition by a factor $2^\Delta\simeq4.3$ in training
steps on this band. The fitted shift varies by only $0.167$ in $\tau$ across the analysed quantiles,
small relative to $\Delta=2.103$. The $R4-R2$ difference in $\tau$ is $1.365$ $[1.14,1.59]$,
approximately $2.6$-fold in steps. The near-uniform translation is supported for
$R4-R0$; $R4-R2$ independently establishes the direction and magnitude. The onset of unsigned
target--alternative margin change remains stable: depth delays persistent signed commitment.
Random reassignment of the same facts across depth arms, with shared initial
weights and batch streams, identifies evidence depth as the cause of the delay while holding fact
composition fixed.

\subsection{Generalisation across architectures and evidence constructions}

A separate fully crossed crossover exchanges shallow and deep assignment within matched
reciprocal-fact blocks while holding initial parameters, minibatch streams, exposure counts,
sequence lengths and the global shallow:deep mixture fixed. The design crosses three compact
four-layer decoders---absolute-position/GELU GPT-2, rotary/GELU GPT-NeoX and
RoPE/RMSNorm/SwiGLU Llama---with distant copy-back-reference and four-cue-parity controlled
languages. It uses eight seeds and 384 matched reciprocal-fact blocks per cell, for 96 runs, and
evaluates the persistent 0.75-nat acquisition time on a 100-point checkpoint grid through step
44,800. The crossed bootstrap uses 10,000 resamples of seed and matched block.

For copy-back-reference and four-cue parity respectively, the mean paired differences in $\tau$ are $1.017$
$[0.876,1.161]$ and $1.205$ $[1.060,1.353]$ for GPT-2, $0.996$
$[0.849,1.150]$ and $1.345$ $[1.196,1.497]$ for GPT-NeoX, and $0.810$
$[0.699,0.929]$ and $1.514$ $[1.365,1.669]$ for Llama (95\% intervals; Extended Data
Fig.~\ref*{Main-fig:ed-acq-generalisation}b,c). All six intervals and all 48 seed means are positive. All twelve fixed-treatment and
randomised-label control intervals contain zero. Averaging architectures equally, parity exceeds
copy-back-reference by $0.414$ $[0.291,0.536]$ in $\tau$; the corresponding within-architecture
contrasts are $0.189$ $[-0.015,0.384]$ for GPT-2, $0.349$ $[0.145,0.552]$ for GPT-NeoX and
$0.704$ $[0.500,0.912]$ for Llama. Evidence depth therefore delays acquisition across
architectures, while evidence construction modulates that delay on average. This mean paired difference
in logarithmic acquisition time complements the lower-tail
translation above and leaves its $\Delta=2.103$ value unchanged.

\subsection{Held-out loss aligns cumulative acquisition curves across scale}

Main-text Fig.~\ref*{Main-fig:acq}b,c shows the cumulative number of persistently acquired facts across the 70M, 160M and 410M checkpoint
ladders in both training-step and held-out-loss coordinates. In the quantified 70M-to-160M
comparison, membership sign agreement is $0.92$--$0.93$, and the trajectory of the $n$-gram
predictor aligns approximately three times more accurately by held-out loss than by training step
(gap $0.130$ against $0.413$), a descriptive comparison. On the 300-fact real-model
battery, the $n$-gram predictor reproduces the trajectories independently at all three sizes without refitting; the
observational level schedule, absolute cross-size alignment error under held-out loss and between-seed variability threshold are also
evaluated at 70M and 160M. These real-checkpoint results establish statistical composition but
remain observational. Causal depth identification and effects at matched $n$-gram evidence come from the
randomised synthetic system, and the held-out-loss alignment advantage is descriptive.

The resulting acquisition law is a composition of four measured objects:
\begin{equation}\label{eq:si-acquisition-synthesis}
 \tau(f)\simeq \tau_0\{\text{deciding-level n-gram margin}\}
 +\Delta\{\text{depth}\},
\end{equation}
with the cumulative acquisition curve re-indexed across scale by held-out loss. Equation~\eqref{eq:si-logtime-law} summarises the
near-uniform measured shift over the analysed quantile band;
Eq.~\eqref{eq:si-acquisition-synthesis} is the compact
first-order reading that connects the measured population-level components. Per-fact additive fits
reach $R^2$ of $0.72$--$0.83$ with a measured $n$-gram-margin-by-depth interaction; the direct causal
result is the near-uniform lower-tail quantile shift of Eq.~\eqref{eq:si-logtime-law}.

\section{Projected-margin dynamics and conditional acquisition time}\label{si:evidence-flux}

The persistent acquisition time above is an empirical population summary. This note supplies a
model-relative accounting framework for a fixed weighted margin projection and states the
additional assumptions needed to infer first-passage or logarithmic-time relations from its trajectory.

Let $\mathcal F$ be a finite population of facts, with fixed weights $w_f\ge0$ satisfying
$\sum_fw_f=1$. Let $z_f\in\R^R$ be fixed, response-independent resolved evidence coordinates,
and let $m_f(\theta)$ be the oriented answer margin, so larger values favour the declared answer.
For a parameter path $\theta(t)$ define the weighted margin projection
\begin{equation}\label{eq:si-evidence-front}
 \beta_r(t)=\E_w[z_{fr}m_f(\theta(t))].
\end{equation}
The estimand fixes the fact population, weights and response-independent coordinates and measures
the trajectory of the declared weighted margin projection.

\subsection{Exact continuous and finite-step accounting}

\begin{theorem}[exact rate of change of the weighted margin projection]\label{thm:si-front-flux}
Suppose $\theta$ is differentiable at $t$ and every $m_f$ is differentiable at $\theta(t)$, and put
$q_f(t)=\nabla_\theta m_f(\theta(t))$ and
$\bar q_r(t)=\E_w[z_{fr}q_f(t)]$. Then:
\resultclause{Instantaneous rate}
\begin{equation}\label{eq:si-front-flux}
 \dot\beta_r(t)=\E_w[z_{fr}q_f(t)^\top\dot\theta(t)]
 =\bar q_r(t)^\top\dot\theta(t).
\end{equation}
\resultclause{Integrated change}
If $\theta$ is absolutely continuous on $[t_0,t_1]$ and every $m_f$ is $C^1$ on a
neighbourhood of the parameter path, then $\beta_r$ is absolutely continuous, the
instantaneous identity holds almost everywhere, and
\begin{equation}\label{eq:si-front-integral}
 \beta_r(t_1)-\beta_r(t_0)
 =\int_{t_0}^{t_1}\bar q_r(t)^\top\dot\theta(t)\dd t.
\end{equation}
\end{theorem}

\begin{proof}
The fact set is finite and $z_f,w_f$ are fixed, so differentiation commutes with the weighted
sum. The chain rule gives $\dot m_f=q_f^\top\dot\theta$; finite linearity gives
Eq.~\eqref{eq:si-front-flux}. Under the stated regularity assumptions, the chain rule for
absolutely continuous paths holds almost everywhere, and the fundamental theorem of calculus gives
Eq.~\eqref{eq:si-front-integral}.
\end{proof}

The fixed-coordinate assumption is essential. If $z_f=z_f(t)$, the product rule adds
$\E_w[\dot z_{fr}(t)m_f(t)]$ to Eq.~\eqref{eq:si-front-flux}; a checkpoint-dependent resolver can
therefore create apparent movement of the projection even when the margins do not move.

\begin{theorem}[exact finite-step path accounting]\label{thm:si-front-discrete}
Let $\theta_0,\ldots,\theta_K$ be any finite parameter path,
$\Delta\theta_k=\theta_{k+1}-\theta_k$ and
$\Delta m_{fk}=m_f(\theta_{k+1})-m_f(\theta_k)$. Define the finite-step projected-margin change
$W_{rk}=\E_w[z_{fr}\Delta m_{fk}]$. Then:
\resultclause{Telescope}
\begin{equation}\label{eq:si-front-telescope}
 \beta_r(K)-\beta_r(0)=\sum_{k=0}^{K-1}W_{rk}.
\end{equation}
\resultclause{Arbitrary tangent}
For any supplied covectors $q_{fk}$, let
\begin{equation}\label{eq:si-front-remainder-def}
 R_{fk}=\Delta m_{fk}-q_{fk}^\top\Delta\theta_k,\qquad
 \bar q_{rk}=\E_w[z_{fr}q_{fk}],\qquad
 \bar R_{rk}=\E_w[z_{fr}R_{fk}].
\end{equation}
Then exactly
\begin{equation}\label{eq:si-front-remainder}
 \beta_r(K)-\beta_r(0)=
 \sum_{k=0}^{K-1}\{\bar q_{rk}^\top\Delta\theta_k+\bar R_{rk}\}.
\end{equation}
\resultclause{Zero-remainder criterion}
If every margin is continuously differentiable on
the $k$th update segment and
\(
 \widetilde q_{fk}=\int_0^1\nabla m_f(\theta_k+s\Delta\theta_k)\dd s,
\)
then
\begin{equation}\label{eq:si-front-orthogonality}
 R_{fk}=0\quad\Longleftrightarrow\quad
 (\widetilde q_{fk}-q_{fk})^\top\Delta\theta_k=0.
\end{equation}
Thus $q_{fk}=\widetilde q_{fk}$ is sufficient but not necessary: one update identifies only the
covector projection along its actual displacement.
\end{theorem}

\begin{proof}
Equation~\eqref{eq:si-front-telescope} follows by summing
$\beta_r(k+1)-\beta_r(k)$. Substituting
$\Delta m_{fk}=q_{fk}^\top\Delta\theta_k+R_{fk}$ and using finite linearity gives
Eq.~\eqref{eq:si-front-remainder}. Along the update segment, the chain rule and fundamental
theorem of calculus give
$\Delta m_{fk}=\widetilde q_{fk}^\top\Delta\theta_k$; subtracting the supplied linear term gives
Eq.~\eqref{eq:si-front-orthogonality}.
\end{proof}

The endpoint telescope requires no smoothness and is the primary exact finite-step identity.
Using a left-endpoint or fixed-initialisation tangent is also exact only when its remainder is
retained. A zero pooled remainder can conceal cancellation among nonzero fact-level remainders.

\subsection{Signed rate in a positive-semidefinite geometry}

Fix a mode and time, abbreviate $q=\bar q_r$ and $y=\dot\theta$ (or
$y=\Delta\theta_k$), and let $A\succeq0$ be a declared local metric or effective optimiser
geometry. Supply solve certificates
\begin{equation}\label{eq:si-front-solves}
 Ax=q,\qquad Ay=g,
\end{equation}
where $g$ is the signed update covector whose metric solve is the actual velocity. Define
\begin{equation}\label{eq:si-front-factors}
 S=q^\top x,\qquad E=g^\top y=y^\top Ay,\qquad
 V=q^\top y,\qquad a=\frac{V}{\sqrt S\sqrt E}.
\end{equation}

\begin{theorem}[positive-semidefinite projected-margin rate factorisation]
\label{thm:si-front-factorisation}
If Eq.~\eqref{eq:si-front-solves} holds and $S,E>0$, then
\begin{equation}\label{eq:si-front-factorisation}
 V=\sqrt S\sqrt E\,a,\qquad |a|\le1,\qquad V^2\le SE.
\end{equation}
The sign of $a$ records whether the update increases or decreases the weighted margin projection.
For singular $A$, the solve certificates require $q,g\in\operatorname{Range}(A)$; null-space
choices do not alter $S,E$ or $V$. If $S=0$ or $E=0$, then $V=0$ but the normalised alignment is
undefined.
\end{theorem}

\begin{proof}
Because $q=Ax$ and $g=Ay$, one has $V=x^\top Ay$, $S=x^\top Ax$ and $E=y^\top Ay$.
Cauchy--Schwarz for the positive-semidefinite form gives $V^2\le SE$; the signed factorisation
and unit bound follow from the definition of $a$.
\end{proof}

\subsection{Conditional first-passage and logarithmic-time relations}

Let two projected-margin trajectories start from the same level $\beta_0$, and put
$C_j(k)=\beta_j(k)-\beta_0$ for $j\in\{s,d\}$.

\begin{theorem}[cumulative-change first-passage ordering]\label{thm:si-front-passage}
Fix a common threshold $h$.
\resultclause{Exact dominance}
If the deep path first reaches $h$ at index $k$ and
$C_s(j)\ge C_d(j)$ for $0\le j\le k$, then the shallow path has reached $h$ by index $k$.
No monotonicity of either path is required.
\resultclause{Uniform-error form}
If $\varepsilon\ge0$ and
$C_d(j)\le C_s(j)+\varepsilon$ through $k$, then a deep first crossing of
$h+\varepsilon$ implies a shallow crossing of $h$ by $k$.
The same pointwise argument gives inclusion of continuous-time hit sets and hence orders their
infima whenever the deep hit set is nonempty.
\end{theorem}

\begin{proof}
At the deep crossing, exact dominance gives
$\beta_s(k)=\beta_0+C_s(k)\ge\beta_0+C_d(k)=\beta_d(k)\ge h$.
Under uniform error,
$\beta_s(k)\ge\beta_d(k)-\varepsilon\ge h$. Earlier or later increments are irrelevant.
\end{proof}

This theorem concerns ordinary first entry. The empirical acquisition time in
Eq.~\eqref{eq:si-acquisition-clock} requires the margin to remain above threshold at all later
observed checkpoints. Ordering that persistent acquisition time requires an additional persistence
argument; sparse local rate measurements alone do not establish cumulative-change dominance.

\begin{theorem}[conditional exponential response]\label{thm:si-front-exponential}
Suppose a projected-margin trajectory has the normalised response
\[
 \beta_r(t)=\beta_{r,\infty}
 -(\beta_{r,\infty}-\beta_{r,0})e^{-\kappa_rt},\qquad \kappa_r>0,
\]
with $\beta_{r,\infty}>\beta_{r,0}$. At fractional threshold $0<\rho<1$,
\begin{equation}\label{eq:si-front-hit-time}
 T_r(\rho)=\frac{-\log(1-\rho)}{\kappa_r}.
\end{equation}
Consequently $\kappa_s\ge\kappa_d>0$ implies $T_s(\rho)\le T_d(\rho)$, and
\begin{equation}\label{eq:si-front-log-translation}
 \log_2T_d(\rho)-\log_2T_s(\rho)=\log_2\frac{\kappa_s}{\kappa_d}.
\end{equation}
The threshold-independent translation follows from the shared normalised response shape and
constant rates; it is not implied by the rate identity alone.
\end{theorem}

\begin{proof}
Solving $1-e^{-\kappa_rt}=\rho$ gives Eq.~\eqref{eq:si-front-hit-time}. Positivity of
$-\log(1-\rho)$ reverses the rate ordering into a time ordering. Taking the time ratio and its
logarithm gives Eq.~\eqref{eq:si-front-log-translation}.
\end{proof}

\begin{corollary}[distributional scaling]\label{si:c:front-scale}
Let $T_s,T_d$ be nonnegative random acquisition times. If
$T_d\overset{d}=cT_s$ for $c>0$, equivalently
$F_d(t)=F_s(t/c)$ for all $t$, then every lower quantile
$Q_j(p)=\inf\{t:F_j(t)\ge p\}$ satisfies
\begin{equation}\label{eq:si-front-quantile}
 Q_d(p)=cQ_s(p),\qquad 0<p<1.
\end{equation}
For positive quantiles this is a constant logarithmic-time translation by $\log_2c$.
\end{corollary}

\begin{proof}
Multiplication by $c>0$ preserves order, so scaling maps the entire CDF level set for the shallow
condition onto the corresponding level set for the deep condition. Its infimum is therefore
multiplied by $c$.
\end{proof}

Population-level scaling need not be a same-fact multiplier. Conversely, a shift over one
quantile band does not by itself establish full distributional scaling.

\begin{theorem}[learner-independent depth monotonicity is impossible]
\label{thm:si-front-depth-counterexample}
Consider the positive-rate cumulative-change model $T_j=W_j/\kappa_j$, with required change
$W_j>0$ and effective rate $\kappa_j>0$. Within this model there exist a fixed shallow/deep
evidence pair with nominal depths $d_s<d_d$ and two positive rate profiles whose acquisition orders are opposite.
Thus nominal depth alone cannot order acquisition time throughout this model without an
assumption on learner-relative effective rates or an equivalent accessibility object.
\end{theorem}

\begin{proof}
Give both items one unit of required change. The rate profile
$(\kappa_s,\kappa_d)=(2,1)$ has $(T_s,T_d)=(1/2,1)$, whereas the profile $(1,2)$ has
$(T_s,T_d)=(1,1/2)$. The evidence depths are unchanged and both rate profiles are positive.
\end{proof}

Randomised evidence-depth assignment identifies the causal acquisition effect in Supplementary
Note~\ref{si:acquisition}; the weighted-margin-projection identities provide conditional path accounting for
the declared evidence projection.

\section{Geometric motion, margins and local objective rate}\label{si:motion}

\begin{theorem}[continuous scores and discontinuous decisions]\label{thm:si-motion-budget}
Let $\rho(p,q)=\norm{\sqrt p-\sqrt q}_2$.
\resultclause{Continuous score}
If a score $S$ is $L$-Lipschitz in this metric, then
\begin{equation}\label{eq:si-score-speed}
 |S(p)-S(q)|\le L\rho(p,q).
\end{equation}
\resultclause{Discrete decision}
For a discrete argmax decision, let $y$ be the unique maximiser of $p$ and
$m=p_y-\max_{j\ne y}p_j$ its margin. If the maximiser changes under $q$, then
\begin{equation}\label{eq:si-flip-margin}
 \operatorname{TV}(p,q)\ge m/2,
 \qquad \rho(p,q)\ge m/2.
\end{equation}
\end{theorem}

\begin{proof}
Equation~\eqref{eq:si-score-speed} is the definition of Lipschitz continuity. If $z\ne y$
maximises $q$, then $q_z\ge q_y$ and
\[
 m\le p_y-p_z
 \le |p_y-q_y|+|p_z-q_z|
 \le\norm{p-q}_1=2\operatorname{TV}(p,q).
\]
Finally $\operatorname{TV}(p,q)\le\rho(p,q)$ follows from
$\norm{p-q}_1\le2\norm{\sqrt p-\sqrt q}_2$. This proves
Eq.~\eqref{eq:si-flip-margin}.
\end{proof}

The continuous bound is checked without violation in 5{,}704 implementation cases. The discrete
margin theorem has zero violations over 45{,}315 observed cross-seed flips, or 90{,}630 oriented
checks. On the held-out LAMBADA battery~\cite{lambada} the flip frequency falls from $0.519$ to zero as the
normalised margin $z=m/(2\rho)$ crosses one, where zero is forced by the theorem. This connects a
smooth geometric acquisition trajectory to an apparently abrupt benchmark transition.

\begin{theorem}[Cauchy--Schwarz bound for the objective rate]\label{thm:si-speed-factorisation}
Let $A$ be positive definite, let $q$ be a behaviour covector and $g$ a training-loss covector,
and solve $Ax=q$, $Ay=g$. Define
\[
 \text{sensitivity}=q^\top x,\qquad
 \text{update norm}^2=g^\top y,\qquad
 \text{objective rate}=q^\top y.
\]
Then
\begin{equation}\label{eq:si-speed-bound}
 \text{objective rate}^2\le \text{sensitivity}\times\text{update norm}^2.
\end{equation}
When both quadratic forms are positive, define
$\text{alignment}=\langle x,y\rangle_A/(\norm x_A\norm y_A)$. Then exactly
\begin{equation}\label{eq:si-speed-exact}
 \text{objective rate}^2=
 \text{sensitivity}\times\text{update norm}^2\times\text{alignment}^2.
\end{equation}
Equivalently, $\text{objective rate}^2/\text{sensitivity}\le\text{update norm}^2$.
\end{theorem}

\begin{proof}
Use the $A$-inner product $\langle u,v\rangle_A=u^\top Av$. Then
$\text{sensitivity}=\norm x_A^2$, $\text{update norm}^2=\norm y_A^2$ and
$\text{objective rate}=q^\top y=x^\top Ay=\langle x,y\rangle_A$. Cauchy--Schwarz proves
Eq.~\eqref{eq:si-speed-bound}; dividing its two sides by the product of the positive
quadratic forms and using the definition of alignment gives Eq.~\eqref{eq:si-speed-exact}.
\end{proof}

Objective sensitivity alone therefore cannot determine acquisition rate. With $A=I$, $q=(1,0)$ and unit
training covectors $(1,0)$ and $(0,1)$, sensitivity and squared update norm are identical but the objective rate is one
or zero. On 80 held-out Pythia-70M examples, evaluated with a declared final-read-out
preconditioner, mean alignment squared is $0.6811$ (95\% interval $0.6503$--$0.7117$). The
factorisation holds exactly and the correlation between log sensitivity and log squared objective rate is $0.541$.
The missing quantity linking controllability to learning is objective--update alignment.

\begin{proposition}[read-out subspace kinematics]\label{prop:si-kinematics}
Let $Q(t)$ have orthonormal columns and set $P(t)=Q(t)Q(t)^\top$ and
$B=(I-P)\dot Q$. Then
\begin{equation}\label{eq:si-projector-kinematics}
 \dot P=BQ^\top+QB^\top,\qquad Q^\top B=0,\qquad
 \norm{\dot P}_{\rm F}^2=2\norm B_{\rm F}^2.
\end{equation}
Only the normal component $B$ rotates the subspace. In particular, an update confined to the
current column span, including scalar decoupled weight decay, produces no rotation.
\end{proposition}

\begin{proof}
Differentiate $P=QQ^\top$. Differentiating $Q^\top Q=I$ shows that
$Q^\top\dot Q$ is skew-symmetric, so its two contributions to $\dot P$ cancel. The remaining
normal component is $B$, giving the first identity and $Q^\top B=0$. Orthogonality makes the two
summands Frobenius-orthogonal and equal in norm, which proves the final identity. If
$\dot Q$ is in the span of $Q$, then $B=0$.
\end{proof}

The same approximation residual that governs static recovery also controls how strongly unresolved
language can rotate the read-out subspace.

\begin{theorem}[language residual bounds read-out rotation]\label{thm:si-language-force}
Fix a strictly positive reference law $\rho$ and put
\[
 P_\rho=I-\sqrt\rho\sqrt\rho^{\,\top},\qquad
 C_\rho=P_\rho\operatorname{diag}(\rho)^{1/2},\qquad
 \kappa_\rho=\norm{C_\rho}_{\rm op}.
\]
For an untied affine read-out $W_t$, let
$A_t=C_\rho W_t$, $R_t=A_t^\top A_t$, $U_t=A_tR_t^{-1/2}$ and
$P_t=U_tU_t^\top$ on a constant-full-rank interval. Let the true next-token law be
$\widetilde\pi_X$, let $\pi_X$ be a declared resolved core, and define
\[
 \zeta=\E_X\norm{\sqrt{\widetilde\pi_X}-\sqrt{\pi_X}}^2,
 \qquad
 \psi_t(X)=R_t^{-1/2}h_t(X),\qquad
 \Psi_t=\{\E\norm{\psi_t(X)}^2\}^{1/2}.
\]
Under the population read-out flow
\[
 \dot W_t=\gamma_t\E[(e_Y-p_t)h_t(X)^\top]-\lambda_tW_t+\mathcal D_t,
 \qquad Y\mid X\sim\widetilde\pi_X,\qquad \gamma_t\ge0,
\]
the horizontal velocity $B_t=(I-P_t)\dot A_tR_t^{-1/2}$ decomposes exactly as
\begin{equation}\label{eq:si-force-decomposition}
 B_t=B_t^{\rm core}+E_t^{\rm lang}+E_t^{\rm mech},
\end{equation}
where
\begin{align*}
 B_t^{\rm core}&=\gamma_t(I-P_t)C_\rho
   \E[(\pi_X-p_t)\psi_t(X)^\top],\\
 E_t^{\rm lang}&=\gamma_t(I-P_t)C_\rho
   \E[(\widetilde\pi_X-\pi_X)\psi_t(X)^\top],\\
 E_t^{\rm mech}&=(I-P_t)C_\rho\mathcal D_tR_t^{-1/2}.
\end{align*}
Scalar weight decay contributes exactly zero, and
\begin{equation}\label{eq:si-language-force}
 \norm{E_t^{\rm lang}}_{\rm F}
 \le2\gamma_t\kappa_\rho\Psi_t\sqrt\zeta,
 \qquad
 \norm{\dot P_t^{\rm actual}-\dot P_t^{\rm core+mech}}_{\rm F}
 \le2\sqrt2\gamma_t\kappa_\rho\Psi_t\sqrt\zeta.
\end{equation}
\end{theorem}

\begin{proof}
Conditioning on $X$ gives $\E[e_Y\mid X]=\widetilde\pi_X$. Apply $C_\rho$ to the
flow, substitute into $B_t$, add and subtract $\pi_X$, and use
$(I-P_t)A_t=0$. This proves Eq.~\eqref{eq:si-force-decomposition}, including the
zero contribution of $-\lambda_tA_t$. For probability vectors,
\[
 \norm{p-q}_2\le2\norm{\sqrt p-\sqrt q}_2,
\]
because $|p_i-q_i|=|\sqrt p_i-\sqrt q_i|(\sqrt p_i+\sqrt q_i)$ and the final
factor is at most two. Projection is contractive, so Frobenius Cauchy--Schwarz gives
\[
 \norm{E_t^{\rm lang}}_{\rm F}
 \le\gamma_t\kappa_\rho
 \{\E\norm{\widetilde\pi_X-\pi_X}_2^2\}^{1/2}\Psi_t
 \le2\gamma_t\kappa_\rho\Psi_t\sqrt\zeta.
\]
Finally, Proposition~\ref{prop:si-kinematics} applied to the velocity difference
$E_t^{\rm lang}$ multiplies its Frobenius norm by $\sqrt2$, proving the projector bound.
\end{proof}

A controlled smooth-path experiment supports the projector identity, the scalar-decay null and
Eq.~\eqref{eq:si-language-force} in all 24 held-out seeds. The corpus $n$-gram, evidence-depth and
held-out-loss alignment experiments establish the acquisition-time results above.

\paragraph{Cyclic concepts and relation-induced probability displacements.}
For a cyclic registry (weekdays, months), the advance-one-step map is fitted as the metric
isometry (rotation) in the Fourier-1 phase representation that best carries each member's distribution to the
next. The rotation angle is compared with the ideal $2\pi/n$ (weekdays $0.902$ vs $2\pi/7 = 0.898$).
The off-rotation component is the metric-isometry defect reported in the double dissociation
($0.29$ for cyclic concepts, $1.02$ for relations), and the closure defect is the Fisher--Rao
distance between the start point and the composition of $n$ steps.

Relations (capital-of,
gender, tense, comparative) are scored by probability-displacement alignment, the mean cosine alignment of the
probability-displacement directions across instances (raw $0.49$ for relations, near zero for
cyclic), reported against a permuted-geometry control that preserves marginals ($0.33$).

\sipart{IV}{Controllable}

\section{Minimum-disturbance control and coordinate dependence of Euclidean cost}\label{si:minimal-control}
This section collects consequences of the chain that the operations of the
main text use directly.

Let $A$ be a positive-definite local cost metric, $q\ne0$ a behaviour covector and
$b\ne0$ a desired first-order change, so feasible interventions satisfy $q^\top\delta=b$.
For the undamped output problem $A=G$ on the resolved tangent space; for the stable
finite-resolution problem $A=G+\alpha R$, where $R$ is the declared reference metric.

\begin{theorem}[minimum disturbance and the exact relative cost of Euclidean control]
\label{thm:si-minimum-disturbance}
\resultclause{Minimum}
The unique minimum-cost intervention is
\begin{equation}\label{eq:si-natural-step}
 \delta_{\rm N}=\frac{bA^{-1}q}{q^\top A^{-1}q},
 \qquad
 \frac12\delta_{\rm N}^\top A\delta_{\rm N}
 =\frac{b^2}{2q^\top A^{-1}q}.
\end{equation}
\resultclause{Exact gap}
Every feasible $\delta$ obeys the exact gap identity
\begin{equation}\label{eq:si-offtarget-gap}
 \frac12\delta^\top A\delta-\frac{b^2}{2q^\top A^{-1}q}
 =\frac12\norm{\delta-\delta_{\rm N}}_A^2.
\end{equation}
\resultclause{Euclidean-to-natural cost ratio}
In particular, the Euclidean step $\delta_{\rm E}=bq/(q^\top q)$ has cost ratio
\begin{equation}\label{eq:si-kantorovich-price}
 \Gamma_A(q)
 :=\frac{\delta_{\rm E}^\top A\delta_{\rm E}}
          {\delta_{\rm N}^\top A\delta_{\rm N}}
 =\frac{(q^\top Aq)(q^\top A^{-1}q)}{(q^\top q)^2}\ge1.
\end{equation}
\resultclause{Spectral bound}
If the spectrum of $A$ lies in $[m,M]$, then
\begin{equation}\label{eq:si-kantorovich-bound}
 1\le\Gamma_A(q)\le\frac{(M+m)^2}{4Mm}.
\end{equation}
Equality on the left holds precisely when $q$ is supported on one eigenspace of $A$.
\resultclause{Damping}
For $B=B^\top\succeq0$, $\alpha>0$ and $A_\alpha=B+\alpha I$,
$\Gamma_{A_\alpha}(q)$ and its worst-case ceiling are
non-increasing in $\alpha$ and converge to one: damping removes the exploitable
anisotropy together with the natural-gradient advantage.
\end{theorem}

\begin{proof}
Lagrange stationarity for minimising $\tfrac12\delta^\top A\delta$ subject to
$q^\top\delta=b$ gives $A\delta=\lambda q$; imposing the constraint gives
Eq.~\eqref{eq:si-natural-step}. For any other feasible $\delta=\delta_{\rm N}+e$,
$q^\top e=0$ and hence
$\delta_{\rm N}^\top Ae=bq^\top e/(q^\top A^{-1}q)=0$. Expanding the quadratic proves
Eq.~\eqref{eq:si-offtarget-gap}; positive definiteness gives uniqueness. Substituting
$\delta_{\rm E}$ and $\delta_{\rm N}$ proves Eq.~\eqref{eq:si-kantorovich-price}.

For the bounds, diagonalise $A$ and put
$a_i=q_i^2/(q^\top q)$, so $\sum_i a_i=1$. Then
\[
 \Gamma_A(q)=\left(\sum_i a_i\lambda_i\right)
 \left(\sum_i\frac{a_i}{\lambda_i}\right).
\]
Cauchy--Schwarz gives
the lower bound. Moreover $(\lambda_i-m)(M-\lambda_i)\ge0$ implies
$\lambda_i+mM/\lambda_i\le M+m$. Averaging and applying the arithmetic--geometric
mean inequality gives
\[
 2\sqrt{mM\Gamma_A(q)}
 \le \sum_i a_i\lambda_i+mM\sum_i\frac{a_i}{\lambda_i}
 \le M+m,
\]
which proves Eq.~\eqref{eq:si-kantorovich-bound}. Equality in the lower
Cauchy--Schwarz bound requires a common eigenvalue wherever $a_i>0$.

Finally, with $x_i=\lambda_i(B)+\alpha$ and weighted averages denoted by $\E_a$,
\[
 \frac{\mathrm d}{\mathrm d\alpha}\Gamma_{A_\alpha}(q)
 =\E_a(x^{-1})-\E_a(x)\E_a(x^{-2})\le0,
\]
because $x$ and $x^{-2}$ are oppositely ordered, so
$\E_a(x\,x^{-2})\le\E_a(x)\E_a(x^{-2})$. The spectral interval is
$[\lambda_{\min}(B)+\alpha,\lambda_{\max}(B)+\alpha]$, whose ratio and
Kantorovich ceiling decrease to one.
\end{proof}

Equation~\eqref{eq:si-offtarget-gap} is the local off-target decomposition used in the
experiments: the first term is total regularised disturbance at matched objective change,
the second is the unavoidable cost of that change, and their difference is exactly half the squared
$A$-distance from the natural step. On a positive-semidefinite $G$, the same statement holds on its
resolved range with $G^{-1}$ replaced by $G^\dagger$, provided $q\in\operatorname{Range}(G)$
and $q\ne0$.

At finite damping, the empirical battery measures realised output KL, whereas
Eq.~\eqref{eq:si-kantorovich-price} compares regularised costs in $A$. The comparison therefore holds
$A=G+\alpha I$ fixed and, for $\Pi_G(u)=(u^\top Gu)/(q^\top u)^2$, predicts the local
Euclidean-to-damped-natural-gradient $G$-cost ratio with
\[
 R_{\rm pred}(q)=\frac{\Pi_G(q)}{\Pi_G(A^{-1}q)}.
\]
The two ratios agree in the undamped limit and in special finite-damping cases, such as an objective
supported on one eigenspace; they generally differ at non-zero damping. The held-out evaluation
covers 1,188 predictor cells at three matched target magnitudes, giving 3,564 deployed responses,
of which 3,515 are usable (98.6\% coverage). Across the usable responses, the median ratio of
measured to predicted advantage is $0.967$ (95\% interval $0.957$--$0.975$), and the log--log
slope of measured on predicted advantage is $0.982$ (95\% interval $0.966$--$0.997$).
At intervention fractions $\Delta C/C=0.01,0.03,0.10$, the model-balanced geometric means of the
measured-to-predicted ratio are $0.918$, $0.860$ and $0.755$, respectively; every one of the 11
within-model paths decreases at both adjacent steps. This quantifies the expected attenuation of a
local prediction as intervention magnitude grows.
The objective-resolved metric anisotropy therefore quantitatively predicts intervention cost across
the cross-model battery; a scalar condition number alone does not supply this prediction.

\begin{proposition}[chart-covariant control]\label{prop:si-chart-control}
Under an invertible chart change $h'=Sh$, let
\[
 A'=S^{-\top}AS^{-1},\qquad q'=S^{-\top}q,\qquad \delta'=S\delta.
\]
Then objective change, cost and the natural step are invariant:
$q'^\top\delta'=q^\top\delta$, $\delta'^\top A'\delta'=\delta^\top A\delta$,
$\delta'_{\rm N}=S\delta_{\rm N}$. The cost ratio of any transported comparator
$\delta'_{\rm C}=S\delta_{\rm C}$ to the natural step is likewise invariant. By contrast, the
fresh coordinate-Euclidean comparator $q'/(q'^\top q')$ is generally not the transport of
$q/(q^\top q)$, so $\Gamma_{A'}(q')$ need not equal $\Gamma_A(q)$: coordinate-Euclidean control
depends on the chosen chart. If
$A=G+\alpha R$, covariance requires $R'=S^{-\top}RS^{-1}$. Resetting the reference to
a fresh identity in the new chart instead maps back to the native reference $S^\top S$.
\end{proposition}

\begin{proof}
The first two identities follow by substitution. Since
$A'^{-1}=SA^{-1}S^\top$,
$A'^{-1}q'=SA^{-1}q$ and $q'^\top A'^{-1}q'=q^\top A^{-1}q$, proving covariance
of Eq.~\eqref{eq:si-natural-step}. The cost ratio is the ratio of two invariant
quadratic costs only when both interventions are transported; the coordinate-Euclidean vectors
do not obey that transport law in general. Finally, a fresh new-chart identity
contributes $\alpha I$ to $A'$; multiplying the equation by $S^\top$ and writing
$\delta'=S\delta$ gives the native term $\alpha S^\top S\delta$.
\end{proof}

\subsection{Relation to direct preference optimisation}\label{si:dpo}
A natural policy gradient for a KL-regularised reinforcement-learning-from-human-feedback (RLHF)
objective has the form
$F^{-1}\nabla_\theta(\text{reward})$, where $F$ is the policy Fisher matrix. Direct preference
optimisation (DPO)~\cite{dpo} parameterises an implicit reward through
$\beta\log[\pi/\pi_{\mathrm{ref}}]$ and optimises model parameters. This inference-time construction
instead uses the activation-space pullback metric and the preference covector,
$\delta h=(G+\alpha I)^{-1}\nabla_h[\log p(y_w\mid x)-\log p(y_l\mid x)]$. The resulting
activation-space natural-gradient update is distinct from DPO's parameter update.

\subsection{Mass-preserving Fisher coarse-graining and interpretable partitions}\label{si:coarse}
Grouping the support tokens into $K$ cells and replacing each token's read-out row by its group
mean gives a coarse output Fisher $H_{\text{coarse}}=\mathrm{Cov}_g(\bar w_g)$, the between-group
covariance of the group-mean unembeddings, which by the law of total covariance equals $H$ minus the
within-group term and is exact when $K$ reaches the support size. For the exponential-tilt family
defined by a scalar token value $\phi$, a partition preserves all Fisher information about the tilt
when $\phi$ is constant within each cell. Approximating the complete steering update also requires
retaining read-out covariance. Bins of $\phi$ are therefore refined by probability-weighted clusters
of the read-out directions $w_a$ and evaluate the resulting update against the ungrouped solve. In
the measured cost model, the exact matrix-free solve is less expensive per step than the tested
partitions fine enough to reproduce its update (main-text Methods).

For sentiment, the resulting interpretable cells correspond
approximately to sentiment level crossed with lexical role. This partition serves interpretation
rather than scalability: at the same update fidelity, the tested token groupings require several times more
cells than the fixed-tolerance Krylov solve requires iterations.

\subsection{The full cubic correction and the erosion predictor}\label{si:cubic}
Let $\eta(h)$ be the output natural parameters and take a straight activation path
$h(t)=h_0+t\delta h$. Write $a=D\eta[\delta h]$, $b=D^2\eta[\delta h,\delta h]$ and
$C=\operatorname{diag}(p)-pp^\top$. Direct expansion of output KL gives
\begin{equation}
 \mathrm{KL}(p_0\Vert p_t)
 =\tfrac12t^2 a^\top C a
 +\tfrac16t^3\!\left\{T(a,a,a)+3a^\top Cb\right\}+O(t^4),
\end{equation}
where $T(a,a,a)=\mathbb{E}_p[(a-\mathbb{E}_pa)^3]$ is the Amari--Chentsov contraction.
The first cubic term is the third central moment pulled back through the Jacobian. The second is
the directional Hessian of the network above the intervention; it vanishes at an affine read-out
and is generally non-zero at an internal layer. The erosion analysis computes the first component
exactly from one Jacobian--vector product and one matrix multiplication; restricted to the
moderate-intervention regime, in which matched targets are reached along monotone segments of the
scan, its rank correlation with gentle-to-aggressive erosion is $+0.49$ to $+0.75$ across
410M--2.8B, and the association reverses sign beyond that regime. In a 410M full-curvature analysis (four prompts,
layers 4, 12 and 20 plus affine final-layer anchors), adding the network-Hessian term reduces mean
absolute error against a parity-cancelling directional finite-difference estimate from 0.345 to
0.0176, a ratio of 0.051; the affine anchors return the map-Hessian term exactly zero. The full
scale-ladder erosion analysis therefore reports the Amari--Chentsov and network-Hessian components
separately and jointly.

\paragraph{Scan and regime comparisons.}
The analysis uses a two-segment scan grid extending the intervention range
sixteenfold (so that unreached matched targets become genuine exclusions after scan-boundary
artefacts are removed), a fixed sign-consistent crossing bracket, and average-tie rank correlations.

The association reverses on the pooled reach-extended population. Supplementary Note~\ref{si:minimal-control} analyses
the moderate-intervention regime, comprising items whose crossings lie on monotone scan segments,
with a partial rank correlation controlling the
gentle-regime advantage separating the predictive signal from its mechanical coupling to the
outcome's denominator. On the 410M subset reaching targets within the narrower scan range, the
correlation is $+0.74$.

\subsection{Relinearised finite intervention paths}
Both Fisher and Euclidean paths were relinearised after every accepted step, with an exact
prompt-KL trust region of $0.02$ nats and common cumulative objective checkpoints of $0.05$, $0.1$,
$0.2$ and $0.4$ nats. Expansion and bisection matched the reached changes to within 1\%.
Cumulative local cost is the sum of $\mathrm{KL}(p_{j-1}\Vert p_j)$ over accepted steps; it measures
the path and is distinct from endpoint or unrelated-prompt disturbance.
The completed panel contains 216 prompts in 18 model--objective cells (16 models from six
families). All 864 prompt--target pairs reach the common checkpoints. Cell--target mean
log Euclidean/Fisher path-cost ratios range from $2.4403$ to $4.8452$, giving geometric-mean
ratios $11.48$--$127.12$. All 72 paired 95\% prompt-bootstrap intervals are positive in log space.
The full set of ratios is displayed in main Figure~\ref*{Main-fig:price}b.

\subsection{Reusable interventions with a reference metric}
BLOOM-560M, Pythia-410M, Pythia-1.4B and Qwen2-1.5B were each given four donor prompts per
objective, eight unseen target prompts per objective and four reference prompt--continuation
pairs. Truth preference and anti-sycophancy used continuation log-odds objectives. A single
activation update per model and objective was shared across prompts, injected at the final
prompt position after block $\operatorname{round}(0.45L)$ for $L$ blocks (zero-based index). The mean donor objective
gradient was preconditioned by the reference-prompt Fisher, by the donor-prompt Fisher, or left
Euclidean. All arms recomputed their gradients and metrics after accepted steps and reached
matched donor effects of $0.05$, $0.1$ and $0.2$ nats within 1.5\%. Fisher damping was $0.03$ times
the estimated largest eigenvalue. No target gradient or target-specific coefficient was used.

The primary cost was mean sequence KL over the reference continuations, evaluated by teacher
forcing and summing over continuation positions. Reference prompts also define the metric, so
the readout evaluates continuation disturbance on that reference set. Across all 24
model--objective--checkpoint settings, the Euclidean/reference-Fisher ratio is $2.8986$--$6.2882$;
the donor-Fisher/reference-Fisher ratio is at least $2.129$. All six model-bootstrap intervals
for the Euclidean/reference-Fisher log ratio are positive, with the smallest lower bound
$1.1429$. The smallest reference-Fisher mean objective change on unseen targets is $0.012758$
nats; all 24 means are positive. These are matched-donor comparisons: held-out target effects
are measured rather than matched.

A two-objective solve requests donor increments of $0.05$ nats for truth preference and
anti-sycophancy using the same reference metric. Both mean target effects are positive in each
of the four models and reference-sequence KL is lower than for the corresponding Euclidean
composition. The reusable object is a within-model activation update, applied after construction
to new prompts. The experiment does not require aligned hidden coordinates between models.

\section{Matrix-free computation}\label{si:instrument}

The intervention algorithms avoid materialising the full matrices $H$, $J$ and
$G=J^\top H J$. Two implementations of the native-reference solve are used
$(G+\alpha I)^{-1}q$; both apply the relevant linear operators to vectors, and a general reference
$R$ is handled by whitening or generalised preconditioning.

\begin{proposition}[dimension-free CG bound at relative damping]\label{si:p:cg}
Let $G\succeq0$ with $\lambda_{\max}(G)>0$, set
$\alpha=c\lambda_{\max}(G)$ with $c>0$, and solve
$(G+\alpha I)x=b$ by conjugate gradients. The regularised condition
number satisfies $\kappa\le(1+c)/c$. If $x_0\ne x_*$, then in exact arithmetic
\[
 \frac{\lVert x_t-x_*\rVert_{G+\alpha I}}
      {\lVert x_0-x_*\rVert_{G+\alpha I}}
 \le 2\left(\frac{\sqrt\kappa-1}{\sqrt\kappa+1}\right)^t.
\]
The residual obeys the same bound in the inverse-operator norm and, in Euclidean
relative norm, the right side multiplied by at most $\sqrt\kappa$. Hence fixed
relative damping and tolerance give a worst-case iteration bound independent of
width. At $c=10^{-2}$, the standard bound gives at most $85$ iterations for Euclidean
relative residual $10^{-6}$.
\end{proposition}

\begin{proof}
The eigenvalues of $G+\alpha I$ lie in
$[\alpha,\lambda_{\max}(G)+\alpha]$. The displayed inequality is the standard
Chebyshev convergence bound for conjugate gradients. Since
$r_t=(G+\alpha I)(x_*-x_t)$, its inverse-operator norm equals the error norm;
Euclidean norm equivalence costs at most $\sqrt\kappa$.
\end{proof}

\emph{Low-rank Woodbury.} Restricting the output Fisher to the top-$|S|$ tokens by mass and
renormalising their probabilities defines a truncated-support approximation $H_S=BB^\top$, factored
exactly as $B=W_U[S,:]^\top\,(\mathrm{diag}(p_S)-p_Sp_S^\top)^{1/2}\in\mathbb{R}^{d\times|S|}$.
Forming $M=J^\top B$ costs $|S|$ vector--Jacobian products (one per column of $B$), after which
$G_S=MM^\top$ and the damped solve reduces, by the Woodbury identity, to an $|S|\times|S|$ system:
$(G_S+\alpha I)^{-1}q=\alpha^{-1}\!\big(q-M(\alpha I+M^\top M)^{-1}M^\top q\big)$.

\emph{Exact conjugate gradients.} For the full vocabulary, $Gv$ is computed in two passes: a
Jacobian--vector product (JVP) returns $Jv$, the exact output-Fisher product
$H(Jv)=W_U^\top(\mathrm{diag}(p)-pp^\top)W_U(Jv)$ is applied in closed form, and a
vector--Jacobian product (VJP) returns $J^\top(\cdot)$. Conjugate gradients on $G+\alpha I$ then
solves the native-reference system. At fixed tolerance, iteration count measures convergence of
the solver at that resolution.
Both routes use explicit automatic-differentiation primitives (reverse-mode VJP via
\texttt{autograd.grad} and forward-mode JVP via double backward).

\begin{enumerate}
\item Power-iterate a few steps to estimate $\lambda_{\max}(G)$; set $\alpha=c\,\lambda_{\max}$, $c\!\sim\!10^{-2}$, fixing damping relative to the spectrum.
\item Solve $(G+\alpha I)^{-1}q$ by Woodbury (low-rank) or conjugate gradients (CG; exact).
\item Range-restrict via Lanczos: project onto the eigenspace retained by numerical spectral truncation, so the step carries no null-space (output-inert) component.
\end{enumerate}

For the aggregated training Fisher $F=\sum_x J_x^\top H_x J_x$, the additive $F$-vector product is accumulated in chunks of examples, holding only one example's autograd graph in memory at a time. All solves run in float32: $G$ is severely ill-conditioned ($\kappa\sim10^{5}$--$10^{7}$), and in the evaluated Pythia-410M and Pythia-1.4B cases the bfloat16 and float32 recovered directions were only weakly to moderately aligned (absolute cosines $0.056$--$0.542$ across two contexts per model).

\paragraph{Dense spectral computation.}
At the evaluated widths ($d \le 1024$), forming the truncated operators directly and diagonalising
them is faster than the matrix-free route for spectra. The matrix-free implementation is retained
for its asymptotic storage scaling: additional operator storage grows linearly with activation
dimension and omits a vocabulary-by-activation Jacobian, while dense-operator storage grows
quadratically. Output evaluation itself still scales with vocabulary size.

\sipart{V}{Evidence, translation and reproducibility}

\section{Experimental protocols}\label{si:protocols}

\subsection{Experimental design and statistical controls}

The empirical analyses separate model-selection or exploratory data from held-out evaluation data
where applicable. For held-out batteries, the population, split, unit of analysis, endpoints and
numerical tolerances are fixed before evaluation. For thresholded endpoints, response-independent
resolution was measured against the instrument floor before evaluation. Reported summaries are
reconstructed from the underlying rows.

The estimand determines the unit of replication. Context batteries are resampled by context;
fact-acquisition studies cross training seed with fact or reciprocal-fact block; intervention
frontiers compare methods on the same prompt and objective sets; dictionary analyses use three paired
seeds. Intervals use the analysis-specific resampling units described in Methods and figure
captions; paired comparisons share each resampling draw across their arms. For cross-model agreement,
factual-relation alignments, corpus mediation, residual eigenvector overlap and band profiles,
95\% percentile intervals use 5,000 full-size context draws with replacement, paired across fixed
model rosters and contrasts. Distance-matrix analyses average tied ranks and exclude comparisons
between repeated copies of the same original context; convergence averages the 29 displayed model
pairs. Separately labelled 80\%-subsample ranges describe stability under context removal.
Nulls retain the nuisance structure relevant to the
claim: permutation within objective--layer strata for the intervention-cost-ratio prediction, displacement-magnitude-matched
rotation for shared eigenspaces, and fixed-half label randomisation for the acquisition controls.

Three batteries deliberately test distinct boundaries of the theory. Identification uses both
local curvature and global profiled-risk certificates. Shared geometry is evaluated on natural,
templated and factual-relation contexts, with a common-outcome subset for the quantitative risk
bounds. Acquisition combines a real-checkpoint forecast applied without refitting, randomised evidence-depth
assignment and cross-size alignment by held-out loss. The components are kept separate in the raw analyses and
combined only in the synthesis of Supplementary Note~\ref{si:acquisition}.

\subsection{Matched-effect comparisons}

Comparisons reported at matched response are made at the same realised change in the scalar objective $\phi$. For
each method and prompt, binary search over the step scale is used until the realised $\Delta\phi$
reaches a target. When the objective changes in discrete increments, the two
steps that bracket the target are interpolated and all other quantities evaluated at that matched
$\Delta\phi$~\cite{fishback,park2026}. The instruction-model frontiers instead evaluate each
prompt-specific direction on a method-specific strength grid shared across the 12 prompts, average
behaviour and off-target KL at each grid point, and compare interpolated frontiers at equal aggregate
mean behaviour change. The amortised contrastive-activation-addition baseline uses the mean covector
direction over the contrast set and the same frontier convention. The six-size Pythia factual sweep reports its three requested
shifts as nominal targets.

Off-target change is reported under three conventions. The first is a concept-decomposed
Kullback--Leibler (KL) divergence that separates the intended change on the objective tokens from
changes elsewhere. The second, \texttt{off\_target\_kl\_set}, evaluates the KL divergence over the
complement of the objective's support. The third, used for scalar objectives, is the
second-order Fisher residual
$\mathrm{KL}(p_0\Vert p_s)-\Delta\phi^2/(2\,\mathrm{Var}_{p_0}\phi)$, which subtracts the
local quadratic disturbance predicted by the intended change alone. For finite KL divergence,
this residual approximates rather than exactly decomposes off-target change. The win
rate is the fraction of prompts for which the natural-gradient step has lower off-target change at
the comparison setting.

For matched-response analyses, advantages are ratios of off-target change at matched objective
change. In the nominal-target Pythia sweep, ratios and win rates use the two arms returned at each
requested target setting. Confidence intervals are
paired bootstraps over a fixed set of 12 shared contrastive pairs, resampled in log-ratio space and
reported as 95\% intervals. Because the same 12 pairs are used at every model scale, depth and width
comparisons are paired and tested in the same space. The full prompt sets, contrastive pairs, random
seeds and exact model revisions for each analysis are available from the corresponding author on
reasonable request.

\section{Knowledge-editing details}\label{si:editing}

Each fact is edited by a contrastive natural-gradient step gated by a subject-string rule and evaluated
on CounterFact with GPT-2-XL. The rule activates when a prompt contains the edit subject after light
text normalisation and otherwise leaves the model unchanged. On the 100-edit sample, it activates
on all edit and paraphrase prompts (100\% paraphrase activation rate) and on 0.2\% of neighbourhood
prompts. Under the standard EasyEdit metrics~\cite{easyedit}---efficacy score (ES), paraphrase score
(PS), neighbourhood score (NS) and their harmonic mean $S$---the routed Fisher system obtains
$S=100.0$ (ES/PS/NS $=100/100/100$). For context, the same benchmark implementation gives
$S=85.8$ for ROME~\cite{rome} ($98/77/85$), $S=70.8$ for MEMIT~\cite{memit} ($86/50/93$),
$S=43.8$ for GRACE~\cite{grace} ($92/22/82$) and $S=30.2$ for SERAC~\cite{serac}
($20/27/81$). Because routing and method-access conditions differ, these values provide benchmark
context rather than a controlled method ranking: the Fisher system computes a query-time edit
gated by a subject-string rule, whereas the comparators are persisted editors.

A learned scope-classifier variant replaces the subject-string test with a trained classifier. It obtains
$S=89.0$ and covers $99.3\%$ of subject aliases, whereas the subject-string rule obtains $S=92.5$ at
$32\%$ alias coverage. The subject-string rule is the default for the reported locality set.

A stricter probability-margin protocol on 300 edits, in which an edit counts only if the new fact's
probability exceeds the old fact's, gives $S=92.8$ (ES $100$, PS $99.5$, NS $81.5$). The NS is
indistinguishable at the displayed precision from the unedited-model baseline of $81.6$. The
apply-operator ablation replaces the pullback-metric step with Euclidean and unpreconditioned
counterparts at matched edit efficacy ($P=0.5$) and measures off-target KL divergence on unrelated
prompts: Fisher $0.872$ ($1.0\times$), Euclidean $3.700$ ($4.2\times$) and unpreconditioned $1.217$
($1.4\times$).

The fixed routed-edit comparison removes the query-time-solve asymmetry. On GPT-2, each
Fisher and Euclidean delta is solved once on a canonical CounterFact prompt, fixed and reused, and
both arms use the same gating rule. Both operators attain the fixed $+5$ log-odds calibration shift on
all 30 held-out edits. The paired geometric-mean ratio of Euclidean to Fisher other-token KL
divergence is $10.57$ (95\% bootstrap interval $8.52$--$13.37$). This fixed routed-edit result directly
compares Fisher and Euclidean application. Ranking against ROME, MEMIT or other persisted editors
requires a common fixed-edit protocol and gating rule.
Supplementary Table~\ref{tab:si-editing} assembles the matched fixed-edit comparison and the
separate query-time results.

\begin{table}[ht]
  \centering
  \caption{\textbf{Knowledge editing in full} (CounterFact).
  Top: the matched fixed-edit comparison on GPT-2, with deltas fixed after one
  solve and a shared gating rule. The remaining sections report the separate GPT-2-XL query-time-routed
  benchmark: standard EasyEdit metrics (100 edits), the stricter probability-margin protocol
  (300 edits) and the earlier apply-operator screen. The cross-method EasyEdit values use common benchmark
  scores under different routing and data-access conditions.
  }
  \label{tab:si-editing}
  \footnotesize
  \begin{tabular}{@{}l c c c c@{}}
    \toprule
    \multicolumn{5}{@{}l}{\emph{Matched fixed routed edits (GPT-2; 30 held-out edits; shared gating rule)}} \\
    Comparison & $+5$ shift attained & \multicolumn{2}{c}{other-token KL advantage} & delta use \\
    \midrule
    Fisher versus Euclidean & 30/30 each & \multicolumn{2}{c}{$10.57\times$ ($8.52$--$13.37$)} & solved once, reused \\
    \midrule
    \multicolumn{5}{@{}l}{\emph{EasyEdit standard metrics (GPT-2-XL; 100 edits; descriptive)}} \\
    Method & Efficacy & Paraphrase & Locality & $S$ \\
    \midrule
    Routed Fisher (ours, training-free) & 100.0 & 100.0 & 100.0 & \textbf{100.0} \\
    ROME   & 98.0 & 77.0 & 85.0 & 85.8 \\
    MEMIT  & 86.0 & 50.0 & 93.0 & 70.8 \\
    GRACE  & 92.0 & 22.0 & 82.0 & 43.8 \\
    SERAC  & 20.0 & 27.0 & 81.0 & 30.2 \\
    \midrule
    \multicolumn{5}{@{}l}{\emph{Probability-margin protocol (300 edits, mass-edit)}} \\
    & Efficacy & Paraphrase & Neighbourhood (base $81.6$) & $S$ \\
    \midrule
    Routed Fisher & 100.0 & 99.5 & 81.5 & 92.8 \\
    \midrule
    \multicolumn{5}{@{}l}{\emph{Apply-operator ablation at matched efficacy $P(\mathrm{target}){=}0.5$ (100 edits)}} \\
    & \multicolumn{2}{c}{off-target KL (median)} & fluency & target reached \\
    \midrule
    Fisher apply & \multicolumn{2}{c}{$0.872$ $(1.0\times)$} & 3.84 & 100\% \\
    Euclidean apply & \multicolumn{2}{c}{$3.700$ $(4.2\times)$} & 3.74 & 94\% \\
    Unpreconditioned apply & \multicolumn{2}{c}{$1.217$ $(1.4\times)$} & 3.80 & 100\% \\
    \midrule
    \multicolumn{5}{@{}l}{\emph{Subject-string gating}: paraphrase 100\%;\ neighbourhood 0.2\%.} \\
    \bottomrule
  \end{tabular}
\end{table}
\FloatBarrier

\section{The instruct-model programme}\label{si:instruct}

Three alignment objectives are steered by the same untuned preference covector
$q=\nabla_h[\log p(y_w)-\log p(y_l)]$ (main-text Methods): sycophancy (honest-over-flattering
continuations), truthfulness (capital-city true/false pairs) and
positive writing style. Capability retention is measured as off-target KL divergence on held-out
neutral prompts that are semantically distinct from the steered attribute. The 20-item
multiple-choice check changes in increments of $1/20=0.05$ and showed no Fisher--Euclidean
difference. This discrete endpoint did not resolve the distributional difference captured by
off-target KL, which differed by a factor of $36$--$50$ on Qwen2. For localised objectives, the analyses
measure on-prompt off-target KL divergence. For the diffuse style objective, intended changes also
affect this quantity, so retention is measured on the held-out neutral prompts.

At equal interpolated aggregate mean behaviour change, the natural-gradient update reduces off-target
KL divergence relative to the Euclidean update by factors of $8$--$74$ for sycophancy, $30$--$190$
for truthfulness and
$4$--$550$ for style. Contrastive activation addition~\cite{caa} often does not reach the target
change for these objectives. Objectives combine linearly as
$(G+\alpha I)^{-1}(w_1q_1+w_2q_2)$, with one weight per objective; the weight required for a given
balance is model-dependent. The results replicate on Qwen1.5-1.8B-Chat.

\begin{figure}[htbp]
  \centering
  \includegraphics{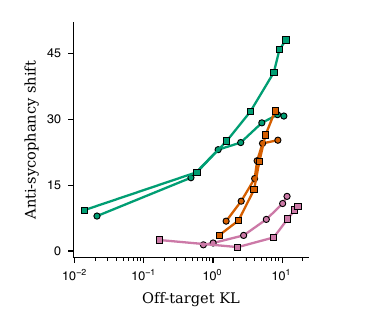}\hspace{0.20in}\includegraphics{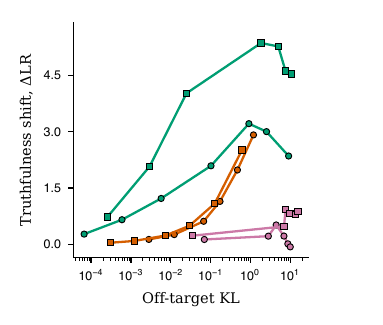}\par
  \includegraphics{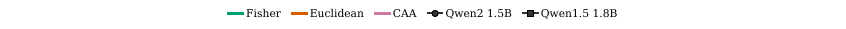}
  \caption{\textbf{Instruction-model control frontiers.} Left, anti-sycophancy; right, capital-city
  truth preferences. Qwen2-1.5B-Instruct and Qwen1.5-1.8B-Chat each contribute the complete
  19-point sweep: seven natural-gradient, six Euclidean-gradient and six contrastive-activation-addition
  settings. Cost is off-target KL on held-out neutral prompts; the horizontal coordinate is mean
  objective change. Interpolated comparisons match aggregate effects. The style and composition
  frontiers are shown in Extended Data Figure~\ref*{Main-fig:ed-intervention-frontiers}.}
  \label{fig:si-instruct-frontiers}
\end{figure}
\FloatBarrier

\section{Sparse-autoencoder feature ablations}\label{si:codezero}

The feature experiment uses one Top-$k$ sparse autoencoder at layer 12 of
Pythia-410M. Feature selection uses 18 discovery contexts; evaluation uses 18 disjoint context
clusters. Of 71 selected features, 57 are active in evaluation, giving 303 unique context--feature
ablations. Each intervention sets the active code to zero before decoding.
The measured cost is exact full-vocabulary $\mathrm{KL}(p_0\Vert p_{\rm zero})$; the Fisher
prediction $\tfrac12\Delta h^\top G\Delta h$ is computed before observing the causal outcome.

Across the 303 interventions, the Fisher prediction has Spearman correlation $0.997$ with measured
KL, compared with $0.896$ for activation magnitude. Their paired difference is $0.101$ (95\%
cluster-bootstrap interval $0.069$--$0.136$). For outcomes independent of the distributional cost
approximation, the objective-effect-to-Fisher-cost ratio correlates with the held-out selected-token
probability-change fraction at $0.935$ versus $0.506$ for attribution-patching magnitude, and with sign agreement at $0.746$ versus
$0.354$. The corresponding paired-difference intervals are $0.296$--$0.553$ and
$0.289$--$0.492$. The objective-effect-to-Fisher-cost ratio is the stronger correlate of the selected-token probability-change fraction and sign agreement,
whereas attribution-patching magnitude is the stronger correlate of absolute signed effect
($0.987$ versus $0.525$).

\section{Dictionary-learning details}\label{si:dictlearn}

Top-$k$ sparse autoencoders~\cite{topksae} are trained with only the decoder update preconditioned by the
natural gradient in the code geometry~\cite{amari1998natural},
$\Delta D \propto (\mathbb{E}[a a^\top] + \varepsilon I)^{-1}\nabla_D \mathcal{L}$, where $a$
denotes the sparse code. The code Gram matrix $\mathbb{E}[a a^\top]$ is not formed explicitly; the
preconditioner is applied to each batch by a Woodbury solve on the low-rank batch code matrix and
agrees with the explicit inverse to $7\times10^{-13}$. Residual-stream activations have
heavy-tailed norms, with a maximum approximately 15 times the median in the training data. In the
reported configuration, per-token norms are winsorised at approximately the 95th percentile before
training.

Comparisons use three paired seeds. Mean feature activation frequency is approximately $k/L$ by
construction for a Top-$k$ autoencoder with dictionary size $L$ and is reported as a diagnostic.
Because recovery alone need not detect poorly used features, reconstruction variance explained and
the fraction of features activated at least once are both reported alongside it. Disentanglement is scored by Hungarian assignment
of learned decoder atoms to planted fine-grained features using absolute cosine similarity, followed
by the mean matched cosine. In the evaluated 410M configuration with $\beta=0.5$,
natural-gradient preconditioning increases mean matched cosine from $0.167$ to $0.480$ ($+0.313$),
with an increase in all three seeds. Mean reconstruction variance explained increases from $0.535$
to $0.622$, and the fraction of features activated at least once from $21.8\%$ to $62.3\%$. Scale comparisons hold the
per-feature training budget fixed.

\section{Cross-model convergence: protocols and controls}\label{si:convergence}

Three context batteries are used. The natural battery is 200 sentence prefixes drawn from the
held-out WikiText test split~\cite{wikitext}, cut mid-sentence at 8--20 words; the templated battery
is 160 contexts from eight topic templates; the semantic battery is 120 prompts across eight
factual relations whose next token is the answer. The natural battery supplies the primary
estimates.

Agreement between two models is the Spearman correlation between the upper triangles of their
context-by-context Fisher--Rao distance matrices. On the templated battery, output-distribution
geometry gives higher agreement than mid-layer activations under three measures used in the
representation-similarity literature: mutual $k$-nearest-neighbour alignment (0.778 versus 0.710),
linear centred-kernel alignment (0.975 versus 0.935) and distance-matrix agreement (0.935 versus
0.755). In the reparameterisation experiment, each model's mid-layer states are transformed by an
independent random invertible map with condition number $\kappa$ swept from 1 to $10^4$;
output-distribution agreement is constant to machine precision at every
$\kappa$ while centred-kernel alignment and mutual $k$-nearest-neighbour alignment degrade by 10\% and
15\% by $\kappa = 10^4$ and cosine distance-matrix agreement by 28\% by $\kappa = 10^3$ (its
apparent recovery at $\kappa = 10^4$ coincides with near-singular transformed activations).

The corpus-specificity control illustrates why the natural battery is primary: the same-corpus
mediation advantage is $+0.27$ (95\% interval $+0.14$ to $+0.38$) on the templated battery but
falls to $+0.03$ ($-0.04$ to $+0.10$) on natural text at matched tokenizer and scale; the interval
spans zero.

Reference mediation conditions each target-model pair's agreement on the geometry of one reference
model. The mediated fraction rises from 7\% with a 70-million-parameter reference to a peak of 54\%
with a 2.8-billion-parameter reference and then saturates,
with 7-billion-parameter references from other families mediating 41--42\%. The qualitative result
is stable across the tested output metrics: cross-model agreement is 0.952 under Fisher--Rao, 0.947
under Jensen--Shannon, 0.935 under total variation and 0.892 under a Euclidean metric on
probabilities. The selection of Fisher--Rao specifically follows from the reparameterisation result
and Chentsov's theorem.

As a separate activation-space grounding control, over 100 concrete concepts the mean cosine-distance
geometry of last-token, last-hidden-state representations from Pythia-1.4B and GPT-2 Large agrees
with the representational geometry of DINOv2~\cite{dinov2}, evaluated on CIFAR-100~\cite{cifar100}, at 0.385
(permutation $P<0.001$), 84\% of which survives partialling out textual co-occurrence; the
result is stable across two vision models and two language models.

A recent critique shows that representational-similarity metrics inflate with model scale, because
spectral measures such as centred-kernel alignment carry a non-vanishing null baseline that grows
with representation dimension, and recommends recalibrating every score against a permutation
null~\cite{groger2026aristotelian}. That calibration was applied to the reported metrics ($200$
row-permutations per model pair; ten models from $70$ million to $7$ billion parameters across six
families, thirty-five cross-family pairs on the natural battery). The rank agreement of distance
matrices used here has a permutation null of $0.001$, independent of dimension, whereas centred-kernel
alignment on the same mid-layer representations carries a null baseline of $0.316$, reproducing the
confound: across the ladder its raw score more than doubles with scale ($0.35$ to $0.73$), the
inflation identified by the critique, while the output-distribution rank agreement stays high and flat
(calibrated $0.85$ to $0.89$).

After calibration, the Fisher--Rao output-distribution agreement is essentially unchanged
($0.879$ to a calibrated $0.869$, retaining $99\%$) and remains the strongest
agreement among the representations compared, whereas mid-layer centred-kernel alignment is reduced
most strongly (retaining $74\%$) and has the weakest significance. Thus, the dimension-dependent
null identified by the critique affects centred-kernel alignment; the rank correlation of distance
matrices used for the primary output-distribution analysis remains stable.

\section{The displacement-magnitude-matched null and the anisotropy analysis}\label{si:nulls}

Claims that two models share eigenvector structure require a null that preserves displacement magnitude. The
decision axes of the output Fisher metric, expressed as probability displacements over a common
vocabulary, assign each word a displacement magnitude and a direction within the
spectrum of the metric. A shuffled-word null destroys both and makes sharing look ubiquitous: the
top-six cross-family subspace overlap is 0.698 against a shuffled null of 0.025. A
displacement-magnitude-matched null preserves each word's displacement magnitude and randomises only the
directions; against it the same overlap has a null of 0.636, so the residual eigenvector overlap is
$+0.063$ (95\% interval $+0.056$ to $+0.070$), about 9\% of the raw overlap. The remaining 91\%
is the shared displacement-magnitude structure, which follows from the shared output distributions alone. The
residual component is concentrated in the top modes (templated battery, where per-band residuals are resolvable), decaying from $+0.080$ in the first five to
$+0.013$ past mode ninety, and differs by training lineage: $+0.088$ within a model family
against $+0.064$ across families, with a shared tokenizer alone conferring no advantage
($+0.061$).

Separately, raw cosine similarities of embedding rows are dominated by a shared mean direction
and can invert cross-model comparisons outright: one model's raw unembedding geometry
anti-correlated with every other model's before correction, and the cross-family agreement rose
from 0.498 to 0.885 after subtracting the vocabulary mean. Covariance-based quantities are immune:
the categorical factor of the output Fisher metric annihilates the all-ones direction, so adding
any constant vector to all unembedding rows leaves the metric, its eigenvectors and the
displacement analysis exactly unchanged. Accordingly, the shared-structure residual computed from
raw and mean-centred unembeddings is identical ($+0.063$ in both cases), whereas cosine analyses
require mean-centring for this comparison.

An objective-level robustness analysis probes this link beyond the band-level co-localisation of
the main text. Across four small models (GPT-2, GPT-Neo-125M, Pythia-70M and Pythia-410M; 180
model-pair~$\times$~objective items), partial Spearman associations conditioned on generalised
rank fraction and log generalised eigenvalue are tested against Freedman--Lane residual
permutation nulls within model-pair~$\times$~objective blocks (5{,}000 permutations, Holm
multiplicity adjustment). The Fisher-cost--sharing and objective-sensitivity--sharing associations
survive (both adjusted $p = 0.001$, and the direction is stable under leave-one-model-out and
leave-one-objective-out re-analysis), while the direct item-level cost--meaning association at
matched training effect does not (observed $-0.016$, adjusted $p = 1.0$) and sharing--meaning is
marginal (adjusted $p = 0.074$). The association with semantic relevance is therefore supported at the band level of the
main text, where answer-relevant motion concentrates in the same top modes. Because model pairs
share endpoints, the pooled analysis uses a
per-block permutation scheme that preserves this dependence.

\section{Meaning decomposition and transfer probes}\label{si:meaning-protocols}

The consensus geometry is the mean of the models' rank-transformed distance matrices, and a
model's residual is its matrix with the consensus partialled out. Against the external semantic
geometry defined by the \texttt{all-MiniLM-L6-v2} Sentence-BERT encoder~\cite{sbert}, the residual has negligible alignment on each battery ($-0.001$ on natural, $+0.003$ on templated,
$+0.051$ on the semantic battery), while the consensus carries 0.044, 0.102 and 0.440
respectively. The alignment increases when the next token directly encodes the answer to a factual
relation. On the
semantic battery the prompts within a relation share a template, so the surface geometry is
itself strongly aligned (0.614); the consensus retains $+0.231$ (95\% interval $+0.159$ to
$+0.316$) after partialling the surface out, and the surface retains $+0.518$ after partialling
the meaning out, so template and meaning are separate strong drivers there.

Transfer probes use relative representations: each prompt is its vector of Fisher--Rao distances to a fixed
40-prompt anchor set, $z$-scored per model, and a multinomial logistic probe for the eight-way
topic label is trained on one model and tested on held-out prompts of another, over all ordered
model pairs and five folds. Accuracy is 0.724 within model, 0.659 across models and 0.700 for
the consensus alone, against a chance level of 0.125; the residual supports 0.407 within its own
model and 0.091, below chance, across models.

\subsection{Native-sample human semantic geometry}
The native-sample panel uses 512 sentence-final contexts from the Peelle et al. completion
norms~\cite{peelle2020norms}. At each of six Pythia sizes, 32 samples per context were drawn at
temperature one from the full next-token law, without top-$k$ or nucleus filtering. Sampling
had an eight-token limit; completions were parsed at the first punctuation, newline or
end-of-sequence boundary. All samples, including empty completions and those reaching the limit,
entered the analysis. The limit was reached by $27.2$--$36.3\%$ of samples across models.

Each normalised completion was encoded by the fixed 384-dimensional
\texttt{all-MiniLM-L6-v2} encoder, then mapped to 768 random Fourier features for an equal mixture
of Gaussian kernels with bandwidths $0.25$, $0.5$ and $1.0$. Sixteen disjoint folds contained
32 contexts each. Completion feature means define the semantic distributional representation;
its squared Euclidean distance and centred Gram-matrix CKA compare model and human predictions.
The four pairings of independent human-participant halves and model-sample halves were averaged.
CKA was divided by the square root of human and model split-half CKA, using the same fold.
Thus these endpoints describe semantic kernel geometry, whereas the native next-token
Fisher--Rao distances in the cross-model panel are computed directly from probabilities.

For 70M, 160M, 410M, 1B, 1.4B and 2.8B, mean squared model--human semantic distances are
$0.33885$, $0.32634$, $0.30841$, $0.30316$, $0.30046$ and $0.29576$.
The slope against natural-log parameter count is $-0.011809$ (95\% paired fold-bootstrap
interval $-0.013017$ to $-0.010627$). Reliability-corrected semantic CKA values are $0.82101$,
$0.82680$, $0.84162$, $0.84450$, $0.84500$ and $0.84005$, with slope $0.006290$
($0.000070$--$0.012880$); the overall trend need not be monotone between adjacent sizes.
A character 3--5-gram feature control also gives decreasing model--human distance, with slope
$-0.012915$ ($-0.014727$ to $-0.011243$), while its CKA slope interval includes zero.

\subsection{Predictive fit and developmental human alignment}
A separate 384-context subset of the same corpus was divided into 16 folds. Model probabilities
were evaluated on complete human-observed completion-plus-end-of-sequence events and normalised
within this support; token-sequence collisions were retained. Risk is the context mean of
$\sum_y(\sqrt{p_y}-\sqrt{h_y})^2$ for model and human conditional laws $p$ and $h$.
CKA is averaged over the two human halves and divided by the square root of human split-half
reliability; the model feature means are deterministic conditional expectations. For each fold, a linear relation
between conditional root-probability risk and corrected semantic CKA was fitted on six Pythia
sizes and five Pythia-160M checkpoints. Its risk slope was $-0.162883$ (95\% fold-bootstrap
interval $-0.202616$ to $-0.125890$). Applied without refitting to five OLMo-2-1B
checkpoints~\cite{olmo2} and BLOOM-560M, GPT-2 Large, GPT-Neo-1.3B, Mamba-130M and Qwen2-1.5B,
the relation gives correlation $0.758320$ and root-mean-square error $0.029465$ over all
160 model--fold observations. These rows share contexts and checkpoints and do not constitute
160 independently trained models.

OLMo alignment increases across checkpoint order, with slope $0.018891$
($0.013438$--$0.025056$). The Pythia trajectory slope is $0.002663$
($-0.002270$--$0.008102$). The residual log-parameter coefficient after conditional risk is
$0.002872$ ($-0.005020$--$0.009876$). Together with the cross-family prediction, these comparisons
support predictive fit as an organising variable for human alignment. The conditional-law
analysis and the native-sampling analysis evaluate different completion distributions.

\subsection{Leading directions in an independent human probability geometry}
The de Varda et al. corpus~\cite{devarda2024cloze} supplies 1,726 context positions in 205
sentences. Human counts used the released exclusion, first-word and spelling-correction rules.
Both model and human laws were mapped to 64 semantic outcome bins fixed independently of this
corpus, plus a residual outcome retaining the remaining mass. Human laws used a Dirichlet
posterior mean with total concentration $0.5$ and a fixed external reference law $\rho$.
Coordinates were the centred log probabilities weighted by $\sqrt\rho$, as in the
standardised read-out representation. Four rotations of whole-sentence folds separated estimation
of the human reference, estimation of model directions, calibration and final evaluation.
The leading four model directions were learned without human target responses.

Directional recovery was evaluated from the component of human coordinate variation orthogonal
to the model-derived frame. Writing $R_{\mathrm{orientation}}$ for this component and $m_z$ for
the smallest eigenvalue of human in-frame variation, the directional lower bound is
$1-R_{\mathrm{orientation}}/(4m_z)$. Contributions were summed over rotations before taking
the ratio. Simultaneous one-sided 95\% lower bounds over the seven families used 2,048
whole-sentence bootstrap draws. For Pythia, Mamba, RWKV, GPT-2, GPT-Neo, Qwen2 and BLOOM,
they are $0.34055$, $0.37935$, $0.37457$, $0.38306$, $0.31937$, $0.41116$ and $-0.00120$.
Direct directional capture exceeds a Haar-random four-dimensional subspace by
$0.50$--$0.61$ across the seven families. The direction result also retains its sign when
human participants used for reference estimation and evaluation are disjoint.

These measurements concern directions. Recovering the full human probability law additionally
requires the appropriate origin and coordinates within the frame. Full-reconstruction bounds
remain non-positive in this corpus under the evaluated native and low-rank calibrated maps;
the directional result therefore establishes shared leading structure rather than full-law
recovery.

\subsection{Independent human-completion replication and model-only calibration}
The independent analysis used 640 word positions selected without response values from five DERCo
narratives~\cite{quach2024derco}. The selected positions contain 64,000 responses from 499
participant identities. For Pythia 70M, 160M, 410M, 1B, 1.4B and 2.8B, cross-fitted native
model--human semantic distances are $0.23348$, $0.22450$, $0.21618$, $0.21151$, $0.20084$ and
$0.20335$. The equal-story slope against natural-log parameter count is $-0.008725$ (95\%
interval $-0.012022$ to $-0.005556$; Extended Data Fig.~\ref*{Main-fig:ed-displacement-null}b).
The contexts and completion units differ from the Peelle analysis above, so absolute distances are
not pooled across corpora; the prospectively fixed comparison is the within-corpus slope.

Across the five ordered training stages, Pythia-160M distances are $0.34380$, $0.26562$,
$0.22647$, $0.22604$ and $0.22300$, while OLMo-2-1B distances are $0.30403$, $0.21441$,
$0.20043$, $0.19928$ and $0.19441$. The final-minus-initial changes are $-0.12080$
($-0.13529$ to $-0.10470$) and $-0.10962$ ($-0.12571$ to $-0.09223$), respectively
(Extended Data Fig.~\ref*{Main-fig:ed-displacement-null}c). In the held-family conditional
analysis, the risk-based predictor has root-mean-square error $0.03825$, compared with $0.06187$
for a constant and $0.04991$ for log model size. Its paired advantage over the constant is
$0.02363$ ($0.00264$--$0.04143$); the advantage over size is $0.01166$ and is not resolved
($-0.00203$--$0.03638$). Thus the independent data establish improvement over a constant
predictor but do not distinguish risk from size for this endpoint.

The model-only calibration retained all 64 fixed semantic bins plus the residual outcome and was
learned entirely from model probability laws on the earlier corpus. Its equal-story expected
human log-score gains over each native law are $0.04520$ nats for Pythia, $0.04078$ for Mamba,
$0.03865$ for RWKV, $0.02202$ for GPT-2, $0.01962$ for GPT-Neo, $-0.00297$ for Qwen2 and
$0.05074$ for BLOOM (Extended Data Fig.~\ref*{Main-fig:ed-displacement-null}d). Five of the six
positive family means are positive in every story and GPT-Neo is positive in four of five; Qwen2
is positive in three. By contrast, maps trained on human responses from the earlier corpus reduce
log score in all seven families ($-0.18212$ to $-0.10584$ nats). The transferable gain is therefore
specific to the model/source-law calibration rather than generic cross-corpus fitting. These are
effects over five fixed observed stories, not a population-of-stories threshold claim, and the
coarse outcomes do not establish individual-word recovery.

\clearpage
\section{Reproducibility map}\label{si:repro}

\begin{longtable}{@{}>{\raggedright\arraybackslash}p{0.20\linewidth}
>{\raggedright\arraybackslash}p{0.74\linewidth}@{}}
\caption{\textbf{Evidence map for the central claim chain.} The table organises each strand by its
evidence and role in the manuscript.}\label{si:tab:evidence-map}\label{tab:si-evidence-map}\\
\toprule
Strand & Evidence and role in the manuscript \\
\midrule
\endfirsthead
\toprule
Strand & Evidence and role in the manuscript \\
\midrule
\endhead
Origin and identification & Coordinate-covariance tests, across-scale curvature and profile calibration, initialisation controls and standardised-frame certificates establish the canonical metric and its identified content. \\
Shared geometry & Cross-model agreement and meaning batteries, exact attenuation accounting, common-outcome and common-byte comparisons, held-out prediction, independent human completion comparisons and fresh model-only calibration establish shared predictive structure. \\
Spectral anatomy & Core-to-tail response measurements, effective-dimension agreement without fitted parameters, inheritance bounds, singleton-head/cluster-bulk anatomy and fresh-context tests and the probability/read-out comparison establish how language statistics shape the spectrum. \\
Acquisition schedule & Cross-size alignment by held-out loss, corpus $n$-gram forecasts applied without refitting, randomised logarithmic-time depth shifts, crossed architecture--construction generalisation and real-checkpoint observational composition connect corpus evidence to acquisition time. \\
Motion and learning & Continuous motion checks, the exact flip-margin identity, objective--update factorisation and read-out projector kinematics connect geometric motion to behavioural change. \\
Control geometry & Chart-covariant damping and the prediction of natural-versus-Euclidean intervention cost without fitted parameters establish the minimum-disturbance law and its relative cost; relinearised paths and reusable reference-metric updates extend its measured scope. \\
Translation & Instruction-model frontiers, dictionary learning, sparse-autoencoder feature ablations, matched fixed routed edits and a held-out low-rank adaptation comparison test the same correction across model operations. \\
\bottomrule
\end{longtable}

Pretrained-model experiments include publicly available checkpoints from the Pythia suite~\cite{pythia},
GPT-2~\cite{gpt2}, Qwen2~\cite{qwen2}, Qwen1.5~\cite{qwen15}, GPT-Neo~\cite{gptneo},
Mistral~\cite{mistral7b}, Mamba~\cite{mamba}, RWKV~\cite{rwkv}, BLOOM~\cite{bloom} and
StarCoder2~\cite{starcoder2}. The main-text randomised causal acquisition results use explicitly
controlled synthetic systems. Model execution and intervention solves use float32 unless an
analysis specifies float64 dense spectral arithmetic. Most GPU-resident model-forward runs use
one NVIDIA RTX 5000 Ada with 32\,GB of memory, with additional model runs on NVIDIA A100 GPUs and CPU execution for dense statistical analyses.
A full GPU-resident pullback solve fits through
2.8 billion parameters. At 6.9 billion, the evidence is limited to graph-free output-Fisher spectra,
a CPU-only mid-layer full-vocabulary pullback calibration and a descriptive late-layer CPU-offload
iteration analysis. Given a
random seed, the conjugate-gradient and Woodbury solvers are deterministic; seeds are logged for
bootstrap resampling, dictionary learning and Monte Carlo rollout values. The map above organises
the central claim strands. Materials supporting the reported analyses, including model revisions,
prompt, pair and objective-set listings, relevant analysis code and summary logs, are available from
the corresponding author on reasonable request; code and data availability are stated in the main text.

\renewcommand{\refname}{Supplementary References}
\bibliographystyle{unsrtnat}
\bibliography{refs}

%% file: preamble.tex
\ifreview
  \usepackage[a4paper,margin=1in]{geometry}
\else
  \usepackage[a4paper,margin=1.05in,headsep=14pt]{geometry}
\fi
\usepackage{setspace}
\ifreview\fi
\usepackage{lineno}

\usepackage{amsmath,amssymb,mathtools}

\ifreview\else
  \usepackage{newtxtext,newtxmath}
\fi

\newcommand*\patchAmsMathEnvironmentForLineno[1]{%
  \expandafter\let\csname old#1\expandafter\endcsname\csname #1\endcsname
  \expandafter\let\csname oldend#1\expandafter\endcsname\csname end#1\endcsname
  \renewenvironment{#1}%
    {\linenomath\csname old#1\endcsname}%
    {\csname oldend#1\endcsname\endlinenomath}}%
\newcommand*\patchBothAmsMathEnvironmentsForLineno[1]{%
  \patchAmsMathEnvironmentForLineno{#1}%
  \patchAmsMathEnvironmentForLineno{#1*}}%
\AtBeginDocument{%
  \patchBothAmsMathEnvironmentsForLineno{equation}%
  \patchBothAmsMathEnvironmentsForLineno{align}%
  \patchBothAmsMathEnvironmentsForLineno{flalign}%
  \patchBothAmsMathEnvironmentsForLineno{alignat}%
  \patchBothAmsMathEnvironmentsForLineno{gather}%
  \patchBothAmsMathEnvironmentsForLineno{multline}}

\usepackage{graphicx}
\graphicspath{{figs/}}

\newcommand{\panel}[2]{%
  \vtop{%
    \hbox{{\fontsize{8}{9}\selectfont\sffamily\bfseries #1}}%
    \vskip 1.5pt
    \hbox{\includegraphics{#2}}}}
\newlength{\panelsep}
\usepackage{booktabs}
\usepackage{array}
\usepackage{siunitx}
\usepackage[labelfont=bf,font=small]{caption}

\usepackage{microtype}
\usepackage[T1]{fontenc}

\usepackage{xcolor}
\newif\ifpendingvisible
\pendingvisibletrue

\definecolor{canonical}{HTML}{009E73}
\definecolor{euclid}{HTML}{D55E00}

\usepackage[numbers,super,sort&compress]{natbib}
\usepackage{hyperref}
\SetLinkTargetFilter{main-#1}
\usepackage{xr}
\IfFileExists{si.aux}{\externaldocument[SI-][nocite]{si}}{\input{references-si.tex}}
\hypersetup{colorlinks=true,allcolors=blue!50!black}
\usepackage{cleveref}

%% file: sections/introduction.tex
A language model's predictive behaviour does not privilege any coordinate system for its internal
activations: the same input--output function can be realised in any coordinate system related by an
invertible linear transformation.
Euclidean distances between activations are not invariant under these reparameterisations, yet
Euclidean inner products underlie common measures of activation distance, feature importance,
steering and fine-tuning regularisation. On the other hand, a model's next-token probability
distribution can retain substantial information about its internal state and input: decoder-only
Transformer hidden-state mappings are almost surely injective under mild
conditions~\cite{nikolaou2025injective}, and much of an input prompt can be
reconstructed from the next-token distribution alone~\cite{morris2024inversion,nazir2025pils}. A
Riemannian metric assigns local lengths and angles to infinitesimal changes on a space. Chentsov's
theorem shows that invariance under transformations preserving statistical information gives the
Fisher--Rao metric as the unique canonical metric on the space of probability distributions, up to an
overall scale~\cite{cencov1982,rao1945,amari2016}. Behaviour space therefore has a privileged geometry
even when representation space does not.

Pulling the output Fisher metric back through the network measures an activation change by the local
output change it produces. The corresponding damped
natural gradient achieves a specified objective change at minimum regularised local output
cost~\cite{amari1998natural,martens2020natural}.

The same geometry supports comparison across models: for common contexts, each model gives a matrix of
pairwise Fisher--Rao distances between its next-token distributions. Comparing these matrices
requires no shared vocabulary, architecture or activation coordinates. This recasts the convergence
phenomenon behind the Platonic Representation Hypothesis~\cite{huh2024platonic} in an output geometry
where, under the stated conditions, convergence follows from predictive fit rather than from a
chosen activation-similarity measure~\cite{kornblith2019cka,moschella2023relative}. Activation
geometry has no comparable standing:
the same behaviour admits invertible hidden-state reparameterisations with different Euclidean
activation geometries, leaving activation-space convergence as an empirical property of training.

Closed-form natural-gradient steps grounded in Fisher information geometry have recently been
validated for activation steering~\cite{fishback,park2026}, with FishBack extending to
8-billion-parameter models. Nonlinear activation-space steering has also been
explored~\cite{curveball}; representation-level steering baselines are established
practice~\cite{caa,repe,actadd,iti}; and
identifiable properties of next-token predictors have been characterised
theoretically~\cite{marconato2024allornone}. Analyses of approximate agreement distinguish
predictive closeness from representational similarity, including limitations of KL-based guarantees
and sufficient conditions based on logit distance~\cite{nielsen2025closeness,nielsen2026logit}.
Approximate identifiability has also been studied for representations with nonlinear
decoders~\cite{nelson2026identifiability}.

Related work has linked next-token statistics to representation
structure~\cite{zhao2024implicit,zhao2025semantics}, identified statistical symmetries shaping
geometric structure~\cite{cyclicsymmetry}, and studied transient semantic organization during
training~\cite{zhao2026structure}. Corpus correlations have also been linked to hierarchical language
acquisition and data-limited loss scaling~\cite{cagnetta2024acquisition,cagnetta2026scaling}.
Other studies document cross-model convergence and cross-architecture steering transfer in internal
representations~\cite{kim2026subspace,agarwal2026transfer}. Here we connect these strands into a
framework defined by the geometry of model outputs: predictive behaviour identifies the geometry,
language statistics predict its structure and acquisition, and its measured anisotropy predicts
intervention cost across architectures and operations.

Predictive behaviour determines the canonical output geometry up to symmetries that preserve
outputs, with explicit bounds on recovery error, whereas activation geometry remains
coordinate-dependent under invertible reparameterisation. Across ten independently trained models
spanning transformer, state-space and recurrent architectures, the identified relational geometries
have a mean rank agreement of 0.88, compared with 0.62 for mid-layer activation geometries;
their shared component supports semantic-category transfer. Human completion geometry becomes
closer with predictive fit, and a relation fitted on one model family predicts agreement in others. Controlled language assignments causally determine the learned geometry across architectures.
Token probabilities and learned read-out structure explain complementary aspects of its spectrum,
with effective dimension predicted without fitted parameters.
At fixed relative damping and solver tolerance, the damped natural-gradient solve has a worst-case
conjugate-gradient iteration bound independent of model width. Corpus statistics also predict when behaviours are acquired: in a
randomised synthetic-language test, assigning otherwise identical facts to deeper statistical
evidence delays early acquisition more than fourfold. The geometry also
prescribes minimum-disturbance interventions. A measured anisotropy ratio, which quantifies how
unevenly different activation directions affect the output, predicts their advantage over Euclidean
control without fitted parameters. Averaging the metric over reference prompts produces reusable
updates that transfer to unseen prompts while reducing reference-sequence change. Applying the same correction improves steering, knowledge
editing, feature attribution, dictionary learning and fine-tuning, with up to two orders of
magnitude less off-target change.

%% file: sections/results.tex
\section*{Results}

\subsection*{Predictive behaviour fixes a canonical, identifiable output geometry}

\begin{figure}[htbp]
  \centering
  \makebox[\linewidth]{\panel{a}{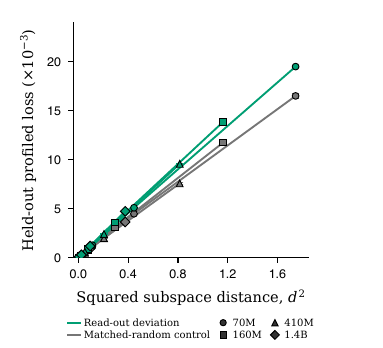}\hspace{\panelsep}\panel{b}{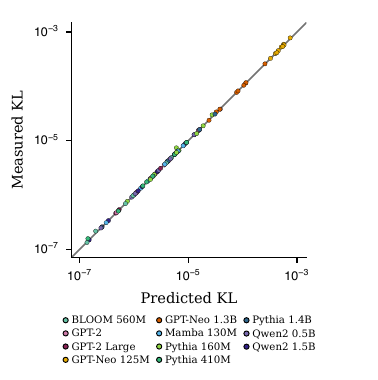}}\par\vspace{-3pt}
  \caption{\textbf{Behaviour identifies the resolved read-out, and geometry predicts local output change.}
  \textbf{a}, Behavioural identification: held-out predictive loss after re-optimising nuisance
  parameters grows linearly with the squared subspace distance between a perturbed read-out subspace
  and the resolved subspace
  (Pythia 70M--1.4B). Markers are the 40 model--condition--angle means and bars span the two
  independently calibrated held-out folds (smaller than the markers); paths connect angles within
  each model and condition. Green, read-out deviation; grey, matched-random control; circles, 70M;
  squares, 160M; triangles, 410M; diamonds, 1.4B. The local
  fits have $R^2 \ge 0.9999$ on all eight paths, and matched-random control directions have
  systematically smaller loss--distance slopes (slope ratios $1.14$--$1.27$). The reference subspace is estimated from a stronger model (Supplementary Note~\ref*{SI-si:identification}).
  \textbf{b}, Predicted and measured output change under small activation perturbations.
  Each point is one model--depth--prompt combination, giving 99 measurements across 11 models,
  six families and quarter-, half- and three-quarter-depth injection. The prediction
  $\tfrac12\delta h^\top G\delta h$ uses the metric before the intervention, without a fitted
  scale. Points show the coarsest resolved local steps; colour identifies the model and the grey
  line marks equality. The measured-to-predicted KL ratio has median $1.000$ and
  range $0.948$--$1.224$
  (Methods; Supplementary Note~\ref*{SI-si:theory}).}
  \label{fig:ident}
\end{figure}

A language model's output is a probability distribution over its vocabulary. The Fisher--Rao
metric measures local separation between such distributions and is unique up to scale by its
invariance under sufficient-statistic transformations~\cite{cencov1982,rao1945,amari2016}. Let $h$
be the activation at the chosen intervention layer and $h_L(h)$ the final hidden state it produces.
Writing $p_h=\mathrm{softmax}(W_Uh_L(h)+b_U)$ for the output law, with unembedding rows $w_a$ and
$J_h=\partial h_L/\partial h$, the output Fisher $H_h=\mathrm{Cov}_{p_h}(w)$ defines the
(possibly degenerate) pullback Fisher metric $G_h=J_h^\top H_hJ_h$ at the intervention layer. For a small intervention,
\begin{equation}
  \mathrm{KL}\!\left(p_h\,\Vert\,p_{h+\delta h}\right)
  = \tfrac12\,\delta h^\top G_h\,\delta h
  + O\!\left(\lVert\delta h\rVert^3\right).
  \label{eq:metric}
\end{equation}
The location subscript is suppressed below. The quadratic form measures local output change and
matches the realised divergence locally across 99 model--depth--objective cells spanning 11 models,
six families and 125M--1.5B parameters (median ratio $1.000$; Fig.~\ref{fig:ident}b), so geometric
length predicts behavioural change. For an
objective gradient $q=\nabla_h\phi$, the damped natural-gradient step
$\delta h\propto(G+\alpha R)^{-1}q$~\cite{amari1998natural,martens2020natural} is
coordinate-covariant: transporting it across a seeded family of activation reparameterisations
reproduces the native direction to $<10^{-12}$, whereas resetting $R$ to the identity changes it by order one
(Extended Data Fig.~\ref{fig:ed-coordinate-controls}a; Supplementary Note~\ref*{SI-si:theory}). The connection to training is exact: excess
population cross-entropy equals the Kullback--Leibler divergence to a model-induced Gibbs law whose
conditional curvature is this output Fisher metric. Next-token training therefore induces the
geometry (Supplementary Note~\ref*{SI-si:gibbs}).

This geometry is also identifiable from behaviour alone: predictive behaviour singles out the
language-defined read-out subspace at matched rank rather than a choice of hidden coordinates. Let
$T$ denote this resolved subspace, $S$ a candidate subspace, $d(S,T)$ their chordal
distance (a standard measure of subspace separation), and $\mathcal K(S)$ the profiled predictive
risk after nuisance parameters have been optimised out. Under the stated finite-support and
inverse-chart conditions, behavioural identification obeys the global quadratic margin
\begin{equation}
  \mathcal K(S) \;\ge\; c\,d^2(S,T).
  \label{eq:ident}
\end{equation}
Here $c>0$ is a certified lower bound on predictive cost per unit squared subspace displacement,
determined by the reference law and context weights (Supplementary Note~\ref*{SI-si:identification}).
Zero profiled risk therefore identifies the resolved subspace exactly at matched rank. More generally,
finite-rank approximation and excess prediction error bound recovery at a square-root rate, so two
accurate rank-matched models of the same language must converge on the same subspace (Supplementary
Note~\ref*{SI-si:identification}).
Held-out profiled loss grows quadratically with subspace distance across four model sizes
($R^2\ge0.9999$ on every path), with steeper growth for actual read-out deviations than for
matched-random controls (Fig.~\ref{fig:ident}a).
The language law determines the learned geometry in a controlled intervention. Transformer,
gated-recurrent and diagonal-recurrent models at two capacities were trained on each of eight
pairs of synthetic languages with known conditional laws, identical token frequencies and
conditional entropy. At matched predictive accuracy, changing
the assigned law recovers more than 99\% of the imposed squared geometric separation, whereas changing
architecture within a law leaves the geometry nearly unchanged.
The assigned law is also a closer geometric match than its counterfactual in all eight pairs
(Supplementary Note~\ref*{SI-si:identification}).
By contrast, an
invertible reparameterisation of a hidden layer, compensated downstream, leaves every
conditional law unchanged: behaviour identifies the resolved output geometry while activation
geometry retains an exact $\mathrm{GL}(d)$ gauge freedom, the freedom to apply any invertible linear
change of activation coordinates (Supplementary Note~\ref*{SI-si:identification}). The resolved output geometry is therefore
invariant under behaviour-preserving reparameterisation.

\subsection*{Independently trained models share the geometry}

\begin{figure}[htbp]
  \centering
  \makebox[\linewidth]{\panel{a}{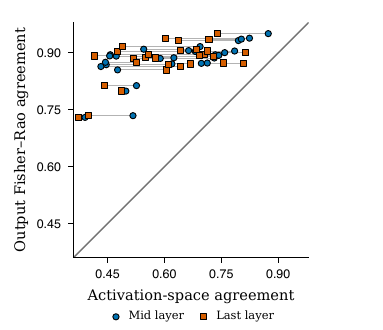}\hspace{\panelsep}\panel{b}{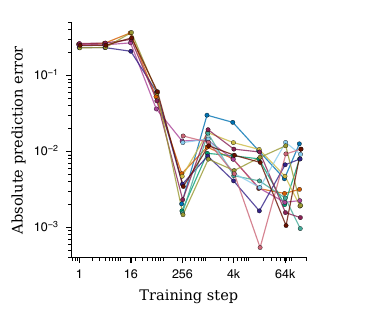}}
  \makebox[\linewidth]{\panel{c}{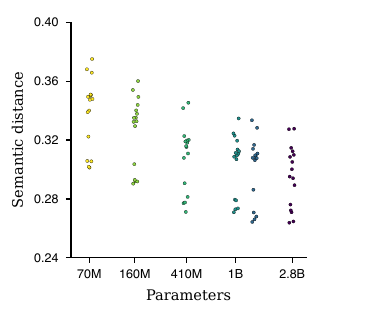}\hspace{\panelsep}\panel{d}{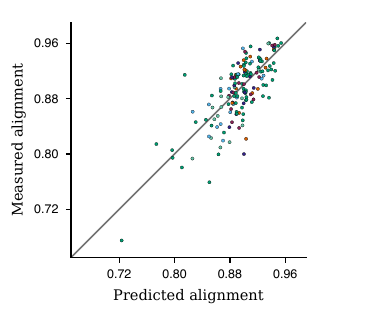}}\par\vspace{-3pt}
  \makebox[\linewidth]{\includegraphics{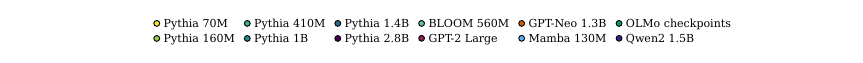}}
  \caption{\textbf{Predictive geometry is shared across models and approaches human completion structure.}
  \textbf{a}, Output Fisher--Rao agreement against activation-space agreement on natural text for
  all 29 cross-tokenizer, cross-family dyads among ten models (70M--7B; transformer, state-space
  and recurrent). Each dyad contributes a blue mid-layer circle and vermilion last-layer square
  joined by a horizontal line; the grey line is equality. The output relation is stronger than either
  activation relation for every displayed mark.
  \textbf{b}, Absolute held-out error of the exchangeable six-scalar prediction across ten public
  training checkpoints. Circles show all 96 defined predictions; colours distinguish individual
  comparisons and lines follow them across training. Six predictions are defined per checkpoint
  at steps 1--64 and twelve from step 256; estimator definitions and early-checkpoint coverage
  are given in Supplementary Note~\ref*{SI-si:stabatten}.
  \textbf{c}, Native model samples approach human completions with scale. Points show all 96
  context-fold means, 16 at each Pythia size; colours identify models in panels c and d.
  The ordinate is squared distance between semantic kernel means,
  cross-fitted over participant and model-sample halves (512 contexts, 32 samples per context;
  eight-token generation limit). No human-answer restriction is applied to sampling.
  \textbf{d}, A relation between conditional prediction risk and human semantic alignment fitted
  on Pythia is applied without refitting to five OLMo checkpoints and five external
  models. Points show all 160 model--context-fold comparisons; checkpoints and folds
  are repeated measurements. Alignment is reliability-corrected kernel CKA on a separate
  384-context battery, using model laws conditional on the observed completion events.
  The grey line is equality; correlation $0.758$, root-mean-square error $0.0295$.}
  \label{fig:shared}
\end{figure}

On a common outcome space, the convergence result in Supplementary Note~\ref*{SI-si:stabatten} bounds
relational-geometry differences by predictive error. Native-vocabulary agreement across tokenizers
is tested empirically here. For a common battery of contexts, each
model defines a matrix of Fisher--Rao distances among its next-token distributions, and models
are compared by the Spearman correlation between the matrix upper triangles. This asks whether they
order the same context pairs from near to far and needs no shared tokenizer, architecture or
activation alignment (Methods). Across ten models
spanning transformer, state-space and recurrent architectures, seven training pipelines, four
tokenizers and 70 million to 7 billion parameters, the mean rank agreement on natural text is
0.88, against 0.62 for mid-layer and 0.61 for last-layer activation geometries, a gap whose
bootstrap interval excludes zero (Fig.~\ref{fig:shared}a). Under random anisotropic
reparameterisations that leave behaviour unchanged, activation-based measures degrade steadily
while output-geometry agreement is constant to machine precision (Extended Data Fig.~\ref{fig:ed-coordinate-controls}c), and partialling out
orthographic surface geometry or the strongest corpus $n$-gram predictor leaves at least 98\% of
the agreement in place (Supplementary Note~\ref*{SI-si:convergence}). As a tokenizer-independent check, each next-token
law was mapped to the distribution of the first byte of the remaining text. In this common outcome
space, cross-tokenizer agreement is 0.91 and coarse-graining contracts the divergence on every pair,
so the sharing is not an artefact of token-level bookkeeping (Extended Data Fig.~\ref{fig:ed-coordinate-controls}b;
Supplementary Note~\ref*{SI-si:stabatten}).

The agreement can also be accounted for quantitatively. Each distance-matrix entry changes at most
in proportion to the root-probability distance $\lVert\sqrt p-\sqrt q\rVert_2$, so excess risk
forces relational convergence at a proved rate (Supplementary
Note~\ref*{SI-si:stabatten}). On same-tokenizer pairs the
resulting data-dependent certificate gives lower bounds of
$0.61$--$0.67$ against observed $0.76$--$0.83$: the models disagree substantially per context
(median root-probability distance $0.30$--$0.71$), but the disagreement directions are nearly orthogonal to the
relational structure (coherence $\approx0.03$; Supplementary
Table~\ref*{SI-si:tab:stability-certificates}). With a fixed reference geometry, agreement
decomposes into six variance and covariance terms: one reference, two residual, two
reference--residual and one residual cross-term. This identity
reconstructs every measured configuration to numerical precision (112 model-pair configurations).
Fitted on one half of the contexts, its exchangeable restriction predicts held-out agreement with
pooled median absolute error $0.009$ across the 96 defined cells. From step 256, predictions are defined for every cell and checkpoint-median errors are below $0.014$ (Fig.~\ref{fig:shared}b; Supplementary
Note~\ref*{SI-si:stabatten}).

The shared component carries semantic alignment and category transfer. Averaging rank-transformed distance
matrices gives a cross-model consensus, and each model's deviation from it is its residual. Consensus
geometry aligns with an independent sentence-encoder semantic geometry on every battery; the
alignment rises
to 0.44 when the next token answers a factual relation and survives a surface control at
$+0.23$, whereas residual alignment is indistinguishable from zero on the natural and
templated batteries and small ($+0.05$) on the semantic battery (Supplementary Note~\ref*{SI-si:meaning})~\cite{sbert}. Separately, the mean cosine-distance geometry of
last-token, last-hidden-state representations from two language models aligns with the representational geometry of a vision model trained without
language (rank agreement 0.39, permutation $P<0.001$), with 84\% of the association retained after controlling for textual co-occurrence~\cite{dinov2}. A linear semantic probe
trained on one model transfers across models with eight-way accuracy of 0.66 ($0.70$ through the
consensus geometry alone), close to the within-model 0.72 and far above the 0.125 chance level,
whereas transfer through the residual
is below chance (Supplementary Notes~\ref*{SI-si:meaning} and~\ref*{SI-si:meaning-protocols}). What is shared beyond the distributions themselves is
narrow: against a displacement-magnitude-matched null, residual eigenvector sharing between families is about
9\% of the raw overlap and is concentrated in the leading modes (Extended Data Fig.~\ref{fig:ed-displacement-null}); the
residual is also graded by training lineage (Supplementary Note~\ref*{SI-si:nulls}).

The comparison extends to human predictions. On 512 sentence contexts with human completion
norms~\cite{peelle2020norms}, semantic distributions estimated from native model samples become closer
to human completions with model scale: mean squared distance falls from about $0.34$ at 70M to
$0.30$ at 2.8B parameters (Fig.~\ref{fig:shared}c). Reliability-corrected alignment of the relational
semantic geometry is about $0.82$--$0.85$ across the six sizes (Methods).
Predictive fit also accounts for human alignment across architectures and training. On a separate
384-context battery, with model laws conditioned on human-observed completion events,
a risk-to-alignment relation fitted on Pythia sizes and checkpoints predicts
OLMo checkpoints and five external models without refitting, with correlation $0.76$ and
root-mean-square error $0.029$ (Fig.~\ref{fig:shared}d).

A separate corpus of 1,726 human next-word predictions provides a direct test in the output
probability geometry~\cite{devarda2024cloze}. After mapping model and human laws to the same 65
outcomes, model-derived leading directions recover human directional structure above a random
subspace control. Simultaneous lower bounds are positive in six of seven model families, with
alignment retained across disjoint participant groups (Supplementary Note~\ref*{SI-si:meaning-protocols}). These comparisons
connect shared model geometry to human predictive structure.

The native scale trend replicated on independent data from 640 word positions in five previously
unused narratives (64,000 responses)~\cite{quach2024derco}. Across six Pythia sizes, semantic
distance fell with log parameter count (slope $-0.00872$, 95\% interval $-0.01202$ to
$-0.00556$), and it also fell from the initial to final checkpoint in both Pythia and OLMo
(Extended Data Fig.~\ref{fig:ed-displacement-null}b,c). A calibration learned only from model
probability laws, and fixed before the new human responses, improved expected human log score for
the complete 65-outcome coarse distribution in six of seven model families (gains $0.0196$--$0.0507$
nats; Qwen2, $-0.0030$ nats; Extended Data Fig.~\ref{fig:ed-displacement-null}d). Shared model-law
structure therefore supports out-of-corpus prediction as well as directional alignment.

\subsection*{The geometry inherits the statistics of language}

In the standardised centred read-out representation, output-Fisher eigenvalues rank directions by local output effect,
and damping defines the effective dimension, a soft count of resolved modes. Let $u_a$ be token $a$'s
centred, orthonormalised read-out row after removing common-logit shifts (Methods), with
$\bar u=\sum_a p_a u_a$. The profile $q_a=p_a\lVert u_a-\bar u\rVert^2$ weights probability by
squared distance from this centre; $q_{(k)}$ lists these values in descending order. Setting
$\widehat\lambda_k=q_{(k)}$ gives a prediction of effective dimension without fitted parameters:
\begin{equation}
  \widehat N_{\mathrm{eff}}(\alpha)
  \;=\; \sum_{k=1}^{r}\frac{q_{(k)}}{q_{(k)}+\alpha}.
  \label{eq:imprint}
\end{equation}
Here $r$ is the read-out rank, and each term contributes between zero and one according to whether its
mode is resolved at scale $\alpha$. Replacing $q_{(k)}$ with the measured eigenvalues gives the
identity $N_{\mathrm{eff}}(\alpha)=\sum_k\lambda_k/(\lambda_k+\alpha)$. An inheritance theorem bounds
each leading measured eigenvalue above and below by the corresponding profile value, up to explicit
frame and tail constants. A more robust fractional-frame form requires only that a fixed fraction
of the leading direction-Gram modes remain above a declared threshold, yielding a rank-shifted
lower bound and a direction-free tail upper bound (Methods; Supplementary
Note~\ref*{SI-si:anatomy}).
Output-Fisher eigenvalues fall approximately inversely with rank,
$\lambda_i\sim i^{-\beta}$ with $\beta\approx1$, over the
rank-$2$--$80$ core window from 70M to 6.9B parameters and across families and vocabularies
(Supplementary Note~\ref*{SI-si:anatomy}). In the matched external-family panel, rank-$8$--$32$ exponents
remain of order one across all nine models from 125M to 1.5B parameters
(Supplementary Note~\ref*{SI-si:anatomy}), while the fitted exponent tracks that of the
model's own weighted token profile (median deviation $0.04$, $r=0.91$; Supplementary
Note~\ref*{SI-si:anatomy})~\cite{piantadosi2014zipf,zipf1949}. In synthetic controls, weakly aligned token
directions inherit the profile exponent, whereas strongly aligned low-rank token directions
decouple from it. On the measured battery, high-probability outcomes require singleton or
near-singleton resolution, whereas the remaining probability mass forms interchangeable clusters
with exact information budgets (Supplementary Note~\ref*{SI-si:anatomy}).

\begin{figure}[htbp]
  \centering
  \makebox[\linewidth]{\panel{a}{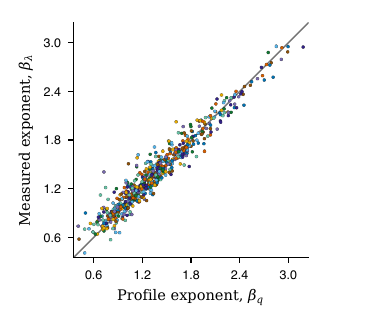}\hspace{\panelsep}\panel{b}{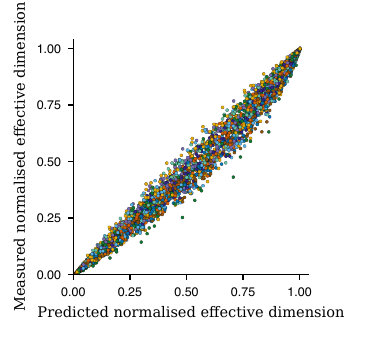}}\par\vspace{6pt}
  \makebox[\linewidth]{\panel{c}{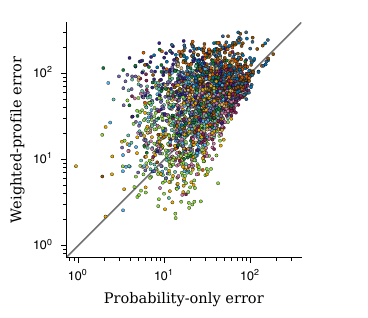}\hspace{\panelsep}\panel{d}{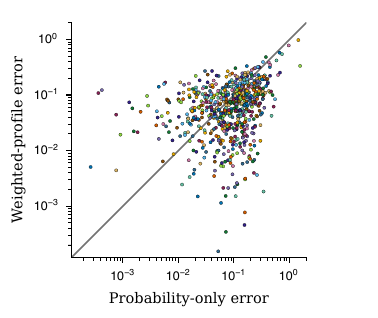}}\par\vspace{-3pt}
  \makebox[\linewidth]{\includegraphics{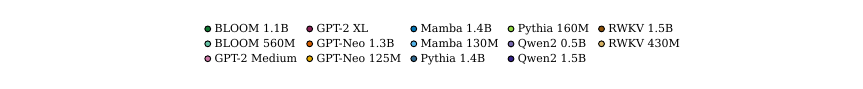}}
  \caption{\textbf{Language statistics shape the spectrum and its effective dimension.}
  \textbf{a}, Profile-predicted against measured spectral exponent over ranks 8--32 for all 576
  held-out model--context cells (nine models from five external families, 64 contexts per model).
  Small circles show every cell and the grey line is exact prediction. Colours identify models
  throughout the figure. The Pearson and Spearman correlations are 0.968 and 0.955; family-median
  absolute exponent errors are $0.0623$--$0.0856$. On a matched 16-context-per-model control battery,
  spectral errors are $4.89$--$9.22$-fold lower than for a same-trace flat spectrum and
  $13.55$--$25.63$-fold lower than after shuffling profile ranks.
  \textbf{b}, Prediction of effective dimension without fitted parameters, the number of modes resolved
  at a given damping, from token statistics. Small circles
  show all 5,760 model--context--resolution values from the same 576 cells, coloured by model; the
  grey diagonal is exact prediction. No
  measured eigenvalue or fitted parameter enters the prediction. Family-median
  effective-dimension RMSEs are $0.0270$--$0.0521$, $6.44$--$7.88$-fold lower than for the
  same-trace flat-spectrum prediction.
  \textbf{c}, Effective-dimension prediction in a matched 14-model, seven-family comparison on
  192 fresh contexts per model. All 2,688 points compare the weighted-profile and probability-only
  prediction errors on the same context, measured as effective-dimension RMSE across damping levels.
  The probability-only profile $\widehat\lambda_k=(r/V)p_{(k)}$ has lower family-median error in all seven families.
  \textbf{d}, All 672 paired spectral-exponent errors on the 48-context exact-spectrum subset per
  model; the weighted profile has lower family-median error in all seven families. Both panels use
  logarithmic axes, a common 27,450-outcome representation and the same standardised read-out;
  the grey diagonal marks equal error. Inference gives equal weight to families. Exact one-sided family
  sign-flip $P=1/128$ for each contrast (Supplementary Note~\ref*{SI-si:anatomy}).}
  \label{fig:imprint}
\end{figure}

The weighted profile also predicts held-out spectra without an eigendecomposition. On the held-out
battery in all five external families, the profile prediction reproduces the spectral profile and
effective-dimension curve and outperforms a same-trace flat spectrum and its effective-dimension
curve, as well as a rank-shuffled profile, in every family. It was fixed before exact spectra were
revealed for a disjoint 64-context battery in
each of nine models from five external families---BLOOM,
GPT-Neo, Mamba, Qwen2 and RWKV. Over ranks 8--32 it predicts the held-out eigenvalues with
family-median errors of $0.0617$--$0.0884$ decades, typical multiplicative deviations of
$1.15$--$1.23$-fold, and spectral-exponent errors of $0.0623$--$0.0856$. Applying the resolution
transform directly to the same fixed profile predicts the complete normalised
$N_{\mathrm{eff}}$ curve with family-median RMSEs of $0.0270$--$0.0521$
(Fig.~\ref{fig:imprint}a,b; Supplementary Table~\ref*{SI-si:tab:external-profile} and Supplementary Fig.~\ref*{SI-si:fig:profile-calibration}). On matched
controls, spectral error is
$4.89$--$9.22$-fold lower than for a same-trace flat spectrum and $13.55$--$25.63$-fold lower
than after shuffling profile ranks; effective-dimension error is $6.44$--$7.88$-fold lower than for the
effective-dimension curve induced by the same-trace flat spectrum. Exact inheritance certificates apply in
$98.4$--$100\%$ of external-family cells
(Methods; Supplementary Note~\ref*{SI-si:anatomy}).

Probability concentration and read-out geometry make distinct contributions to the spectrum.
In a matched comparison of 14 models from seven families on 192 fresh contexts, a probability-only
profile predicts effective dimension more accurately than the weighted profile in every family
(Fig.~\ref{fig:imprint}c). The weighted profile, which includes the learned read-out directions,
predicts the spectral exponent more accurately in every family (Fig.~\ref{fig:imprint}d).
Thus the concentration of probability captures the number of resolved modes, while read-out
structure improves prediction of how their strengths decay. This comparison uses a common outcome
representation; the native-vocabulary prediction in panels a and b provides the complementary
across-vocabulary test (Methods; Supplementary Note~\ref*{SI-si:anatomy}).

Across the external families, the token profile alone predicts effective dimension without fitted
parameters. In a separate finite-size confirmation, a bounded effective-dimension
curve fitted at four anchor dampings was evaluated on 240 interleaved held-out values and outperformed
an equal-parameter unbroken power curve on every one of 80 curves (Extended Data Fig.~\ref{fig:ed-effective-dimension}c,d).
Median held-out error was $0.00493$ for the bounded curve and $0.0814$ for the power curve.
The fixed-relative-damping conjugate-gradient bound is documented in Supplementary Note~\ref*{SI-si:instrument}.

\subsection*{Corpus n-gram statistics predict acquisition, and evidence depth shifts its timing}

\begin{figure}[htbp]
  \centering
  \makebox[\linewidth]{\panel{a}{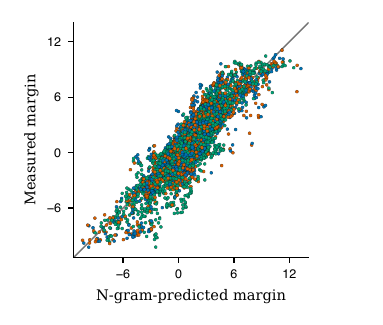}}\par\vspace{6pt}
  \makebox[\linewidth]{\panel{b}{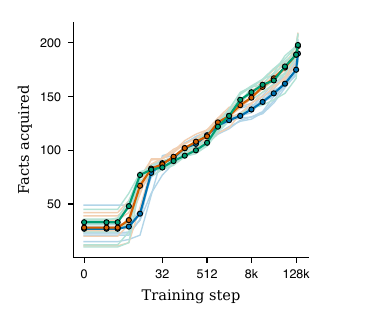}\hspace{\panelsep}\panel{c}{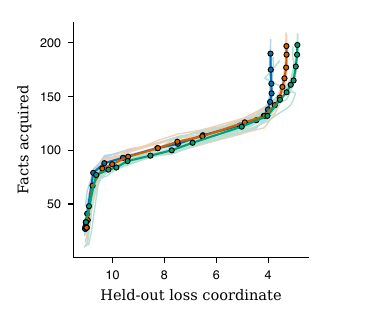}}\par\vspace{-3pt}
  \makebox[\linewidth]{\includegraphics{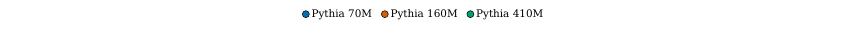}}\par\vspace{-3pt}
  \caption{\textbf{Corpus statistics predict acquisition across model sizes.}
  \textbf{a}, Ex-ante corpus n-gram margins predict subsequent margins without refitting for fresh facts
  disjoint from calibration at Pythia 70M, 160M and 410M. Points are all 9,840 fact--checkpoint
  means over 15 checkpoints: 5,340 from 178 defined facts at 70M and 160M (ten runs)
  and 4,500 from 300 facts at 410M (nine runs);
  the shared model-colour key below the panels distinguishes Pythia 70M, 160M and 410M; the grey
  diagonal is exact prediction. Trajectory $R^2$ is $0.790$, $0.792$ and $0.775$, persistent-status agreement is
  $0.792$, $0.854$ and $0.807$, and continuous timing error is $0.77$, $0.95$ and $0.96$ $\log_2$
  training-step units, respectively. All three timing errors outperform their constant acquisition-time
  and label-permutation baselines.
  \textbf{b}, Cumulative acquisition curves for the same 300 facts against training step at
  70M, 160M and 410M. Thin curves are
  individual seeds and heavy curves with circular markers are medians over nine seeds for each model
  size, using the shared model colours.
  \textbf{c}, The same curves against held-out loss. In the descriptive 70M-to-160M comparison,
  alignment by held-out loss has a median absolute error of 2.1 facts.}
  \label{fig:acq}
\end{figure}

Corpus $n$-gram margins separately predict acquisition timing (Fig.~\ref{fig:acq}). We measure
acquisition time on a logarithmic scale, $\tau=\log_2 t$, where $t$ is the first step after
which the designated answer's log-probability advantage over its paired alternative remains above a
fixed threshold.

Measured before training at unigram, bigram and trigram levels, these margins predict, without
refitting, the complete trajectories of fresh facts
disjoint from calibration at all three tested model sizes: trajectory $R^2=0.775$--$0.792$,
acquisition-status agreement $0.792$--$0.854$, and continuous timing error of
$0.77$--$0.96$ $\log_2$ training-step units, outperforming constant acquisition-time and label-permutation baselines
(Fig.~\ref{fig:acq}a). Evidence depth causally delays acquisition. In a controlled synthetic
language in which otherwise identical facts are randomly assigned to shallow or deep
evidence constructions
with paired initial weights and batch streams, no statistic below the assigned depth distinguishes
the two answers; the complete deciding margin appears at and above that depth. Across the analysed
10th--20th percentiles, acquisition-time quantiles under deep evidence were approximately $4.3$
times the corresponding quantiles under shallow evidence, a near-uniform shift of $2.10$ in
$\log_2$ training steps. This shift was stable under leave-one-seed-out analysis
(Extended Data Fig.~\ref{fig:ed-acq-generalisation}a; Supplementary Note~\ref*{SI-si:acquisition}). The paired randomisation identifies the causal effect. Depth therefore delays
persistent preference for the designated answer, but not the earlier onset of unsigned movement
along the same target--alternative margin direction (Supplementary Note~\ref*{SI-si:acquisition}). Conditional analyses of acquisition-time variation and
of smooth geometric motion are given in Supplementary Notes~\ref*{SI-si:evidence-flux} and~\ref*{SI-si:motion}, respectively.
The causal direction generalises beyond that construction. In a fully crossed randomised study
of GPT-2-, GPT-NeoX- and Llama-style decoders and two independent controlled languages, placing
otherwise matched evidence in the deeper construction delayed persistent acquisition in every
cell by $0.81$--$1.51$ $\log_2$ training-step units, equivalent to $1.75$--$2.86$-fold more training steps. All six
95\% intervals excluded zero, whereas all 12 fixed-treatment and label-randomisation control
intervals contained zero (Extended Data Fig.~\ref{fig:ed-acq-generalisation}b,c; Supplementary Note~\ref*{SI-si:acquisition}). Evidence
depth therefore delays acquisition across architectures; evidence construction modulates
the magnitude on average.

The cumulative number of acquired facts is shown across Pythia 70M, 160M and 410M against
both training-step and held-out-loss coordinates (Fig.~\ref{fig:acq}b,c). In the quantified
70M-to-160M comparison, aligning the cumulative curves by held-out loss gives a median absolute error
of 2.1 facts out of 300, about three times lower than alignment by training step.

The n-gram predictor was independently applied without refitting to the fresh 300-fact battery across
9 seeds and 20 checkpoints at 70M, 160M and 410M (Supplementary
Note~\ref*{SI-si:acquisition}).

\clearpage
\subsection*{The geometry determines minimum-disturbance interventions and their relative cost}

The same geometry determines how to change behaviour at minimum cost, and quantifies in advance the
penalty for ignoring it. Let $A=G+\alpha R\succ0$, with damping $\alpha$ and reference metric $R$. For
objective gradient $q\ne0$ and target change $q^\top\delta=b$, the unique minimum-disturbance step is
$\delta^\star=bA^{-1}q/(q^\top A^{-1}q)$, attaining the minimum regularised cost
$C_{A,\min}(b)=\tfrac12\delta^{\star\top}A\delta^\star=b^2/(2q^\top A^{-1}q)$.
Every other feasible step incurs an excess regularised cost
equal to half its squared $A$-distance from this optimum (Supplementary
Note~\ref*{SI-si:minimal-control}). To predict the realised
output disturbance at finite damping, define the $G$-cost per unit squared objective change as
$\Pi_G(u)=(u^\top Gu)/(q^\top u)^2$; smaller values mean less predicted local output change for the
same target effect. For the coordinate-Euclidean comparator used here ($R=I$), the prediction without
fitted parameters is
\begin{equation}
  R_{\mathrm{pred}}(q)
  = \frac{\Pi_G(q)}{\Pi_G(A^{-1}q)}.
  \label{eq:control}
\end{equation}
This finite-damping cost ratio was fixed before intervention; the idealised ratio in the full
regularised metric is analysed in Supplementary Note~\ref*{SI-si:minimal-control}.

\begin{figure}[!t]
  \centering
  \makebox[\linewidth]{\panel{a}{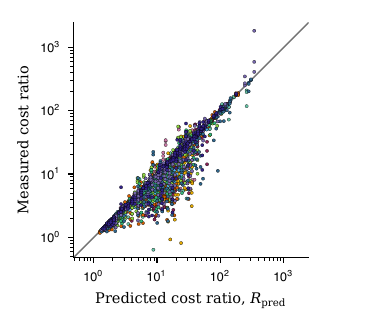}\hspace{\panelsep}\panel{b}{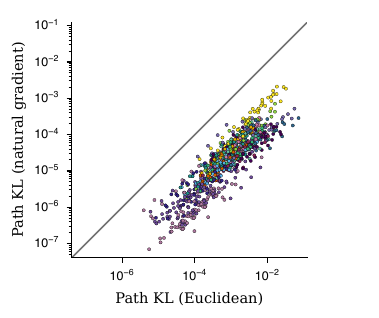}}\par\vspace{6pt}
  \makebox[\linewidth]{\panel{c}{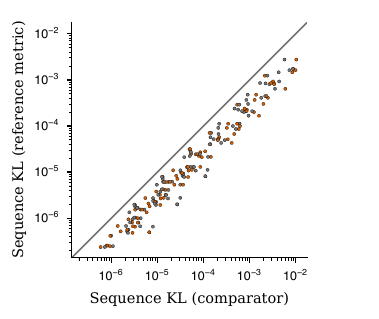}\hspace{\panelsep}\panel{d}{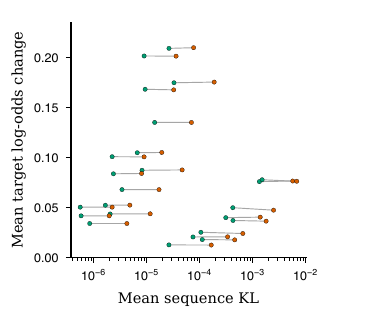}}\par\vspace{-3pt}
  \makebox[\linewidth]{\includegraphics{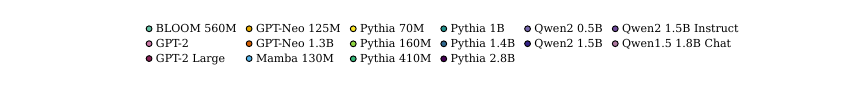}}\par\vspace{-2pt}
  \makebox[\linewidth]{\includegraphics{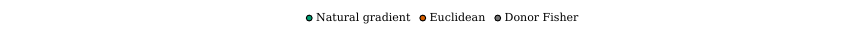}}
  \caption{\textbf{The geometry predicts intervention cost and supports reusable control.}
  \textbf{a}, At matched behaviour change, the measured ratio of off-target output divergence for
  Euclidean versus damped natural-gradient updates is compared with the finite-damping prediction
  $R_{\mathrm{pred}}$, which has no fitted parameters and is computed before intervention. The axes
  are logarithmic and the grey identity line denotes perfect agreement. Small circles show all
  3,515 usable matched-response observations
  (of 3,564) from 1,188 model--objective--layer--context cells: 11 models, six families, three
  objectives, three relative layer depths and 12 contexts per objective, each measured at three
  intervention magnitudes (98.6\% coverage; median measured-to-predicted ratio $0.967$, 95\%
  interval $0.957$--$0.975$; log--log slope $0.982$, 95\% interval $0.966$--$0.997$).
  \textbf{b}, Finite paths with both methods relinearised after every accepted step. All 864 points
  compare Euclidean and Fisher cumulative local KL at matched target changes, from 216 prompts,
  16 models and 18 model--objective cells. The metric sums KL between successive accepted outputs.
  The 72 cell--target geometric-mean cost ratios span $11.5$--$127.1$, with every paired 95\%
  interval above one. Colours identify models in panels a and b.
  \textbf{c}, Reusable control on four base models from three families. All 192 points compare
  reference-Fisher sequence KL with Euclidean (vermilion) or donor-only Fisher (grey) sequence KL
  for the same reference continuation and matched donor change, across two objectives and three
  target changes. Each setting contributes four reference continuations per comparator.
  Reference sequences are teacher-forced continuations of the four prompts used to construct
  the reference metric. Ratios of mean Euclidean/reference-Fisher sequence KL span $2.90$--$6.29$.
  \textbf{d}, The same updates change unseen target prompts without target gradients.
  Each point shows mean sequence KL over four reference continuations and mean target log-odds
  change (nats) over eight unseen prompts, for one model, objective, donor change and method
  (48 points). Green, natural gradient using the reference metric; vermilion, Euclidean.
  Grey segments join matched donor changes. All 24 natural-gradient mean target changes are positive.
  Grey diagonals in panels b and c mark equality.}
  \label{fig:price}
\end{figure}

The cost ratio is predicted without fitted parameters. Across 11 models from six architecture
families, three steering objectives and three relative layer depths, the finite-damping ratio
computed before intervention tracks 3,515 usable matched-response measurements from 1,188 cells
(98.6\% coverage; median measured-to-predicted ratio $0.967$, 95\% interval
$0.957$--$0.975$; log--log slope $0.982$, 95\% interval $0.966$--$0.997$;
Fig.~\ref{fig:price}a).
The advantage persists along finite intervention paths. Recomputing both Fisher and Euclidean
directions after each accepted step reaches four matched objective changes across 216 prompts
and 18 model--objective cells. The Fisher paths accumulate about $11$--$127$ times less local KL
than their Euclidean counterparts, with positive paired intervals in every cell and target
(Fig.~\ref{fig:price}b; Supplementary Note~\ref*{SI-si:minimal-control}).

The metric also supports interventions that can be reused across prompts. A shared activation
update is learned from four donor prompts per objective, using the Fisher metric averaged over
four reference prompts to limit disturbance. The resulting update is then applied to eight unseen
target prompts without computing target gradients. Across BLOOM-560M, Pythia-410M, Pythia-1.4B and
Qwen2-1.5B, it transfers a positive mean truth-preference or anti-sycophancy change in all 24
model--objective--target settings (Fig.~\ref{fig:price}d). At matched donor changes, the reference
metric reduces sequence KL on reference continuations by about $3$--$6$ times relative to Euclidean
control and by at least $2$ times relative to a donor-only Fisher metric
(Fig.~\ref{fig:price}c). A joint solve also transfers both objectives with lower reference-sequence
cost in all four models (Supplementary Note~\ref*{SI-si:minimal-control}).

Prompt-specific control extends to instruction-tuned models: at equal aggregate mean objective
change, natural-gradient frontiers have $8$--$74$ times lower off-target divergence for
anti-sycophancy, $30$--$190$ times for capital-city truth preferences and $4$--$550$ times for
style (Supplementary Note~\ref*{SI-si:instruct}; Extended Data Fig.~\ref{fig:ed-intervention-frontiers}a). Several objectives combine through their
weighted gradient sum, with lower off-target cost across the measured composition frontier
(Extended Data Fig.~\ref{fig:ed-intervention-frontiers}b).

\clearpage
\subsection*{One geometric correction spans model operations}

A single correction, replacing the Euclidean inner product with the pullback metric of the
relevant output map, improves every matched endpoint in Table~\ref{tab:operations}. Its prediction
without fitted parameters generalises across the 11-model, six-family panel in Fig.~\ref{fig:price}a,
and the reference metric supports reusable control across unseen prompts.
Applied to a single fact, the steer becomes a training-free editor: on 30
held-out CounterFact records (GPT-2), with each edit solved once and reused through a
shared prompt-selection rule, the pullback-metric edit had $10.57$-fold lower geometric-mean
other-token divergence than its matched Euclidean counterpart (95\% interval $8.52$--$13.37$);
both operators attained the fixed $+5$ log-odds calibration shift on all 30 edits (Extended Data
Fig.~\ref{fig:ed-intervention-frontiers}c). Separate GPT-2-XL
evaluations cover 100 edits under standard metrics and 300 under a stricter probability-margin
criterion (Supplementary Table~\ref*{SI-tab:si-editing}; Supplementary Note~\ref*{SI-si:editing}). For interpretability, Fisher-weighted output sensitivity
$v^\top Gv$ measures the output change caused by a feature direction; activation magnitude measures
how strongly it activates. Across 303 zero ablations of sparse-autoencoder (SAE) features at Pythia-410M, each setting one active feature to
zero~\cite{topksae,cunningham2023sparse,bricken2023monosemanticity}, the intervention-specific Fisher
cost rank-correlated with the exact divergence at $\rho=0.997$, against $0.896$ for activation
magnitude. The objective-effect-to-Fisher-cost ratio, the squared linear objective effect per unit Fisher cost,
predicted the selected-token probability-change fraction at $0.935$ against $0.506$ for attribution
patching~\cite{attributionpatching}, which remains the better predictor of raw effect
magnitude (Extended Data Fig.~\ref{fig:ed-feature-selectivity}; Supplementary Note~\ref*{SI-si:codezero}).

The correction carries over to training. The code-Gram metric, the second moment of sparse feature
activations, increases Hungarian-matched mean absolute decoder--ground-truth cosine from $0.167$
to $0.480$ ($+0.313$) when used
to precondition a sparse-autoencoder decoder at 410M with planted-signal/background-RMS ratio $\beta=0.5$
(Supplementary Note~\ref*{SI-si:dictlearn}). At the same
scale, natural-gradient low-rank adaptation produces $60$--$250$-fold less off-target change than
Adam at matched on-target change. In a separate held-out comparison at 70M, it improves the
preservation-KL frontier $1.57$-fold relative to tuned AdamW with explicit KL regularisation
(Extended Data Fig.~\ref{fig:ed-lora-preservation}; Supplementary Note~\ref*{SI-si:instrument}).

The size of the available correction is itself set by the geometry. The advantage is largest at
early-to-mid injection layers, where the metric pulled back through the rest of the network is
most anisotropic, and shrinks toward the near-linear read-out (Extended Data Fig.~\ref{fig:ed-layer-magnitude}a). It vanishes
in the isotropic limit, where the natural-gradient and Euclidean steps coincide. Across the same
11-model panel, every measured-to-predicted path decreases over the three declared intervention
fractions; the model-balanced geometric means are $0.918$, $0.860$ and $0.755$ (Extended Data
Fig.~\ref{fig:ed-layer-magnitude}b). This quantifies attenuation of the local prediction as intervention magnitude grows;
Supplementary Note~\ref*{SI-si:minimal-control} analyses the finite-magnitude correction to cubic order. Output sensitivity, residual cross-model sharing and answer-associated
probability motion concentrate in the same leading spectral modes (Extended Data Fig.~\ref{fig:ed-colocalisation}), and
cyclic concepts trace nearly closed rotations, whereas analogy relations instead share a
probability-displacement direction (Extended Data Fig.~\ref{fig:cyclic}).

\clearpage
\begin{table}[t]
  \centering
  \caption{\textbf{One correction spans the evaluated operations.}
  The result column reports the strongest reported matched comparison for each operation; the
  final column gives its evidence base. Intervals are 95\% bootstrap
  intervals. AUC, area under the curve; KL, Kullback--Leibler; LoRA, low-rank adaptation.
  Sources and protocols are given in Methods.}
  \label{tab:operations}
  \footnotesize
  \setlength{\tabcolsep}{3pt}
  \begin{tabular}{@{}>{\raggedright\arraybackslash}p{2.15cm}
                       >{\raggedright\arraybackslash}p{3.35cm}
                       >{\raggedright\arraybackslash}p{5.15cm}
                       >{\raggedright\arraybackslash}p{4.15cm}@{}}
    \toprule
    Operation & Geometry-aware construction & Principal result & Evidence base \\
    \midrule
    Steering (non-local) & per-prompt natural gradient $(G{+}\alpha I)^{-1}q$ &
      prediction without fitted parameters tracks the realised advantage across 11 models (median
      measured/predicted $0.967$, 95\% interval $0.957$--$0.975$; slope $0.982$, 95\% interval
      $0.966$--$0.997$); up to $72.7$-fold lower off-target KL
      on factual steering, with $92$--$100\%$ prompt-level wins & 3,515 usable observations from
      1,188 cells; 11 models, six families, three objectives and three relative depths; six-size Pythia factual sweep and
      two instruction-tuned Qwen frontiers \\
    Knowledge editing & fixed Fisher edit with a shared gating rule &
      geometric-mean Euclidean/Fisher other-token KL ratio $10.57$
      (interval $8.52$--$13.37$); fixed $+5$ log-odds shift on 30/30 edits & GPT-2, 30 held-out matched
      CounterFact edits; supporting GPT-2-XL evaluations on 100 standard-metric and 300
      probability-margin edits \\
    Feature importance & intervention-specific Fisher cost $\tfrac12\Delta h^{\top}G\Delta h$ &
      $\rho=0.997$ with exact feature-ablation KL, versus $0.896$ for activation magnitude &
      Pythia-410M; 57 evaluation-active features from a layer-12 Top-$k$ sparse autoencoder;
      18 held-out context clusters and 303 interventions \\
    Attribution & objective-effect-to-Fisher-cost ratio &
      selected-token probability-change fraction $\rho=0.935$ versus $0.506$ for attribution patching; sign agreement
      $0.746$ versus $0.354$ & same 303-intervention Pythia-410M feature-ablation evaluation;
      selected-token probability-change-fraction and sign-agreement endpoints \\
    Dictionary learning & code-Gram natural gradient for the decoder &
      matched absolute-cosine recovery increases $0.167\to0.480$ ($+0.313$); 3/3 seeds &
      planted-feature recovery on real Pythia-410M activations at planted-signal/background-RMS ratio $\beta=0.5$;
      three paired seeds \\
    Fine-tuning & exact natural-gradient LoRA &
      tuned AdamW+KL incurs $1.57\times$ the preservation-KL frontier AUC
      (interval $1.14$--$1.98$); $60$--$250$-fold lower off-target change than Adam at 410M &
      held-out 70M rank-4 frontier: three target gains, 8 seeds and 16 prompts;
      supporting 410M rank-16 matched-target sweep \\
    \bottomrule
  \end{tabular}
\end{table}

%% file: sections/discussion.tex
\section*{Discussion}

\looseness=-1
Next-token training induces a canonical output geometry. Predictive behaviour identifies the
language-determined read-out subspace up to transformations that leave outputs unchanged; this
predicts convergence of independently trained models estimating the same language. Controlled
language assignment supports this origin: changing the law changes the learned geometry, while
different architectures fitted to the same law recover nearly identical geometry. Activation-space
convergence measures, by contrast, must tolerate arbitrary reparameterisations. Among those
evaluated here, agreement is higher in output geometry and reparameterisation-invariant; it remains
high after permutation-null calibration for scale-dependent
baselines~\cite{groger2026aristotelian}.

Their measured agreement is reconstructed exactly from six variance and covariance terms, and its
shared component carries the measured semantic alignment. A semantic-category probe transfers across
models, and the comparison extends to human predictions. Models with lower prediction error agree
more closely with human word choices across families. On independent data, agreement also increases
with scale and training, while a calibration learned only from model laws improves predictions across
the full coarse distribution of human word choices. Direct probability-space comparisons on a
separate human corpus also recover leading directional structure.
These completion-distribution comparisons complement work on the finite resolution of human cloze
estimates and the semantic distinctions captured by model probabilities~\cite{nair2026clozing}.

Token statistics predict the spectrum and its effective dimension, the number of modes resolved
at a given damping, without fitted parameters. The matched mechanism comparison separates two
contributions: probability concentration captures effective dimension, while the weighted read-out
profile improves prediction of spectral shape across all seven families. At fixed relative damping and
solver tolerance, the damped natural-gradient solve has a proved worst-case conjugate-gradient
iteration bound independent of model width; at $c=10^{-2}$ and Euclidean relative residual $10^{-6}$,
the exact-arithmetic bound is 85 iterations.

Separately, corpus n-gram evidence measured before training predicts individual-fact acquisition with
median absolute errors of $0.77$--$0.96$ on the $\log_2$ training-step scale at all three sizes. In the
controlled languages, randomised evidence depth causally delays acquisition.

The same geometry prescribes minimum-disturbance control whose advantage is predicted in
advance by a measured anisotropy ratio. This advantage persists along relinearised finite paths
and extends to updates reused across prompts. Its recurrence across comparison, steering, editing,
feature analysis and training shows that the correction is not confined to a single evaluated
operation. Supplementary Table~\ref*{SI-si:tab:evidence-map} maps these claims to their analytical,
observational and interventional evidence.

Alignment systems face two separate questions: what behaviour should change, and how to make that
change without unnecessarily disturbing the model elsewhere. Human preferences, constitutions,
reward or cost models and other forms of oversight must answer the first question. The output geometry
addresses the second. Given a preference covector and a desired first-order change, the regularised
Fisher pullback gives the activation change with minimum local output disturbance. It can therefore
turn an externally supplied preference into an intervention, complementing existing
representation-level methods for inference-time alignment~\cite{iti,caa,repe}. On instruction-tuned
models, the construction improves the aggregate trade-off between objective change and off-target
change for anti-sycophancy, capital-city truth preferences and writing style. Averaging the metric
over a reference prompt set extends this construction to reusable control. Updates learned from
four donor prompts transfer to eight unseen prompts per objective in four base models, while
reducing reference-sequence change by about three- to sixfold relative to Euclidean control at
matched donor effects. The solve is performed on donors and reference prompts; the resulting
activation update is applied to each target without computing its gradient. Several objectives
also combine in one solve, allowing their balance to be set through externally supplied weights.

The same idea may be useful during post-training. Rather than keeping an update close to the original
model in parameter or activation space, the reference-set construction could be applied in parameter space to limit changes to the
model's predictions on protected prompts. This would give preference
fine-tuning and model customisation a behaviour-level trust region. The held-out LoRA comparison
provides an initial local-preservation result in this direction, although it does not establish
safety-specific retention or broader capability preservation.

Mechanistic interpretability has a related measurement problem. A feature can activate strongly or
admit a clear verbal description without playing an important role in the model's output. Conversely,
an intervention can change a chosen target only by disrupting many other predictions. Fisher-weighted
output sensitivity measures the predicted local output consequence of a specified feature intervention,
while the objective-effect-to-Fisher-cost ratio measures its squared linear effect on a chosen objective per
unit Fisher cost. In the sparse-feature
experiments, these quantities accurately predicted both the output cost of removing a feature and the
selectivity of its effect. They could therefore help prioritise candidate features and circuits in
audits of sycophancy, deception, refusal and other safety-relevant behaviours, and help choose
interventions with a limited behavioural footprint. End-to-end sparse dictionary learning already
prioritises functional importance by minimising the KL divergence between original and reconstructed
model outputs~\cite{braun2024endtoend}. The present sensitivity measurements motivate investigating
local geometric weighting for sparse dictionaries and replacement models, so that limited explanatory
capacity is spent on distinctions that matter to the model's behaviour. These
uses would complement semantic labelling and causal attribution: the metric measures the functional
consequence of a proposed feature, but does not by itself explain what that feature represents.

The comparison results suggest a further use in model auditing. Because relational output geometry
does not require corresponding hidden coordinates, architectures or tokenizers, a fixed battery of
prompts and anchors could be used to compare successive checkpoints, post-training variants and
independently developed models. The consensus--residual decomposition could then separate widely
shared structure from model-specific deviations. The acquisition results could likewise help place
costly evaluations near the predicted learning transitions of target facts or associations,
while routed geometric edits could provide modular factual corrections with a controlled output
footprint. Together, these applications suggest a role for output geometry throughout the model
lifecycle: anticipating behavioural change during training, auditing it across models, interpreting
its internal causes and controlling it when intervention is needed.

The same information-geometric principles have a
direct analogue in quantum computing: on pure quantum
states, the Fubini--Study metric is, up to a constant, the Fisher--Rao metric naturally induced by
optimal quantum measurements~\cite{braunstein1994statistical}. Its pullback underlies quantum
natural-gradient optimisation and variational dynamics, including in the author's recent
work~\cite{picozzi2026metric}.

Across the models and operations tested here, predictive behaviour identifies a canonical output
geometry shared across models and related to human predictions. Token statistics shape its
resolved structure, corpus n-gram evidence predicts when individual facts are acquired, and the
metric supplies calibrated control that can be reused across prompts.

%% file: methods.tex
\section*{Methods}

\subsection*{Models, datasets and geometric definitions}

\paragraph{Models and data.}
Pretrained-model analyses use public checkpoints; causal language-assignment and acquisition
experiments train controlled synthetic systems.
The language-model ladder is the Pythia suite~\cite{pythia} (70M--6.9B).
Alignment-objective experiments use Qwen2-1.5B-Instruct~\cite{qwen2}, replicated on
Qwen1.5-1.8B-Chat~\cite{qwen15}, and knowledge editing uses GPT-2 and GPT-2-XL~\cite{gpt2}. The
spectral analysis of the steering dimension additionally spans GPT-2, GPT-2-XL, Qwen2-1.5B and
BLOOM-1b1~\cite{bloom}, whose vocabularies reach $\sim$250k tokens. The cross-model and dense-response
batteries additionally include GPT-Neo, Mistral-7B, Mamba, RWKV and StarCoder2
checkpoints~\cite{gptneo,mistral7b,mamba,rwkv,starcoder2}. Unless stated otherwise, interventions act on the
residual stream at a middle layer, $L = \mathrm{round}(0.45\times\text{depth})$ (Qwen2-1.5B-Instruct uses layer~14); the feature-importance analyses use layer~12 of Pythia-410M.

Layer-dependence is reported in Extended Data. Objectives and concepts are: morphological and
sentiment (SST-2~\cite{sst2}) contrasts; factual capital-city true/false preference pairs; a cyclic
registry (weekdays, months); sycophancy, truthfulness (TruthfulQA~\cite{truthfulqa} and capital-city
true/false pairs) and writing style in experiments on instruction-tuned models; and
CounterFact~\cite{rome} for knowledge editing. The feature-importance analyses use a trained
Top-$k$ sparse autoencoder~\cite{topksae}
(\nolinkurl{tim-lawson/sae-pythia-410m-deduped-x64-k32-layers-12}; residual-stream Top-$k$, $k{=}32$,
layer~12), which retains the $k$ largest feature activations per token.

\paragraph{Output metric and pullback.}
Write the next-token distribution as $p = \mathrm{softmax}(W_U h_L + b_U)$, where
$h_L\in\mathbb{R}^d$ is the final pre-unembedding activation and
$W_U\in\mathbb{R}^{|V|\times d}$ is the unembedding over a vocabulary of size $|V|$. Denote the
rows of $W_U$ by $w_a$. Pulling the
Fisher--Rao metric of $p$ back to $h_L$ through the softmax gives the \emph{output Fisher}
$H = \mathrm{Cov}_p(w) = W_U^\top(\mathrm{diag}(p) - p p^\top)\,W_U$.
For an intervention on an activation $h$ at layer $\ell$, let $J = \partial h_L/\partial h$ be
the (fixed-context) Jacobian of the network from $\ell$ to the read-out; the (possibly degenerate) pullback Fisher metric is
$G = J^\top H J$, reducing to $H$ at the final layer. $G$ is the second-order model of output
change, $\mathrm{KL}(p \Vert p_{\delta h}) = \tfrac12 \delta h^\top G\, \delta h + O(\|\delta h\|^3)$.

For an objective covector $q=\nabla_h\phi$, the regularised natural-gradient step is
$\delta h\propto(G+\alpha R)^{-1}q$, with positive reference metric $R$.

Under $h \mapsto Ah$, $G \mapsto A^{-\top} G A^{-1}$. The same damped problem is coordinate-covariant
when $R\mapsto A^{-\top}RA^{-1}$. Native-coordinate experiments use $R=I$;
canonicity, coordinate covariance and the numerical coordinate test are detailed in
Supplementary Note~\ref*{SI-si:theory}.

\subsection*{Behavioural identification and causal language assignment}

\paragraph{Profiled-loss experiment.}
The identification test uses a rank-32 resolved reference family estimated from Pythia-6.9B on
500 calibration contexts. For Pythia 70M, 160M, 410M and 1.4B, candidate subspaces follow paths
towards each model's read-out and random controls preserving the same principal angles.
At five positions along each path, the intercept and context coordinates are jointly optimised
on each of two disjoint 100-context calibration folds. Each intercept is then fixed and only
coordinates are optimised on 100 held-out contexts, disjoint from reference construction.
The mean of the two held-out profiled KL losses is compared with squared chordal distance;
Figure~\ref{fig:ident}a displays both fold estimates. Profiling uses float64 and explicit
stationarity tolerances (Supplementary Note~\ref*{SI-si:identification}).

\paragraph{Causal language assignment.}
Three model architectures (transformer, gated recurrence and diagonal recurrence), each at two
capacities, were trained on paired synthetic languages with 64 contexts and 64 outcomes. Training
schedules were selected using two separate pilot language pairs and conditional prediction risk,
then applied to all eight test pairs. The risk requirement was excess cross-entropy below
$\Delta^2/32$, where $\Delta^2$ is the mean squared separation between the two known language
geometries. Geometry was represented by scaled root-probability chord distances among contexts.
The same architecture was compared across laws, and different architectures were compared within
a law at the same capacity. Both differences were normalised by $\Delta^2$. Inference resampled
whole language pairs, retaining all architecture and capacity comparisons within each pair
(Supplementary Note~\ref*{SI-si:identification}).

\subsection*{Cross-model geometry and human completions}

\paragraph{Relational comparison and controls.}
For a common context battery,
each model defines a matrix of Fisher--Rao distances
$d_{ij} = 2\arccos\sum_a \sqrt{p_i(a)\,p_j(a)}$ between its next-token distributions. Two models
are compared by the Spearman correlation of the upper triangles of their matrices (rank
agreement). This is an instance of representational similarity analysis~\cite{kriegeskorte2008rsa}
that requires no shared vocabulary or architecture. Activation geometries are
compared identically using cosine distances of mid- or last-layer states. Three batteries are
used: 200 natural sentence prefixes drawn from the held-out WikiText test split~\cite{wikitext}, 160 templated contexts spanning
eight topics, and 120 factual-relation prompts whose next token is the answer (the natural battery is primary).

The stability analysis uses the per-context Hellinger bound and the excess-risk rank-convergence
theorem (Supplementary Note~\ref*{SI-si:stabatten}). It is carried out on the Bhattacharyya affinity matrices
$C_{ij} = \sum_a \sqrt{p_i(a)\,p_j(a)}$, whose entries determine the corresponding
distance-matrix ranks. The realised-perturbation certificate uses ten same-tokenizer pairs so that per-context divergences are defined on
a shared outcome space, with vocabulary positions beyond the common range merged into a single
bucket. The population excess risks to the unknown training conditional law are not estimated.
The common-byte comparison below provides a cross-tokenizer extension; native-vocabulary
agreement is reported separately (Supplementary Note~\ref*{SI-si:convergence}).

Surface and predictability controls partial out, respectively, a character $n$-gram
geometry of the contexts and the geometry of a corpus $n$-gram predictor with backoff; the
tokenizer control compares same-tokenizer with different-tokenizer model pairs; the corpus control
contrasts reference models that share a target pair's training corpus with reference models that do not, at
matched scale and tokenizer.

In the reparameterisation experiment, each model's hidden states are transformed by
an independent random invertible map of condition number $\kappa$ before activation-based
similarity is computed. Output-geometry agreement is unchanged because the model's input--output
function is fixed, verified numerically.

Reference mediation conditions the agreement of every target-model pair on a reference model's
geometry, sweeping references from 70M to 7B parameters. Unembedding geometries are cosine
similarities of unembedding rows over a vocabulary of words that are single tokens in every model,
after subtracting the vocabulary mean~\cite{mu2018allbutthetop}; covariance-based quantities
(the output Fisher and its eigenvectors) are invariant to that correction and are used uncorrected.

\paragraph{Common-byte comparison.}
Each tokenizer's next-token law is mapped to the distribution of the first byte of the remaining
text, with byte classes providing a second, coarser common outcome space. Byte-level round trips
and non-decreasing Bhattacharyya affinity under both coarse-grainings are checked over all
28 model pairs (Supplementary Note~\ref*{SI-si:stabatten}).

\paragraph{Prediction of cross-model agreement.}
The six-scalar accounting uses a fixed leave-pair-out reference over the context battery; all
variances and covariances are population (ddof $=0$) statistics over a declared pair population,
strict upper triangle for the relational accounting and all ordered pairs including the diagonal
where the theory's independent-draw convention requires it. Risk floors and chord-based checks
apply only where the reference is a genuine conditional law on the same outcome experiment.
The cross-fitted prediction pools fluctuation energy and alignment on a training half of the
contexts, estimates the shared-fluctuation fraction for each target pair from the other pairs,
and predicts the held-out-half agreement with no test-half error statistic entering any
estimate. Predictions cover six ordered pairs of four independently trained runs in both
directions at ten public checkpoints (120 predictions). The estimator is undefined where its
variance denominator is non-positive, which occurs in 24 early-checkpoint cells (steps
$\leq 64$), leaving 96 defined predictions (Supplementary Note~\ref*{SI-si:stabatten}).

\paragraph{Semantic geometry and transfer.}
The consensus geometry is the mean of the rank-transformed distance matrices across models; a
model's residual is its rank-transformed matrix with the consensus partialled out. The external
semantic geometry is the mean-centred cosine geometry of the fixed \texttt{all-MiniLM-L6-v2}
Sentence-BERT encoder~\cite{sbert} over the same contexts. Transfer probes represent each prompt by its Fisher--Rao distances to a fixed
anchor set (relative representations~\cite{moschella2023relative}), $z$-scored per model; a
multinomial logistic probe is trained on one model's representation and evaluated on held-out
prompts through another's, with five-fold cross-validation.

The cross-modal control compares the mean cosine-distance matrix of last-token, last-hidden-state
concept representations from Pythia-1.4B and GPT-2 Large with a concept dissimilarity matrix formed
by averaging DINOv2~\cite{dinov2} features over CIFAR-100~\cite{cifar100} images. This is an activation-space
representational analysis, separate from the output Fisher--Rao analyses.

\paragraph{Eigenvector sharing.}
The eigenstructure analyses band the
output-Fisher spectrum by eigenvalue rank; each band reports the eigenvalue mass, the sharing of
eigenvector structure across model pairs measured as the subspace overlap of eigenvector-induced
probability displacements over the common vocabulary, and the fraction of the band's probability
motion on the highest-probability tokens. Sharing is always reported against a displacement-magnitude-matched
null that preserves each word's displacement magnitude while randomising directions. Most raw
eigenvector overlap reflects similar token-probability profiles; subtracting the null isolates
sharing beyond this displacement-magnitude effect.

\paragraph{Native human-completion comparison.}
Human sentence-final completions were drawn from the norms of Peelle et al.~\cite{peelle2020norms}.
Native sampling used 512 contexts, six Pythia sizes from 70M to 2.8B, and 32 categorical samples per
context at temperature one, without top-$k$ or nucleus filtering. Generation stopped at eight
new tokens; a fixed parser extracted the completion through the first punctuation, newline or
end-of-sequence boundary. Empty and length-limited samples were retained. Each completion was
embedded with the fixed \texttt{all-MiniLM-L6-v2} encoder and mapped through a multiscale Gaussian
kernel approximation with three bandwidths and 768 random Fourier features. Model and human
sample means were compared by squared feature distance and centred kernel alignment (CKA).
Human participants and model samples were split into independent halves; cross-pair values were
averaged and CKA was corrected using human and model split-half reliability. Intervals and
log-parameter slopes used 4,000 paired bootstrap draws over 16 disjoint context folds.

\paragraph{Prediction of human alignment.}
The predictive-fit comparison used a separate 384-context subset, the same semantic features,
six final Pythia sizes, five Pythia-160M checkpoints, five OLMo-2-1B checkpoints~\cite{olmo2}, and
five external models. Here each model law was conditional on complete human-observed
completion-plus-end-of-sequence events, retaining tokenizer collisions. Conditional root-probability
risk predicted reliability-corrected human alignment through a linear relation fitted only to the
Pythia sizes and trajectory, separately within each context fold. This relation was applied to OLMo
and external models without refitting.

\paragraph{Human probability directions.}
A separate next-word corpus~\cite{devarda2024cloze}
provided 1,726 contexts in 205 sentences for comparison of leading log-probability directions
across seven families. Model and human laws were mapped to 64 fixed semantic bins plus a residual
outcome. Whole-sentence splits separated human reference, model direction estimation, calibration
and evaluation; participant splits assessed respondent generalisation (Supplementary Note~\ref*{SI-si:meaning-protocols}).

\paragraph{Independent human-completion replication.}
The DERCo resource~\cite{quach2024derco} supplied word-prediction responses for five narratives.
Before viewing response values, 128 positions per narrative were selected from stimulus metadata by
a fixed hash, giving 640 contexts, 64,000 responses and 499 participant identities. Models received
the preceding ten displayed words. For six final Pythia sizes and five ordered checkpoints each from
Pythia-160M and OLMo-2-1B, 32 native continuations per context were divided between two independently
seeded halves. The first lexical word was embedded with the same fixed semantic encoder and kernel
map as above. Squared distances averaged the four pairings of participant and model-sample halves,
with equal weight for contexts and stories. The 512 bootstrap draws resampled participants globally,
sentences within each fixed story and native sequences; intervals are conditional on the five stories.

\paragraph{Model-only calibration on independent human data.}
The same contexts were mapped to 64 fixed semantic bins plus a residual outcome. For each of seven
model families, a full 65-dimensional affine ridge map in centred log-probability coordinates was
trained to predict leave-tokenizer-cluster-out source-model laws from candidate-model laws on the
earlier corpus. No human response entered the fit or model selection. Four source-only rotations were
averaged equally, and the complete recipe was fixed before the DERCo responses were examined. The
primary comparison was the change from the native model law in expected multinomial log score against
unsmoothed human counts, weighting contexts equally within stories and the five stories equally.
These outcomes test the complete coarse distribution, not recovery of individual word probabilities.

\subsection*{Spectral measurements and effective-dimension prediction}

\paragraph{Spectrum estimation and inheritance.}
Output-Fisher spectra are computed by exact diagonalisation on a renormalised top-$512$ token
support. The target support is the smallest token set containing $0.99$ of the
probability mass. No evaluated context reaches this target within 512 tokens, and $94\%$ still do
not reach it within 1024 tokens; the retained probability mass at 512 tokens averages $0.90$.
A full-vocabulary calculation reproduces the rank-$2$--$80$ core fit (Supplementary Note~\ref*{SI-si:anatomy}).

Power-law exponents $\beta$ and $R^2$ are fitted as log--log slopes over ranks $2$--$80$, and high-sensitivity mode counts are
$\#\{\lambda_i > c\,\lambda_{\max}\}$ at $c = 10^{-1}, 10^{-2}, 10^{-3}$. The empirical inheritance
test compares, per context, the sorted eigenvalues of $H$ with the sorted values of
$p_a \lVert w_a\rVert^2$ (fitted on the same rank window), and the Zipf exponent $s$ of the sorted
profile $p$; this is the uncentred empirical exponent test. The theorem-level analysis uses the
exact centred profile $q_a=p_a\lVert w_a-\sum_b p_bw_b\rVert^2$ over the full vocabulary,
measures its frame and tail constants, checks the conditional eigenvalue bounds, and reports the
realised $\lambda_k/q_{(k)}$ band. Its profile-only $k^{*}_{\mathrm{prof}}$ is evaluated as an
empirical crossover predictor, separately from the conditional trace-floor theorem.
Eigenvector inheritance is scored
as the alignment between the $i$-th eigenvector and
the unit read-out direction of the token with the $i$-th largest $p_a\lVert w_a\rVert^2$.
Synthetic read-out controls and spectral model-selection/coverage checks are detailed in
Supplementary Note~\ref*{SI-si:anatomy}; exponent claims concern the declared core windows.

\paragraph{Standardised read-out representation.}
We use the standardised centred read-out representation
$U=\operatorname{orth}(P_0W_U)$, an orthonormal basis for the centred read-out subspace, where
$P_0=I-|V|^{-1}\mathbf{1}\mathbf{1}^{\top}$ removes the output-softmax gauge. Writing $u_a$ for
row $a$ of $U$ and $\bar u=\sum_a p_a u_a$, each model--context cell has centred weighted profile
$q_a=p_a\lVert u_a-\bar u\rVert^2$. The exact trace identity between the per-context output Fisher and its weighted profile is
checked in every context. Frame certificates are evaluated at
$b=1/2$, $k=64$ on the fixed 56-curve battery, in both the standardised basis and native read-out
coordinates.

\paragraph{External-family prediction.}
The profile-to-spectrum-to-effective-dimension test uses design
choices fixed with Pythia and GPT-2, then evaluates a text-disjoint 64-context battery on
BLOOM-560M/1.1B, GPT-Neo-125M/1.3B, Mamba-130M/1.4B, Qwen2-0.5B/1.5B and RWKV-1.5B. A prior compatibility context is text-disjoint from the scored battery; no previously probed
model--context cell enters the evaluation.

The spectrum predictor fixed before evaluation is
$\widehat\lambda_k=q_{(k)}$ for $1\le k\le r$, and Eq.~\eqref{eq:imprint} applies the
effective-dimension transform to these predicted eigenvalues. The measured quantity is computed
separately from the revealed spectrum as
$N_{\mathrm{eff}}(\alpha)=\sum_k\lambda_k/(\lambda_k+\alpha)$. Spectral error is the median of
$|\log_{10}(\lambda_k/q_{(k)})|$ over $8\le k\le32$, exponent error is
$|\beta_\lambda-\beta_q|$ over the same ranks, and effective-dimension error is the RMSE over ten fixed
relative dampings between measured $N_{\mathrm{eff}}/r$ and the value obtained by applying the
same transform to the leading $r$ profile values. Family summaries are medians over the complete
model-by-context grid; 95\% intervals use 2,000 crossed-bootstrap draws that resample model and
context axes independently. A separately fixed 16-context-per-model battery compares paired
errors with a same-trace flat spectrum and its effective-dimension curve and, for the spectrum, a random permutation
of profile ranks. Exact eigenvalues are revealed only to score the fixed predictions: the
profile prediction itself uses neither eigendecomposition nor a fitted parameter.

\paragraph{Probability and read-out contributions.}
The mechanism comparison used two model sizes from each of seven families: Pythia, GPT-2, GPT-Neo,
Qwen2, BLOOM, Mamba and RWKV. The 192-context battery was balanced across Wikipedia, AG News and
LAMBADA. A common set of 27,450 decoded singleton outcomes defined conditional model laws, with
read-out rows centred and standardised as above. The weighted profile was compared with the
probability-only profile $(r/V)p_{(k)}$, using the same rank $r$, vocabulary size $V$ and damping
values. Effective-dimension errors were computed across ten damping values on all contexts;
spectral-exponent errors used exact spectra on 48 contexts per model over ranks 8--32. Errors were
summarised by their median within each family. The two contrasts were tested by enumerating all
$2^7$ sign patterns of the family log-error ratios (Supplementary Note~\ref*{SI-si:anatomy}).

\paragraph{Finite-size validation.}
A separate test fits the bounded curve $N_{\mathrm{eff}}/d=[1+(c/c^{\star})^{\gamma}]^{-1}$
and an equal-parameter unbroken power curve to four anchor dampings, then scores three interleaved
held-out dampings on each of 80 curves from ten models in seven families. Median curve-level
error ratios use 5,000 crossed family--model--prompt bootstrap draws; the exact grids and
sampling scheme are given in Supplementary Note~\ref*{SI-si:anatomy}.

\subsection*{Acquisition forecasting and causal evidence depth}

\paragraph{Corpus-based forecast.}
Acquisition time is measured on a logarithmic scale, $\tau=\log_2 t$, where $t$ is the training step at
which a fact's margin first crosses a fixed threshold and persists. Ex-ante corpus $n$-gram margins are
computed at three statistical levels, fixed before any model evaluation; a model-size-specific
$n$-gram predictor is fitted once on the discovery battery and applied without refitting to two fresh
batteries whose prefixes never occur in the discovery battery:
180 facts at Pythia 70M and 160M, and 300 facts at Pythia 410M
(timing scored in $\log_2$ training-step units against constant acquisition-time and label-permutation
baselines).

\paragraph{Randomised evidence depth.}
The causal depth
experiment runs in a randomised controlled synthetic language:
otherwise identical facts are assigned by a cyclic $5\times5$ Latin-square design to five evidence-depth rungs across
paired arms with byte-identical initial weights and batch streams (12 fresh seeds, 45 quintets,
225 distinct facts, a 40,000-step horizon on a 145-point logarithmic grid). The primary
analysis estimates the translation of the fixed lower-tail persistent-acquisition quantile band
($p=0.10$--$0.20$) between the extreme rungs, with crossed seed-by-quintet bootstrap intervals,
leave-one-seed-out ranges and an alternative random-number stream as robustness; the shallow
adjacent contrast is dilation-only, and per-fact additive readings are first-order descriptions
with a measured $n$-gram-margin-by-depth interaction.

\paragraph{Architecture and evidence-construction generalisation.}
A separate fully crossed randomisation tests whether the causal direction generalises across
architecture and evidence construction. Compact four-layer, width-256 GPT-2-, GPT-NeoX- and
Llama-style decoders are each trained on distant copy-back-reference and four-cue-parity
controlled languages. Within each architecture, language and seed, a crossover exchanges shallow
and deep assignment within 384 matched reciprocal-fact blocks while holding initial parameters,
minibatch streams, exposure counts, sequence length and the global shallow:deep mixture fixed
(eight seeds per cell; 96 runs). The response is the mean paired deep-minus-shallow persistent-acquisition
delay at the 0.75-nat margin threshold on a 100-point grid through step 44,800. Confidence intervals
use 10,000 crossed bootstrap draws over seed and matched block. Two control families preserve the
nuisance structure: 16 fixed-treatment blocks per cell and randomisation of deep/shallow labels
within the matched blocks.

\paragraph{Cumulative acquisition and observational composition.}
The cumulative acquisition curves use the 300-fact battery at all three sizes, with nine seeds and 20 checkpoints per size; their held-out
loss coordinate is mean token negative log-likelihood on the same fixed 50,000-token Wikitext
validation prefix. The real-model observational composition uses a
fresh 300-fact battery over two model sizes, nine initialisation seeds and twenty public
checkpoints per size (360 response cells) with four measured components: the $n$-gram predictor
applied without refitting, the observational level schedule, cumulative-curve alignment by held-out
loss and a between-seed variability threshold. The loss-versus-step comparison is descriptive; causal effects are tested by the
randomised experiments above. Numerical resolution was measured before held-out evaluation and is
reported in Supplementary Note~\ref*{SI-si:acquisition}.

\subsection*{Intervention computation, calibration and reusable control}

\paragraph{Matrix-free solves and damping.}
Intervention solves apply $G$ without forming $J$ or $G$, using two operators.

The low-rank apply restricts the law to the top-$|S|$ tokens by mass and renormalises their
probabilities, giving $H \approx B B^\top$ with
$B = W_U[S,:]^\top\,(\mathrm{diag}(p_S) - p_S p_S^\top)^{1/2}$, and $M = J^\top B$ obtained by
$|S|$ vector--Jacobian products; the damped solve $(G+\alpha I)^{-1}q$ then follows from a Woodbury
identity requiring only an $|S|\times|S|$ system in the native $R=I$ implementation.

For the exact apply, a Jacobian--vector product computes $Jv$, the exact full-vocabulary
$H(Jv) = W_U^\top(\mathrm{diag}(p)-pp^\top)W_U(Jv)$
is applied in closed form, and a vector--Jacobian product returns $J^\top(\cdot)$, giving $Gv$ in two
passes using standard matrix-free automatic differentiation~\cite{pearlmutter1994}. The native-reference solve uses conjugate gradients (CG) on $G+\alpha I$, while a general $R$
is handled by whitening or preconditioned generalised CG (Supplementary Note~\ref*{SI-si:instrument}).
Implicit pullback-Fisher products and conjugate-gradient solves also appear in distribution-space
attribution~\cite{martino2026fringe}.

For a reference metric $R$, damping is set as $\alpha=c\,\lambda_{\max}(G,R)$, where the largest
generalised eigenvalue is estimated by power iteration in whitened coordinates and $c$ is typically
$10^{-2}$. The native-reference analyses use $R=I$ and therefore reduce to
$\alpha=c\lambda_{\max}(G)$. The operator can be exactly singular, numerically rank-deficient, or
merely ill-conditioned; these cases are distinguished from deliberate low-rank output truncation.
When needed, the solve is restricted to the numerically resolved eigenspace of $G$. Iterations to a relative
residual of $10^{-6}$ quantify fixed-resolution solver convergence.

\paragraph{Objectives and matched-response calibration.}
For a contrastive pair $(y_w, y_l)$ the objective is the summed log-odds $\phi(h) = \sum_t[\log p(y_w^{(t)}\!\mid\!x,y_w^{<t}) -
\log p(y_l^{(t)}\!\mid\!x,y_l^{<t})]$ over the continuation tokens, with covector
$q = \nabla_h \phi$. The objective spans the full continuation. Its relation to direct preference optimisation
(DPO)~\cite{dpo}, which updates parameters rather than activations, is given in Supplementary Note~\ref*{SI-si:minimal-control}.

Matched-response comparisons use the same realised objective change $\Delta\phi$~\cite{fishback,park2026}.
Each method's step scale is binary-searched, with interpolation between bracketing steps where
necessary; all other endpoints are evaluated at that matched change. The instruction-model
frontiers instead evaluate each prompt-specific direction on a method-specific strength grid shared
across the 12 prompts, average behaviour and off-target KL at each grid point, and compare interpolated
frontiers at equal aggregate mean behaviour change. The amortised contrastive-activation-addition
baseline uses the mean covector direction over the contrast set and the same frontier convention.
The six-size Pythia factual sweep
reports its three requested shifts as nominal targets.

\paragraph{Disturbance and paired steering comparisons.}
Off-target change is measured as off-target KL: the KL of the output change evaluated over the
complement of the objective $\phi$'s support (\texttt{off\_target\_kl\_set}). For scalar objectives the analyses additionally use the
Fisher-orthogonal decomposition $\mathrm{KL}(p_0\Vert p_s) - \Delta\phi^2/(2\,\mathrm{Var}_{p_0}\phi)$,
with $p_s$ the distribution after a steer of magnitude $s$, which isolates the disturbance
orthogonal to the intended change (clamped at zero, as the second-order decomposition can go
slightly negative at large steps). Capability retention on the
frontier for instruction-tuned models is off-target KL on held-out neutral prompts that are semantically disjoint
from the steered attribute. Win rates are computed per prompt at the applicable comparison setting.

In the 12-pair steering comparisons, advantages are disturbance ratios and win rates are the
fraction of pairs with strictly lower natural-gradient disturbance at the comparison setting.
Paired bootstrap draws resample the 12 contrastive pairs in log-ratio space. The same pairs are
used across width (410M versus 1.4B, both 24 layers) and depth (1B at 16 layers versus 1.4B at
24 layers); nominal-target sweep summaries use the two arms at each requested target
(Supplementary Note~\ref*{SI-si:protocols}).

\paragraph{Prediction of relative intervention cost.}
The finite-damping cost ratio
$R_{\mathrm{pred}}=\Pi_G(q)/\Pi_G(A^{-1}q)$, with
$\Pi_G(u)=(u^\top Gu)/(q^\top u)^2$ and $A=G+\alpha I$, is computed from predictor state fixed
before any response is measured. The held-out evaluation spans 11 models from six families,
three objectives, three relative layer depths and 12 held-out contexts per objective (1,188 cells),
with three primary matched-effect magnitudes per cell. Damped natural-gradient and Euclidean interventions
are compared by their realised output KL at matched achieved objective change via binary-searched
step scales. Of 3,564 primary measurements, 3,515 meet the fixed reachability, numerical-floor
and effect-matching criteria. Evaluation reports the median measured-to-predicted ratio and log--log
slope with 10,000 prompt-template-cluster bootstrap draws, improvement over an oracle constant and a
stratified permutation test. To resolve intervention-magnitude dependence, the three fractions are
summarised within each model as geometric means of the measured-to-predicted ratio. Paired layer
intervals resample matched objective--context--magnitude tuples within model while preserving the
available layer tuple, and model-balanced pooled intervals resample models. Numerical resolution
is checked with an fp64 deep ladder, detailed in Supplementary Note~\ref*{SI-si:minimal-control}.

\paragraph{Finite intervention paths.}
Finite paths recomputed both Fisher and Euclidean directions after every accepted step, using
a prompt-KL trust region of $0.02$ nats. Scale was chosen by expansion and bisection to reach
cumulative objective changes of $0.05$, $0.1$, $0.2$ and $0.4$ nats within 1\%. Cumulative local KL
sums the divergence between successive accepted output laws; endpoint KL is measured separately.
The panel contains 14 base models and two instruction-tuned models, giving 18 model--objective
cells with 12 prompts each. Paired log-cost intervals used 4,000 prompt bootstrap draws per cell.

\paragraph{Reusable control.}
Reusable control used a common residual-stream activation update near 45\% of network depth
at the final prompt position, estimated from four donor prompts per objective. Three directions were compared: the Euclidean gradient,
the gradient preconditioned by donor-averaged Fisher, and the gradient preconditioned by Fisher
averaged over four reference prompts. All three methods were relinearised along their paths and
matched to donor log-odds changes of $0.05$, $0.1$ and $0.2$ nats. Reference-Fisher damping was
$0.03$ times the estimated largest eigenvalue. The resulting update was applied unchanged to eight
unseen target prompts per objective without target gradients. Cost was the sum of teacher-forced
next-token KL along each reference continuation, averaged over the four reference sequences;
their prompt-end metrics were used during optimisation. A two-objective solve used the same
reference metric and requested $0.05$ nats for each donor objective (Supplementary Note~\ref*{SI-si:minimal-control}).

\paragraph{Layer and intervention-magnitude dependence.}
Layer dependence is measured at layers $4, 8, 12, 20$ of Pythia-410M using advance-one-step
steering at matched objective change. Conditioning is reported in native coordinates and as
generalised anisotropy relative to $R$. The equality condition
at the intervention layer is $G\propto R$ (or, for a particular objective, support of $q$ in one
generalised eigenspace), in which case natural and reference-gradient directions coincide. A condition
on $H$ alone is insufficient unless $J$ is also a scaled isometry.

The cubic KL correction combines the pulled-back Amari--Chentsov term with a network-Hessian
term; the former alone is complete only at an affine read-out. Nested automatic differentiation
computes both terms and directional finite differences check their sum. Erosion prediction is
analysed on moderate-intervention crossings along monotone scan segments, controlling for the
gentle-regime advantage. Scan details and regime-dependent results are in Supplementary Note~\ref*{SI-si:minimal-control}.

\subsection*{Editing, attribution, dictionary learning and fine-tuning}

\paragraph{Knowledge editing.}
Fisher information has also been used to select model components for knowledge
editing~\cite{xu2026fidal}; the comparisons here concern the intervention and its output disturbance.
For 30 held-out CounterFact edits on GPT-2, Fisher and Euclidean displacements are each
computed once on the canonical prompt, calibrated to the same contrastive target shift and
reused unchanged. Both operators share a gating rule evaluated once per query. The primary
contrast is the paired ratio of other-token KL after both reach the target; intervals bootstrap edits.

A separate GPT-2-XL benchmark applies query-time Fisher steps behind a normalised subject-string
rule. It uses EasyEdit metrics on 100 edits and a stricter probability-margin protocol on
300 edits. Different access and amortisation conditions make these descriptive comparisons
with persisted editors (Supplementary Note~\ref*{SI-si:editing}).

\paragraph{Sparse-feature sensitivity and attribution.}
For a sparse-autoencoder feature with decoder direction $v$, the Fisher-weighted output sensitivity is
$v^\top G v$, the second-order KL coefficient per squared unit displacement along the feature.
For zero ablation of an active sparse-autoencoder (SAE) feature, with displacement $\Delta h$, the corresponding predicted cost is
$\tfrac12\Delta h^\top G\Delta h$. The held-out intervention study sets the active feature to zero before
decoding and measures the resulting full-vocabulary KL divergence. Separate discovery
contexts are used to select features and the token IDs and signs defining the linear objective; all reported causal
outcomes use 18 held-out evaluation-context clusters. The selected-token probability-change fraction is the fraction of the
full-vocabulary absolute probability change on the discovery-selected tokens, and sign agreement
measures whether their nonlinear probability changes have the discovery-defined signs. The
objective-effect-to-Fisher-cost ratio is the squared linear objective effect divided by $\Delta h^\top G\Delta h$;
attribution-patching magnitude is the absolute activation--gradient product. The minimal selective basis
is grown by orthogonal matching pursuit in the Fisher metric on the top eigen-directions of $G$, and compared
with greedy per-feature selection at equal basis size.

\paragraph{Interpretable partitions.}
For interpretable token partitions, probability-weighted group means define a mass-preserving
Fisher coarse-graining. Equal-objective-value groups preserve information about a scalar tilt;
read-out clustering refines the approximation to the full update. Fidelity and the
interpretation--computation tradeoff are evaluated against the ungrouped solve
(Supplementary Note~\ref*{SI-si:minimal-control}).

\paragraph{Dictionary learning.}
Top-$k$ sparse autoencoders are trained and precondition the \emph{decoder} update by the natural
gradient in the code geometry: $\Delta D \propto (\mathbb{E}[a a^\top] + \varepsilon I)^{-1}
\nabla_D \mathcal{L}$, where $a$ are the codes. The preconditioner is applied matrix-free per batch by
a Woodbury solve, never forming the code-Gram matrix. Real residual activations are heavy-tailed; in
the reported configuration, per-token norms are winsorised at approximately the 95th percentile
before training.

Comparisons use three paired seeds. Recovery, reconstruction variance explained, feature use and activation frequency are
reported together.

Disentangling is evaluated against planted ground-truth features by the Hungarian-matched mean
absolute cosine between learned decoder atoms and ground-truth features.
Scale comparisons hold the \emph{per-feature} training budget fixed (Supplementary Note~\ref*{SI-si:dictlearn}).

\paragraph{Low-rank adaptation.}
For low-rank adaptation (LoRA~\cite{lora}) the preconditioner is the generalised Gauss--Newton/Fisher of
the output loss aggregated over examples, $F = \sum_x J_x^\top H_x J_x$, applied to both the $A$
and $B$ adapter factors. Exact matrix-free natural-gradient LoRA is compared against a
Kronecker-factored approximate-curvature (K-FAC) method~\cite{kfac} and against the coarse-grained
metric (Supplementary Note~\ref*{SI-si:instrument}).
Off-target preservation is measured on held-out prompts disjoint from the adaptation set.
The held-out comparison uses rank-4 adapters on Pythia-70M, three matched target gains, eight
outer seeds and 16 held-out preservation prompts. Its primary endpoint is the area under the
preservation-KL frontier. Exact natural-gradient LoRA is compared with a tuned AdamW baseline that
includes an explicit KL penalty. Task transfer and capability retention are measured separately;
these endpoints did not support a broader conclusion, so the reported result is restricted to local
preservation.

\paragraph{Cyclic concepts and relations.}
For a cyclic registry (weekdays, months), the advance-one-step map is fitted as the metric
isometry (rotation) in the Fourier-1 phase representation that best carries each member's distribution to the
next. The rotation angle is compared with the ideal $2\pi/n$.
The off-rotation component is the metric-isometry defect reported in the double dissociation, and the closure defect is the Fisher--Rao
distance between the start point and the composition of $n$ steps.

Relations (capital-of,
gender, tense, comparative) are scored by probability-displacement alignment, the mean cosine alignment of the
probability-displacement directions across instances, reported against a permuted-geometry control
that preserves marginals (Supplementary Note~\ref*{SI-si:motion}).

\subsection*{Statistical analysis and reproducibility}

\paragraph{Statistical analysis.}
Confidence intervals are 95\% unless stated otherwise. Resampling follows each experiment's replication
unit and preserves pairing between methods: contexts or context folds for geometry,
whole language pairs for causal assignment, crossed seeds and facts or matched blocks for
acquisition, and prompts or edits for interventions. Model/family resampling and analysis-specific
exclusions are stated with the corresponding protocols. Cross-model agreement, consensus semantic
alignment, corpus mediation and eigenvector-sharing intervals use 5,000 full-size context draws
with replacement, paired across fixed model rosters and contrasts;
distance-matrix analyses average tied ranks and exclude comparisons between repeated copies
of the same original context. Separately labelled 80\%-subsample ranges describe stability
under context removal (Supplementary Note~\ref*{SI-si:protocols}). Probe accuracies average folds and ordered model pairs.

\paragraph{Compute and reproducibility.}
Model execution and intervention solves use float32, with float64 exceptions specified in the
individual protocols. Most GPU-resident model-forward runs use one NVIDIA RTX 5000 Ada (32\,GB),
with additional runs on NVIDIA A100 GPUs and CPU execution for dense statistical analyses. Full GPU-resident pullback solves fit through 2.8B parameters; at 6.9B the evidence comprises
graph-free spectra and CPU-based pullback calibration/offload analyses, detailed in
Supplementary Note~\ref*{SI-si:repro}. For the reported computational analyses,
applicable random seeds, model revisions, prompts and objective-set definitions were recorded and
are available as described in the Data and Code availability statements
(Supplementary Notes~\ref*{SI-si:protocols} and~\ref*{SI-si:repro}).

\paragraph{Use of generative AI.}
Generative AI tools assisted with research and manuscript preparation. The author independently
verified the outputs, made the substantive decisions and takes full responsibility for the work.

%% file: extended_data.tex
\clearpage
\section*{Extended Data}
\captionsetup{labelformat=empty,labelsep=none}
\setcounter{figure}{0}
\renewcommand{\theHfigure}{ED.\arabic{figure}}

\newcommand{\edgridpanel}[2]{%
  \vtop{%
    \hbox{\makebox[2.547in][l]{{\fontsize{8}{9}\selectfont\sffamily\bfseries #1}}}%
    \vskip 1.5pt
    \hbox{\makebox[2.547in][c]{\includegraphics{#2}}}}}

\begin{figure}[!ht]
  \centering
  \makebox[\linewidth]{\edgridpanel{a}{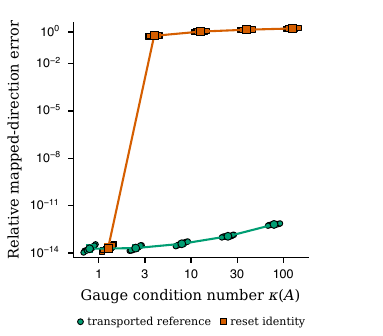}}\par\vspace{6pt}
  \makebox[\linewidth]{\edgridpanel{b}{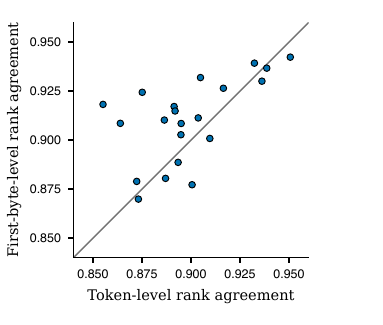}\hspace{\panelsep}\edgridpanel{c}{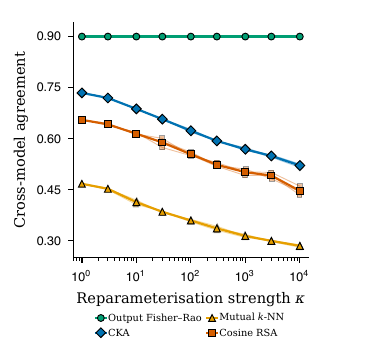}}
  \caption{\textbf{Extended Data Fig.~\thefigure{} $|$ Coordinate covariance and common-outcome controls.}
  \textbf{a}, Coordinate covariance at finite damping. For 12 prompts under one seeded symmetric
  positive-definite gauge family with $\kappa(A)=1,3,10,30,100$, transporting the reference as
  $R'=A^{-\top}RA^{-1}$ reproduces the native damped direction with mapped relative error at most
  $7.4\times10^{-13}$ (green circles). Installing a fresh identity instead changes the regularised
  problem: over non-identity gauges, the mapped-direction error has median 1.23 and minimum 0.54
  (vermilion squares). Points, prompts; connected symbols, medians; error bars, 2.5th--97.5th
  percentiles. Pythia-410M, final
  pre-unembedding activation, top-512 output-Fisher support and $\alpha=0.01\lambda_{\max}$.
  \textbf{b}, Pairwise output-geometry agreement before and after mapping each next-token law to
  its first-byte distribution (21 dyads among eight models). Aggregate agreement is $0.899$ at
  token level and $0.910$ after the push-forward; rank agreement between dyad-level values is
  $0.551$. The grey line marks equality.
  \textbf{c}, Hidden-coordinate reparameterisation at nine anisotropy strengths and three seeds.
  Each point aggregates the same 21 model pairs; faint paths show seeds and heavy paths their
  means. Output Fisher--Rao agreement remains constant, while activation CKA, mutual nearest-neighbour
  overlap and cosine RSA degrade. Small horizontal offsets separate seeds.
  }
  \label{fig:ed-coordinate-controls}
\end{figure}

\begin{figure}[ht]
  \centering
  \makebox[\linewidth]{\panel{a}{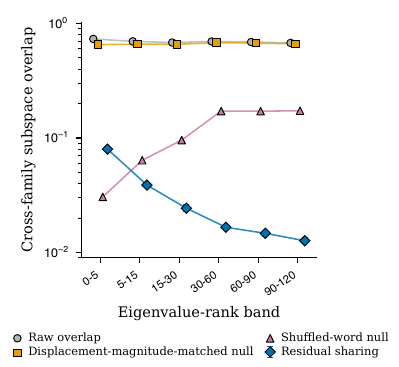}\hspace{\panelsep}\panel{b}{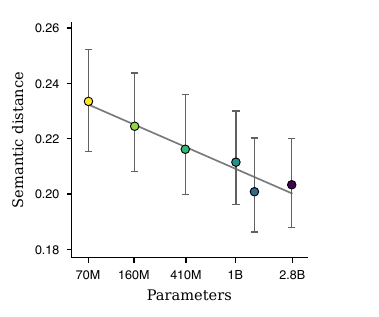}}\par\vspace{9pt}
  \makebox[\linewidth]{\panel{c}{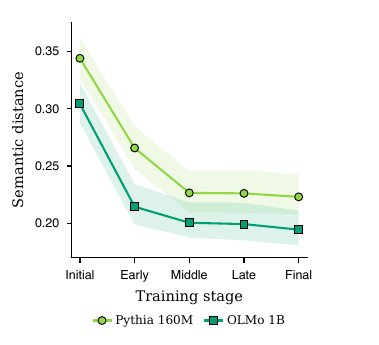}\hspace{\panelsep}\panel{d}{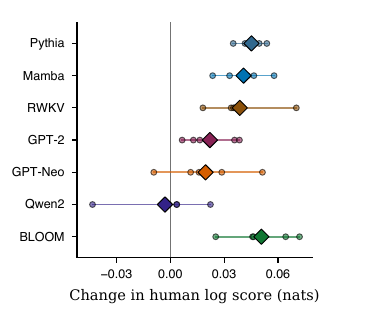}}
  \caption{\textbf{Extended Data Fig.~\thefigure{} $|$ Shared output structure beyond displacement magnitude and on independent human data.}
  \textbf{a}, The 24 displayed estimates are six fixed eigenvalue-rank bands crossed with four summaries:
  raw subspace overlap (grey circles), a displacement-magnitude-matched null that preserves each word's
  displacement magnitude (amber squares), a shuffled-word null (magenta triangles) and the
  residual above the displacement-magnitude-matched null (blue diamonds). Opaque markers have black borders;
  lines connect like summaries across bands. Error bars on the residual span the 2.5th--97.5th
  percentiles across 400 random 80\%-context subsamples. Raw overlap is nearly flat, but the residual
  is concentrated in the leading modes and falls from $0.080$ in ranks 0--5 to $0.013$ in ranks
  90--120. Thus about 89\% of the leading-band
  raw overlap is explained by shared displacement magnitudes, while the smaller residual carries the
  rank-dependent cross-family structure.
  \textbf{b}, Native model samples on 640 word positions from five previously unused narratives.
  Points are the equal-story semantic distances for six final Pythia sizes; bars show conditional 95\%
  bootstrap intervals and the grey line is the registered fit against log parameter count (slope
  $-0.00872$, interval $-0.01202$ to $-0.00556$).
  \textbf{c}, The same distance at five ordered checkpoints from Pythia-160M and OLMo-2-1B.
  Lines show point estimates and shaded regions the conditional 95\% intervals. Final-minus-initial
  changes are $-0.12080$ and $-0.10962$, respectively.
  \textbf{d}, Expected human log-score change after applying the model-only source-trained full 65-outcome map
  instead of each model's native law. Small circles show the five fixed-story differences and diamonds
  their equal-story means; positive values favour calibration. Six of seven family means are positive
  ($0.0196$--$0.0507$ nats); Qwen2 is $-0.0030$ nats. Panels b--d are conditional on the five observed
  stories and make no population-of-stories claim.
  }
  \label{fig:ed-displacement-null}
\end{figure}

\begin{figure}[ht]
  \centering
  \makebox[\linewidth]{\edgridpanel{a}{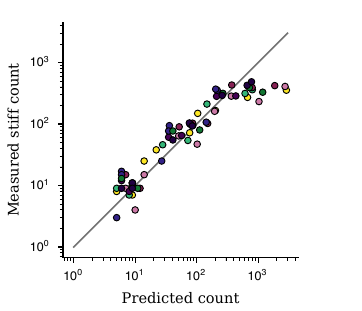}\hspace{\panelsep}\edgridpanel{b}{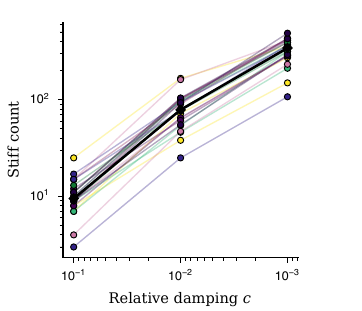}}\par\vspace{2pt}
  \includegraphics{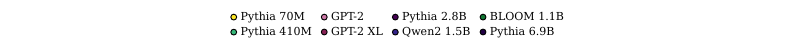}\par\vspace{9pt}
  \makebox[\linewidth]{\edgridpanel{c}{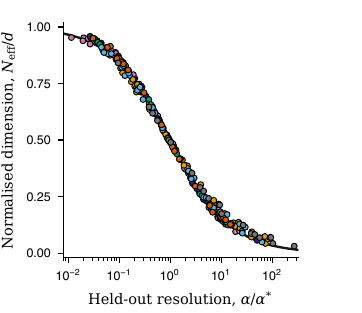}\hspace{\panelsep}\edgridpanel{d}{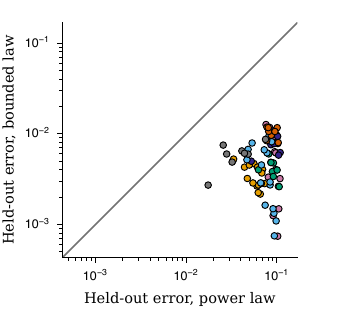}}\par\vspace{2pt}
  \includegraphics{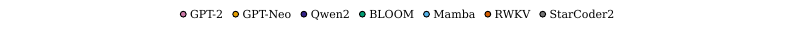}
  \caption{\textbf{Extended Data Fig.~\thefigure{} $|$ Damping reveals bounded effective dimension.}
  \textbf{a}, Predicted versus measured stiff-mode counts for 24 model--context spectra at relative
  damping $10^{-1}$, $10^{-2}$ and $10^{-3}$ (72 points). Predictions are
  $(1/c)^{1/\beta}$ from the power-law exponent fitted over ranks 2--80; colours identify the eight
  models, and the grey line denotes equality.
  \textbf{b}, All 72 resolved-mode counts, joined within each of the 24 spectra; colours identify
  models and black diamonds show the cross-spectrum median. The count grows as damping is lowered.
  The native-reference exponent is empirically stable across the tested models but is not an
  invariant of the metric alone. In held-out estimator
  validation, nominal 95\% interval coverage is $0.90$ for continuous effective dimensions and
  $0.85$ for discrete threshold counts; primary dimension estimates therefore use the continuous
  form (Methods).
  \textbf{c}, Finite-size effective-dimension confirmation on 80 model--prompt curves spanning ten models and
  seven families. For each curve, a bounded effective-dimension curve was fitted at four fixed anchor dampings;
  points are all 240 observations at three interleaved, held-out dampings after rescaling by the
  fitted crossover $\alpha^{*}$. The black line is the bounded effective-dimension curve at the median fitted
  exponent.
  \textbf{d}, Curve-level held-out root-mean-square error for an equal-parameter unbroken power law
  (abscissa) and the bounded effective-dimension curve (ordinate). Every one of 80 points lies below equality. Median
  errors are $0.0814$ and $0.00493$, respectively; the median bounded-to-power error ratio is
  $0.0749$ (95\% crossed-bootstrap interval $0.0428$--$0.1234$).
  }
  \label{fig:ed-effective-dimension}
\end{figure}

\begin{figure}[ht]
  \centering
  \makebox[\linewidth]{\edgridpanel{a}{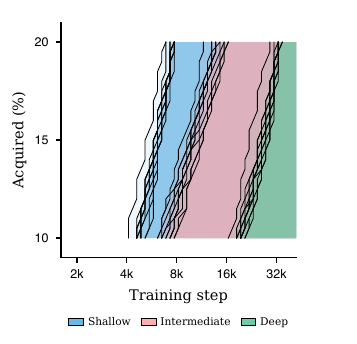}}\par\vspace{6pt}
  \makebox[\linewidth]{\edgridpanel{b}{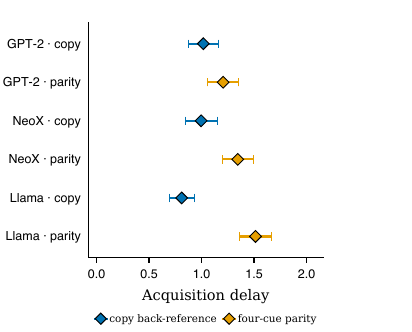}\hspace{\panelsep}\edgridpanel{c}{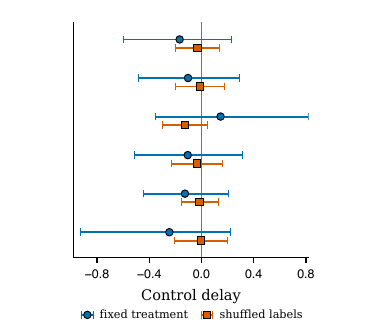}}
  \caption{\textbf{Extended Data Fig.~\thefigure{} $|$ Evidence depth delays acquisition across
  architectures and evidence constructions.}
  \textbf{a}, Randomised evidence depth delays persistent acquisition.
  Curves follow the 10th--20th acquisition-time percentiles for each of 12 training seeds
  of the same transformer, pooling over the same facts and probes within each seed.
  Black curves show individual seeds; blue, pink and green shading to their right marks acquisition
  levels already reached under shallow, intermediate and deep evidence, respectively.
  The pooled deepest-versus-shallowest shift is $\Delta=2.10$ $\log_2$ training-step units
  (crossed seed-by-quintet bootstrap 95\% interval $1.79$--$2.40$), equivalent to
  $4.3\times$ more training steps; the deepest-versus-intermediate contrast is smaller.
  \textbf{b}, Mean paired delay in persistent acquisition for evidence assigned to the deep
  construction relative to otherwise matched evidence in the shallow construction, crossing GPT-2-, GPT-NeoX- and
  Llama-style decoders with copy-back-reference and four-cue-parity controlled languages. Blue
  diamonds denote copy-back-reference cells and amber diamonds four-cue-parity cells; horizontal
  lines are 95\% crossed seed-by-matched-block bootstrap intervals around the cell means
  over eight seeds and 384 matched fact blocks per cell. All six intervals and all 48 seed means are
  positive; effects span $0.810$--$1.514$ $\log_2$ training-step units ($1.75$--$2.86$-fold in training steps). The
  equal-family mean parity-minus-back-reference contrast is $0.414$ $\log_2$ training-step units (95\% interval
  $0.291$--$0.536$).
  \textbf{c}, Corresponding fixed-treatment (blue circles) and label-randomisation (vermilion
  squares) controls. Every one of the
  twelve 95\% intervals contains zero. The grey vertical line in panel c marks zero delay; delays are in $\log_2$ training-step units.}
  \label{fig:ed-acq-generalisation}
\end{figure}

\begin{figure}[ht]
  \centering
  \makebox[\linewidth]{\edgridpanel{a}{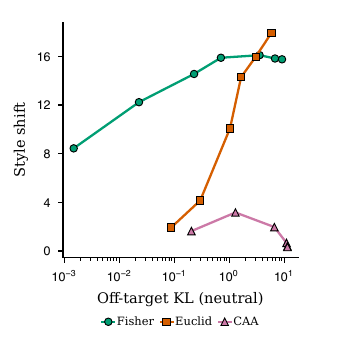}}\par\vspace{6pt}
  \makebox[\linewidth]{\edgridpanel{b}{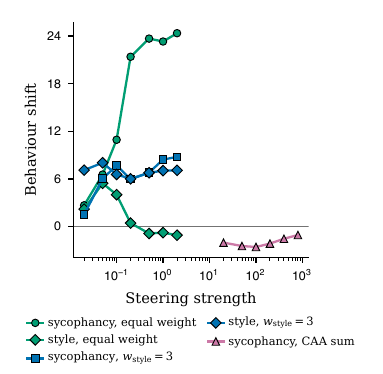}\hspace{\panelsep}\edgridpanel{c}{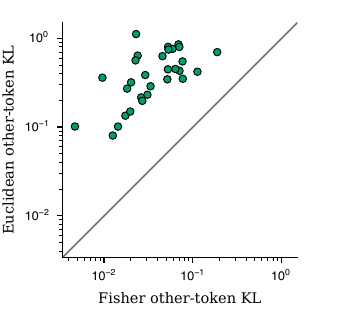}}
  \caption{\textbf{Extended Data Fig.~\thefigure{} $|$ Additional intervention trade-off curves, composition and fixed activation editing.}
  \textbf{a}, The diffuse style objective: behaviour change against off-target change on neutral
  prompts. The complete 19-point operating sweep contains seven Fisher, six Euclidean and six
  contrastive-activation-addition (CAA) settings.
  \textbf{b}, Composition: sycophancy and style steered simultaneously by one solve on the weighted
  covector sum; varying the weight changes the balance between objectives, whereas summed CAA
  vectors change sycophancy in the opposite direction. The 20 distinct operating points produce 34
  displayed objective outcomes; all markers are opaque with black borders.
  \textbf{c}, Matched fixed activation edits on 30 held-out CounterFact records (GPT-2). Each point compares
  Fisher other-token Kullback--Leibler (KL) divergence (abscissa) with Euclidean other-token KL
  divergence (ordinate) after both operators attained the same fixed $+5$ log-odds shift; axes are
  logarithmic and the grey line denotes equality. Euclidean cost is higher for all 30 edits, with a
  geometric-mean Euclidean-to-Fisher ratio of $10.57$ (95\% interval $8.52$--$13.37$).
  }
  \label{fig:ed-intervention-frontiers}
\end{figure}

\begin{figure}[ht]
  \centering
  \makebox[\linewidth]{\edgridpanel{a}{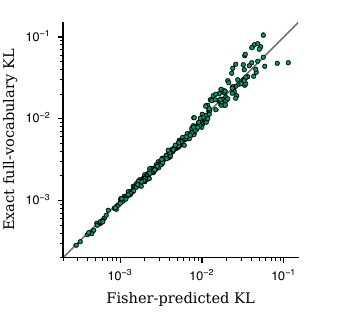}}\par\vspace{6pt}
  \makebox[\linewidth]{\edgridpanel{b}{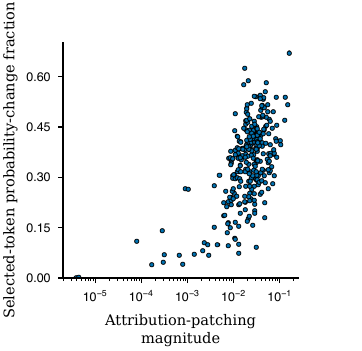}\hspace{\panelsep}\edgridpanel{c}{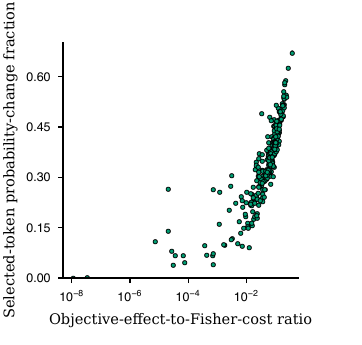}}
  \caption{\textbf{Extended Data Fig.~\thefigure{} $|$ Output geometry predicts the causal cost and selectivity of sparse features.}
  Pythia-410M, layer 12, Top-$k$ sparse autoencoder. Points in every panel are the same 303 unique
  zero ablations of active sparse-autoencoder (SAE) features from 57 evaluation-active features across 18 held-out
  context clusters; feature and objective selection used 18 disjoint discovery contexts. Opaque
  green points in \textbf{a} and \textbf{c} mark Fisher quantities, while blue points in
  \textbf{b} mark attribution patching; every point has a black border.
  \textbf{a}, The Fisher prediction $\tfrac12\Delta h^\top G\Delta h$ against exact
  full-vocabulary $\mathrm{KL}(p_0\Vert p_{\mathrm{zero}})$ (Spearman $\rho=0.997$); grey line,
  equality.
  \textbf{b}, Attribution-patching magnitude against the selected-token probability-change fraction, the fraction of
  full-vocabulary absolute probability change on discovery-selected tokens ($\rho=0.506$).
  \textbf{c}, The objective-effect-to-Fisher-cost ratio against the same endpoint ($\rho=0.935$); its correlation
  exceeds that of attribution patching by $0.429$ (95\% context-cluster bootstrap interval
  $0.296$--$0.553$). The comparison concerns concentration on selected tokens; attribution-patching magnitude remains
  the stronger predictor of absolute effect magnitude.
  }
  \label{fig:ed-feature-selectivity}
\end{figure}

\begin{figure}[ht]
  \centering
  \includegraphics{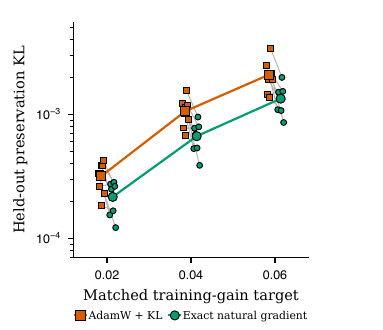}
  \caption{\textbf{Extended Data Fig.~\thefigure{} $|$ Natural-gradient low-rank adaptation lowers held-out preservation cost at matched training gain.}
  Exact natural-gradient LoRA is compared with tuned AdamW with explicit KL regularisation using
  rank-4 adapters on Pythia-70M. Points are all 48 method--target--seed cells from eight paired
  outer seeds, three fixed matched training-gain targets and two methods; each cell averages
  16 held-out preservation prompts. Vermilion squares denote AdamW+KL and green circles exact
  natural gradient. Pale lines join paired methods within seed and target, horizontal offsets are
  fixed display jitter, and the larger connected symbols are method means.
  The ordinate is logarithmic. Every method reached every target. Across this frontier,
  AdamW+KL incurred $1.57\times$ the preservation-KL AUC of exact natural gradient (95\% paired
  hierarchical bootstrap interval $1.14$--$1.98$). The experiment did not support a broader
  task-transfer or capability-retention conclusion.
  }
  \label{fig:ed-lora-preservation}
\end{figure}

\begin{figure}[ht]
  \centering
  \makebox[\linewidth]{\edgridpanel{a}{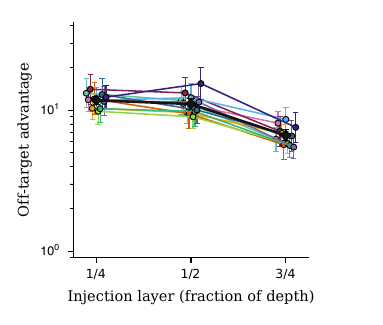}\hspace{\panelsep}\edgridpanel{b}{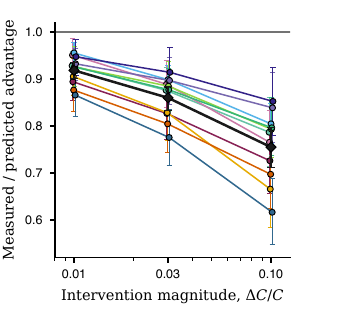}}\par\vspace{2pt}
  \includegraphics{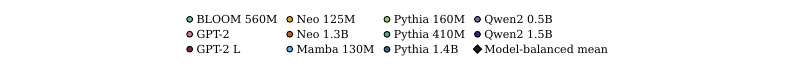}
  \caption{\textbf{Extended Data Fig.~\thefigure{} $|$ Layer and intervention magnitude delimit the geometric correction.}
  \textbf{a}, Off-target advantage by injection layer for 3,515 usable matched-response observations
  from 1,188 model--layer--objective--context cells across 11 models and six families, summarized as
  33 within-model estimates. Coloured circles are geometric means and vertical bars are 95\% paired
  objective--context--magnitude bootstrap intervals that preserve the available layer tuple; lines
  connect the three depths within model. Black diamonds
  are model-balanced geometric means with 95\% model-bootstrap intervals. The pooled means fall from $11.79$ to $11.08$ to $6.62$ at quarter, half and three-quarter
  depth, respectively. Every model mean is lower at three-quarter than at quarter depth; nine of
  eleven paths decrease at both adjacent steps, with one intermediate-depth increase for Mamba
  130M and Qwen2 1.5B.
  \textbf{b}, Measured-to-predicted off-target advantage at intervention fractions
  $\Delta C/C=0.01,0.03,0.10$. Coloured circles are all 33 within-model geometric means (11 models
  crossed with three fractions), with 95\% paired-cell bootstrap intervals; lines connect fractions
  within model. Black diamonds are model-balanced geometric means with model-bootstrap intervals,
  and the grey line denotes exact calibration. All 11 model paths decrease strictly. The
  model-balanced means are $0.918$, $0.860$ and $0.755$, quantifying the expected attenuation of a
  local prediction as intervention magnitude grows.
  }
  \label{fig:ed-layer-magnitude}
\end{figure}

\begin{figure}[ht]
  \centering
  \makebox[\linewidth]{\panel{a}{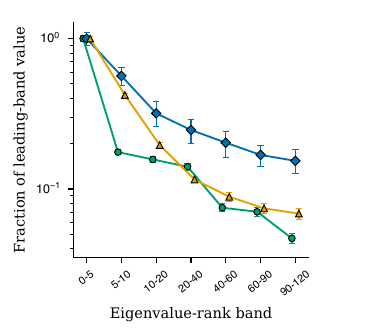}\hspace{\panelsep}\panel{b}{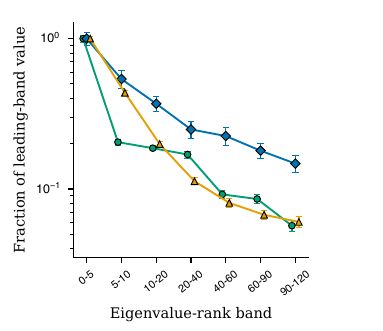}}\par\vspace{2pt}
  \includegraphics{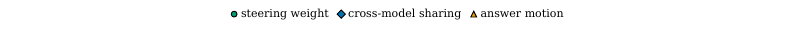}
  \caption{\textbf{Extended Data Fig.~\thefigure{} $|$ Output sensitivity, cross-model sharing and answer
  motion concentrate in the same native-reference modes.}
  Each panel shows all 21 summaries (seven predefined eigenvalue-rank bands $\times$
  three quantities) from six models: the band's eigenvalue-mass share, cross-model eigenvector
  sharing against the displacement-magnitude-matched null and probability motion on the highest-probability
  tokens. To compare their decay directly, each quantity and its 95\% context-bootstrap interval
  are divided by that quantity's leading-band point estimate. On natural text the top five modes carry 55\% of the eigenvalue mass, the largest
  residual sharing ($+0.069$, 95\% interval $0.062$--$0.077$) and the largest answer motion
  ($0.59$), all three decaying together to the last band.
  \textbf{a}, Natural-text battery. \textbf{b}, Templated battery.
  }
  \label{fig:ed-colocalisation}
\end{figure}

\clearpage
\subsection*{Extended Data note: cyclic concepts and two geometries of behaviour}
\begin{figure}[htbp]
  \centering
  \makebox[\linewidth]{\edgridpanel{a}{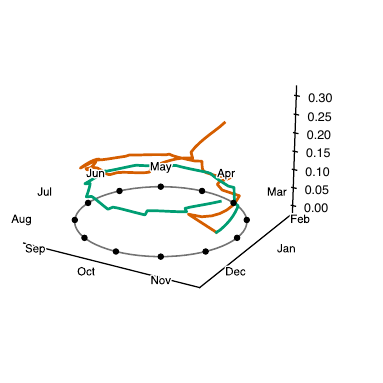}\hspace{\panelsep}\edgridpanel{b}{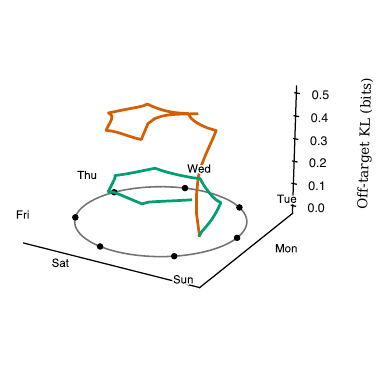}}\par\vspace{2pt}
  \includegraphics{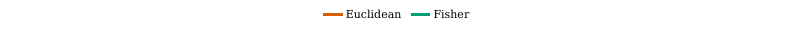}\par\vspace{6pt}
  \makebox[\linewidth]{\edgridpanel{c}{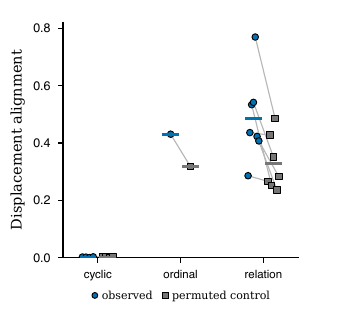}}
  \caption{\textbf{Extended Data Fig.~\thefigure{} $|$ Two geometries of behaviour: cyclic reweighting and
  relation-induced probability displacement} (Pythia-410M).
  \textbf{a},\textbf{b}, Walking the ring by steering month$\to$month (\textbf{a}) or
  day$\to$day (\textbf{b}) around the full cycle at matched concept progress: the plane is the
  concept's phase circle in output space and height is accumulated off-target change
  (bits). Both paths traverse the same ordered concept cycle; the reference phase sectors are
  separated, by construction, by $2\pi/12$ (or $2\pi/7$). After one circuit the endpoint returns
  towards the starting sector with a method-dependent residual displacement, while the
  natural-gradient path accumulates less off-target change.
  \textbf{c}, The dissociation from analogy relations. Paired circles and squares show probability-displacement
  alignment and its permuted control for all twelve concept sets across three categories (24
  displayed values); thin grey lines join each concept's pair and short horizontal bars show
  category means. Relations move mass along a shared probability-displacement direction (mean alignment
  $0.49$; permuted control $0.33$), while cyclic concepts show none ($0.00$) and are
  near-isometries of the metric. The shared colour key applies to both trajectory panels.}
  \label{fig:cyclic}
\end{figure}

Beyond local correction of the preceding operations, the output metric also reveals global
structure in selected concept sets. In the Fourier-1 phase coordinates used in Extended Data
Fig.~\ref{fig:cyclic}a,b, the days of the week and months of the year occupy ordered, equally spaced reference
sectors. Steering once around either full cycle returns the concept phase towards its starting
sector, with the displayed method-dependent residual displacement. At each transition, the
natural-gradient step disturbs unrelated tokens less than the Euclidean step.

Analogy relations behave differently. Pairs such as capital-of, gender and tense do not reweight a
shared set of tokens; they move probability mass between disjoint sets (Paris to Tokyo, \emph{he} to
\emph{she}). The two classes separate as a double dissociation
(Extended Data Fig.~\ref{fig:cyclic}c): across twelve concepts, relations move mass along a shared probability-displacement
direction (high probability-displacement alignment, $0.49$ against $0.33$ for a permuted-geometry control,
essentially zero for cyclic concepts), while cyclic concepts are near-isometries of the metric,
with a low isometry defect ($0.29$ against $1.02$ for relations); ordinal concepts fall between.
The pullback metric characterises probability reweighting, whereas movement between disjoint token
sets additionally requires a ground metric on tokens; the corresponding
structure in embedding space is analysed in a related symmetry study~\cite{cyclicsymmetry}.